%% file: main.tex
\documentclass[runningheads]{llncs}
\usepackage{graphicx}
\usepackage{comment}
\usepackage{amsmath,amssymb} 
\usepackage{color}
\usepackage{epsfig}
\usepackage{multirow}
\usepackage{booktabs}
\usepackage[ruled,vlined]{algorithm2e}
\usepackage{microtype}      
\usepackage{alphalph}
\usepackage{pifont} 
\usepackage{CJKutf8}
\usepackage{wrapfig,lipsum,booktabs}
\usepackage[dvipsnames]{xcolor}
\usepackage{tabularx}
\input{defs}

\newcommand\blfootnote[1]{%
  \begingroup
  \renewcommand\thefootnote{}\footnote{#1}%
  \addtocounter{footnote}{-1}%
  \endgroup
}
\usepackage[pagebackref=true,breaklinks=true,colorlinks,bookmarks=false]{hyperref}

\begin{document}
\pagestyle{headings}
\mainmatter
\def\ECCVSubNumber{4220}  

\title{Measuring Generalisation to Unseen Viewpoints, Articulations, Shapes and Objects for 3D Hand Pose Estimation under Hand-Object Interaction} 



\author{Anil Armagan$^1$ \and Guillermo Garcia-Hernando$^{1,2}$ \and Seungryul Baek$^{1,20}$ \and Shreyas Hampali$^3$ \and Mahdi Rad$^3$ \and Zhaohui Zhang$^4$ \and Shipeng Xie$^4$ \and MingXiu Chen$^4$ \and Boshen Zhang$^5$ \and Fu Xiong$^6$ \and Yang Xiao$^5$ \and Zhiguo Cao$^5$ \and Junsong Yuan$^7$ \and Pengfei Ren$^8$ \and Weiting Huang$^8$ \and Haifeng Sun$^8$ \and Marek Hr\'{u}z$^9$ \and Jakub Kanis$^9$ \and Zden\v{e}k Kr\v{n}oul$^9$ \and Qingfu Wan$^{10}$ \and Shile Li$^{11}$ \and Linlin Yang$^{12}$ \and Dongheui Lee$^{11}$ \and Angela Yao$^{13}$ \and Weiguo Zhou$^{14}$ \and Sijia Mei$^{14}$ \and Yunhui Liu$^{15}$ \and Adrian Spurr$^{16}$ \and Umar Iqbal$^{17}$ \and Pavlo Molchanov$^{17}$ \and Philippe Weinzaepfel$^{18}$ \and Romain Br\'{e}gier$^{18}$ \and Gr\'{e}gory Rogez$^{18}$ \and Vincent Lepetit$^{3,19}$ \and Tae-Kyun Kim$^{1,21}$}

\authorrunning{Armagan et al.}
\titlerunning{Measuring Generalisation for 3D Hand Pose Estimation}
\institute{}
\maketitle

\begin{abstract}
We study how well different types of approaches generalise in the task of 3D hand pose estimation under single hand scenarios and hand-object interaction. We show that the accuracy of state-of-the-art methods can drop, and that they fail mostly on poses absent from the training set. Unfortunately, since the space of hand poses is highly dimensional, it is inherently not feasible to cover the whole space densely, despite recent efforts in collecting large-scale training datasets. This sampling problem is even more severe when hands are interacting with objects and/or inputs are RGB rather than depth images, as RGB images also vary with lighting conditions and colors. To address these issues, we designed a public challenge (\hands) to evaluate the abilities of current 3D hand pose estimators~(HPEs) to interpolate and extrapolate the poses of a training set. More exactly, \hands is designed (a) to evaluate the influence of both depth and color modalities on 3D hand pose estimation, under the presence or absence of objects; (b) to assess the generalisation abilities \wrt~four main axes: shapes, articulations, viewpoints, and objects; (c) to explore the use of a synthetic hand models to fill the gaps of current datasets. Through the challenge, the overall accuracy has dramatically improved over the baseline, especially on extrapolation tasks, from 27mm to 13mm mean joint error. Our analyses highlight the impacts of: Data pre-processing, ensemble approaches, the use of a parametric 3D hand model (MANO), and different HPE methods/backbones.
\end{abstract}

\blfootnote{
$^1$Imperial College London,
$^2$Niantic, Inc.,
$^3$Graz Uni. of Technology,
$^4$Rokid Corp. Ltd.,
$^5$HUST,
$^6$Megvii Research Nanjing,
$^7$SUNY Buffalo,
$^8$BUPT,
$^9$Uni. of West Bohemia,
$^{10}$Fudan Uni.,
$^{11}$TUM,
$^{12}$Uni. of Bonn,
$^{13}$NUS,
$^{14}$Harbin Inst. of Technology,
$^{15}$CUHK,
$^{16}$ETH Zurich,
$^{17}$NVIDIA Research,
$^{18}$NAVER LABS Europe,
$^{19}$ENPC ParisTech,
$^{20}$UNIST,
$^{21}$KAIST. \\Challenge webpage: \url{ https://sites.google.com/view/hands2019/challenge}
}


\section{Introduction}
\label{intro}

3D hand pose estimation is crucial to many applications including natural user-interaction in AR/VR, robotics, teleoperation, and healthcare. The recent successes primarily come from large-scale training sets~\cite{yuan2017bighand}, deep convolutional neural networks~\cite{resnet50,hand_iccvw_2017_1}, and fast optimisation for model fitting~\cite{pso,forth}. State-of-the-art methods now deliver satisfactory performance for viewpoints seen at training time and single hand scenarios. However, as we will show, these methods substantially drop accuracy when applied to egocentric viewpoints for example, and in the presence of significant foreground occlusions. These cases are not well represented on the training sets of existing benchmarks~\cite{gui_fpha_cvpr2018,GANeratedHands_CVPR2018,shhand}. The challenges become even more severe when we consider RGB images and hand-object interaction scenarios. These issues are well aligned with the observations from the former public challenge HANDS'17~\cite{shanxin_cvpr2018}: The state-of-the-art methods dropped accuracy from frontal to egocentric views, and from open to closure hand postures. The average accuracy was also significantly lower under hand-object interaction~\cite{gui_fpha_cvpr2018}.

\begin{figure*}[b!]
\centering
\begin{tabular}{c c c}
{\includegraphics[width=0.3\linewidth,height=0.24\linewidth]{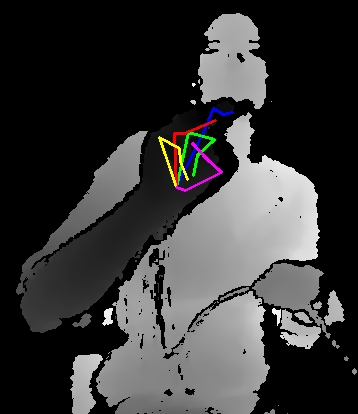}}
& 
{\includegraphics[width=0.3\linewidth,height=0.24\linewidth]{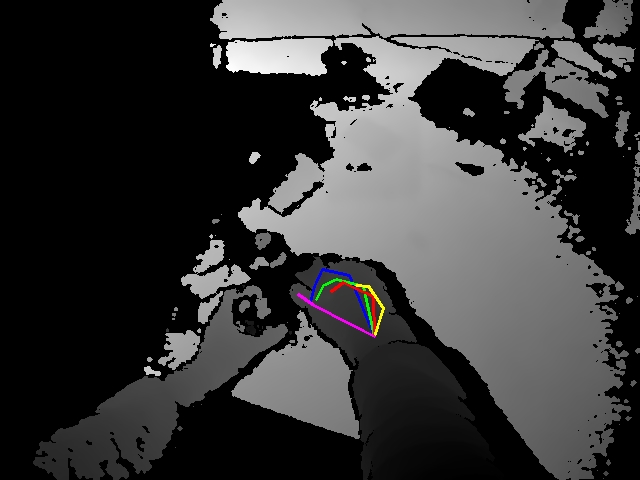}}
& 
{\includegraphics[width=0.3\linewidth,height=0.24\linewidth]{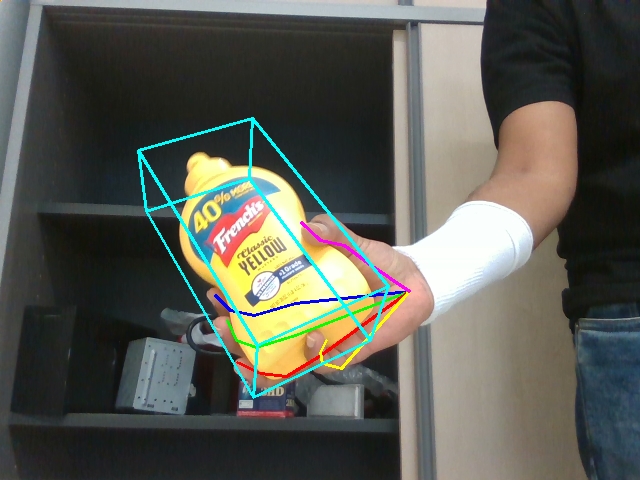}}

\\
{\includegraphics[width=0.3\linewidth,height=0.24\linewidth]{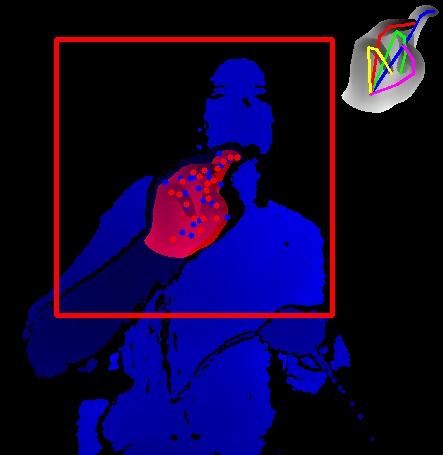}} 
& 
{\includegraphics[width=0.3\linewidth,height=0.24\linewidth]{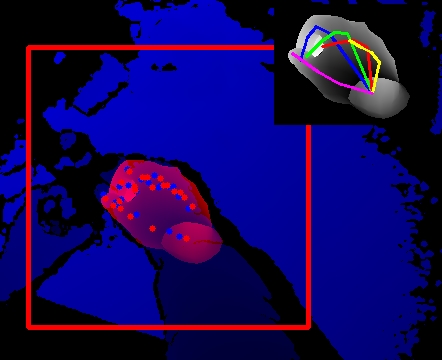}}
& 
{\includegraphics[width=0.3\linewidth,height=0.24\linewidth]{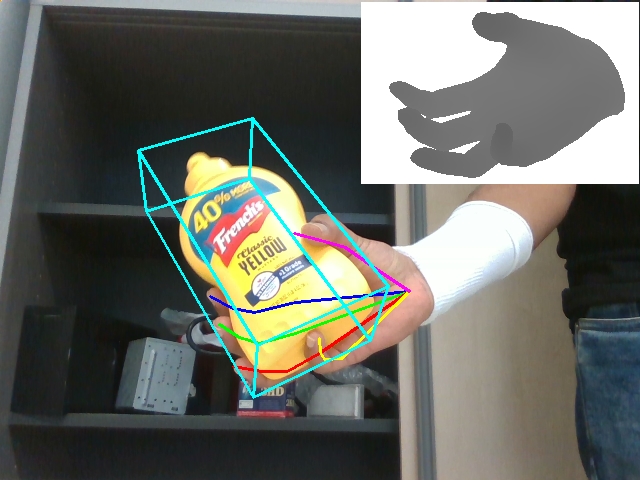}}
\\
{\small (a) Task 1} & {\small (b) Task 2} & {\small (c) Task 3}\vspace{-10pt}
\end{tabular}
\caption{Frames from the three tasks of our challenge. For each task, we show the input depth or RGB image with the ground-truth hand skeleton (top) and a rendering of the fitted 3D hand model as well as a depth rendering of the model (bottom). The ground-truth and estimated joint locations are shown in blue and red respectively.
}
\label{fig:tasks_vis} 
\end{figure*}

Given the difficulty to interpolate and extrapolate poses from the training set, one may opt for creating even larger training sets. Unfortunately, an inherent challenge in 3D hand pose estimation is the very high dimensionality of the problem, as hand poses, hand shapes and camera viewpoints have a large number of degrees-of-freedom that can vary independently. This complexity increases even more when we consider the case of a hand manipulating an object. 
Despite the recent availability of large-scale datasets~\cite{yuan2017bighand}, and the development of complex calibrated multi-view camera systems to help the annotation or synthetic data~\cite{multiviewsystem,hand_cvpr_2017_1,iccv_2017_zimmerman}, capturing a training set that covers completely the domain of the problem remains extremely challenging.

In this work, we therefore study in depth the ability of current methods to interpolate and extrapolate the training set, and how this ability can be improved. To evaluate this ability, we consider the three tasks depicted in Fig.~\ref{fig:tasks_vis}, which vary the input (depth and RGB images) or the camera viewpoints, and introduce the possible manipulation of an object by the hand.
We carefully designed training and testing sets in order to evaluate the generalisation performance to unseen viewpoints, articulations, and shapes of the submitted methods.

\begin{figure}[!t]
    \centering
    \includegraphics[width=0.45\linewidth,height=0.35\linewidth]{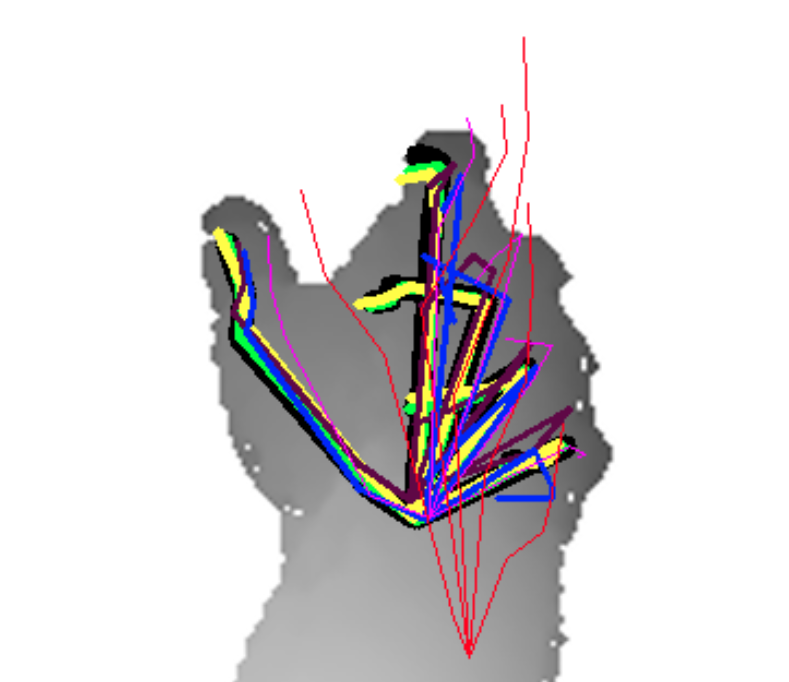}
    \caption{Visualization of \textbf{ground-truth} hand pose and poses with varying level of MJEs, {\color{green}{$<5mm$}}, {\color{yellow}{$<10mm$}}, {\color{blue}{$<20mm$}}, {\color{purple}{$<30mm$}}, {\color{magenta}{$<40mm$}, {\color{red}{$<60mm$}}}. MJE (mm) of the visualized poses are $1.75$, $6.88$, $13.94$, $15.32$, $35.67$, $52.15$, respectively. Best viewed in color.
    }
    \label{fig:varying_handpose_errors}
\end{figure}

\Hands fostered dramatic accuracy improvement compared to a provided baseline, which is a ResNet-50~\cite{resnet50}-based 3D joint regressor trained on our training set, from $\bm{27mm}$ to $\bm{13mm}$. Please see Fig.~\ref{fig:varying_handpose_errors} for visualization of varying level of hand pose errors. This paper provides an in-depth analysis of the different factors that made this improvement possible.

\input{input_Overview}

\input{input_Evaluation_criteria}

\input{input_Dataset_details}

\input{input_Evaluated_methods}

\input{input_Results_and_analysis}

\input{input_supp_analysis}

\clearpage
\section{Conclusion}
We carefully designed structured training and test sets for 3D HPEs and organized a challenge for the hand pose community to show state-of-the-art methods still tend to fail to extrapolate on large pose spaces. Our analyses highlight the impacts of using ensembles, the use of synthetic images, different type of HPEs \eg~2D, 3D or local-estimators and post-processing. Ensemble techniques, both methodologically in 2D and 3D HPEs and in post-processing, help many approaches to boost their performance on  extrapolation.  
The submitted HPEs were proven to be successful while interpolating in all the tasks, but their extrapolation capabilities vary significantly.  Scenarios such as hands interacting with objects present the biggest challenges to extrapolate by most of the evaluated methods both in depth and RGB modalities. \\
Given the limited extrapolation capabilities of the methods, usage of synthetic data is appealing. Only a few methods actually were making use of synthetic data to improve extrapolation.
$570K$ synthetic images used by the winner of Task 1 is still a very small number compared to how large, potentially infinite, it could be. We believe that investigating these possibilities, jointly with data sub-sampling strategies and real-synthetic domain adaptation is a promising and interesting line of work. The question of what would be the outcome if we sample `dense enough' in the continuous and infinite pose space and how 'dense enough' is defined when we are limited by hardware and time is significant to answer.
\\
\noindent \textbf{Acknowledgements.}
This work is partially supported by Huawei Technologies Co. Ltd. and Samsung Electronics. S. Baek was supported by IITP funds from MSIT of Korea (No. 2020-0-01336 AIGS of UNIST, No. 2020-0-00537 Development of 5G based low latency device - edge cloud interaction technology).

\clearpage
%
%
\bibliographystyle{splncs04}
\bibliography{egbib}

\appendix
\input{appendix}

\end{document}

%% file: defs.tex
\usepackage[normalem]{ulem}
\usepackage{xspace}

\makeatletter
\DeclareRobustCommand\onedot{\futurelet\@let@token\@onedot}
\def\@onedot{\ifx\@let@token.\else.\null\fi\xspace}
\def\eg{\emph{e.g}\onedot}

\def\wrt{\emph{w.r.t}\onedot} 
\def\etal{{et al}\onedot}
\makeatother

\newcommand{\taskoneone}{\textit{Rokid}\xspace}
\newcommand{\taskonetwo}{\textit{A2J}\xspace}
\newcommand{\taskonethree}{\textit{AWR}\xspace}
\newcommand{\taskonefour}{\textit{NTIS}\xspace}
\newcommand{\taskonefive}{\textit{Strawberryfg}\xspace}
\newcommand{\taskonesix}{\textit{BT}\xspace}

\newcommand{\tasktwoone}{\textit{NTIS}\xspace}
\newcommand{\tasktwotwo}{\textit{A2J}\xspace}
\newcommand{\tasktwothree}{\textit{CrazyHand}\xspace}
\newcommand{\tasktwofour}{\textit{BT}\xspace}

\newcommand{\taskthreeone}{\textit{ETH\_NVIDIA}\xspace}
\newcommand{\taskthreetwo}{\textit{NLE}\xspace}
\newcommand{\taskthreethree}{\textit{BT}\xspace}

\newcommand{\hands}{HANDS'19\xspace}
\newcommand{\Hands}{HANDS'19\xspace}

\newcommand{\Handschallenge}{HANDS 2019 Challenge}

\newcommand{\pose}{\mathcal{P}_{6D}}

\newcommand{\bm}[1]{\mbox{\boldmath{$#1$}}}

\newcommand{\cmark}{\ding{51}}%
\newcommand{\xmark}{\ding{55}}%

\usepackage{mathtools}
\usepackage{xspace}

\newcommand{\C}{\mathbb{C}}

\def\C{\mathbb{C}}
\def\N{\mathbb{N}}

\newcommand{\ignore}[1]{}

%% file: input_Overview.tex
\section{\Handschallenge~Overview}
\label{sec:handsoverview}
The challenge consists of three different tasks, in which the goal is to predict the 3D locations of the hand joints given an image. For training, images, hand pose annotations, and a 3D parametric hand model~\cite{mano} for synthesizing data are provided. For inference, only the images and bounding boxes of the hands are given to the participants. These tasks are defined as follows:

\textbf{Task 1}: Depth-Based 3D Hand Pose Estimation: This task builds on BigHand2.2M~\cite{yuan2017bighand} dataset, as for the HANDS 2017 challenge~\cite{yuan20172017}. No objects appear in this task. Hands appear in both third person and egocentric viewpoints.

\textbf{Task 2}: Depth-Based 3D Hand Pose Estimation while Interacting with Objects: This task builds on the F-PHAB dataset~\cite{gui_fpha_cvpr2018}. The subject  manipulates objects with their hand, as captured from an egocentric viewpoint. Some object models are provided by~\cite{gui_fpha_cvpr2018}.

\textbf{Task 3}: RGB-Based 3D Hand Pose Estimation while Interacting with Objects: This task builds on the HO-3D~\cite{ho3d_2019} dataset. The subject manipulates objects with their hand, as captured from a third person viewpoint. The objects are used from the YCB dataset~\cite{ycbdata}. The ground truth wrist position of the test images is also provided in this task.

The BigHand2.2M~\cite{yuan2017bighand} and F-PHAB~\cite{gui_fpha_cvpr2018} datasets have been used by 116 and 123 unique institutions to date. \hands received 80 requests to access the datasets with the designed partitions, and 17, 10 and 9 participants have evaluated their methods on Task 1, Task 2 and Task 3, respectively.


%
%


%% file: input_Evaluation_criteria.tex
\section{Evaluation Criteria}
\begin{figure}[!b]
    \centering
    \begin{minipage}{0.48\textwidth}
    \centering
    \includegraphics[width=1.\linewidth,height=0.55\linewidth]{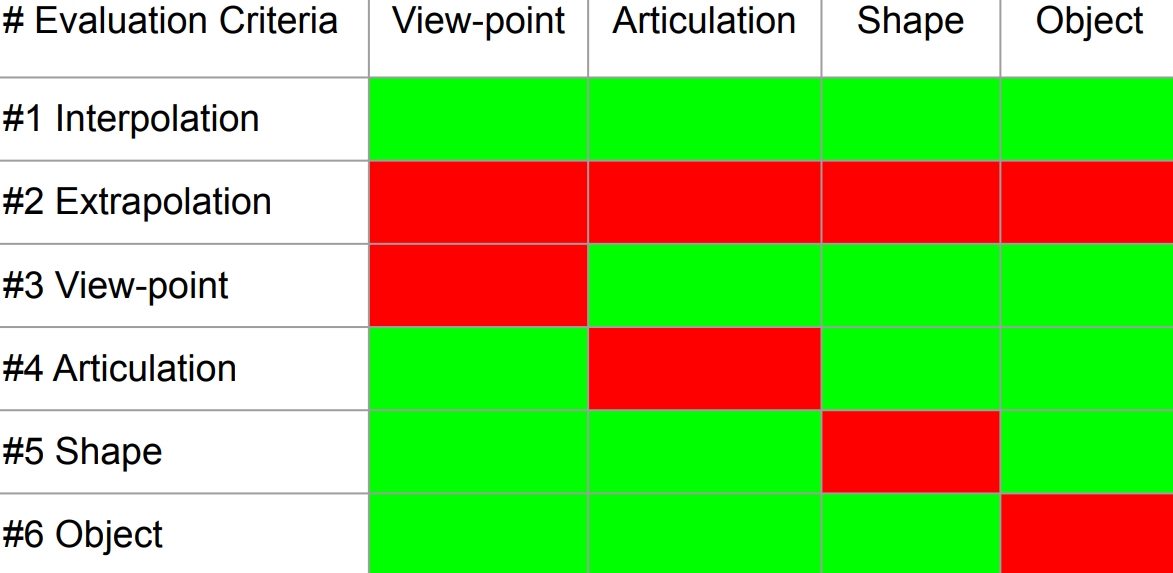}
    \end{minipage}
    \begin{minipage}{0.48\textwidth}
    \centering
    \includegraphics[width=0.8\linewidth,height=0.65\linewidth]{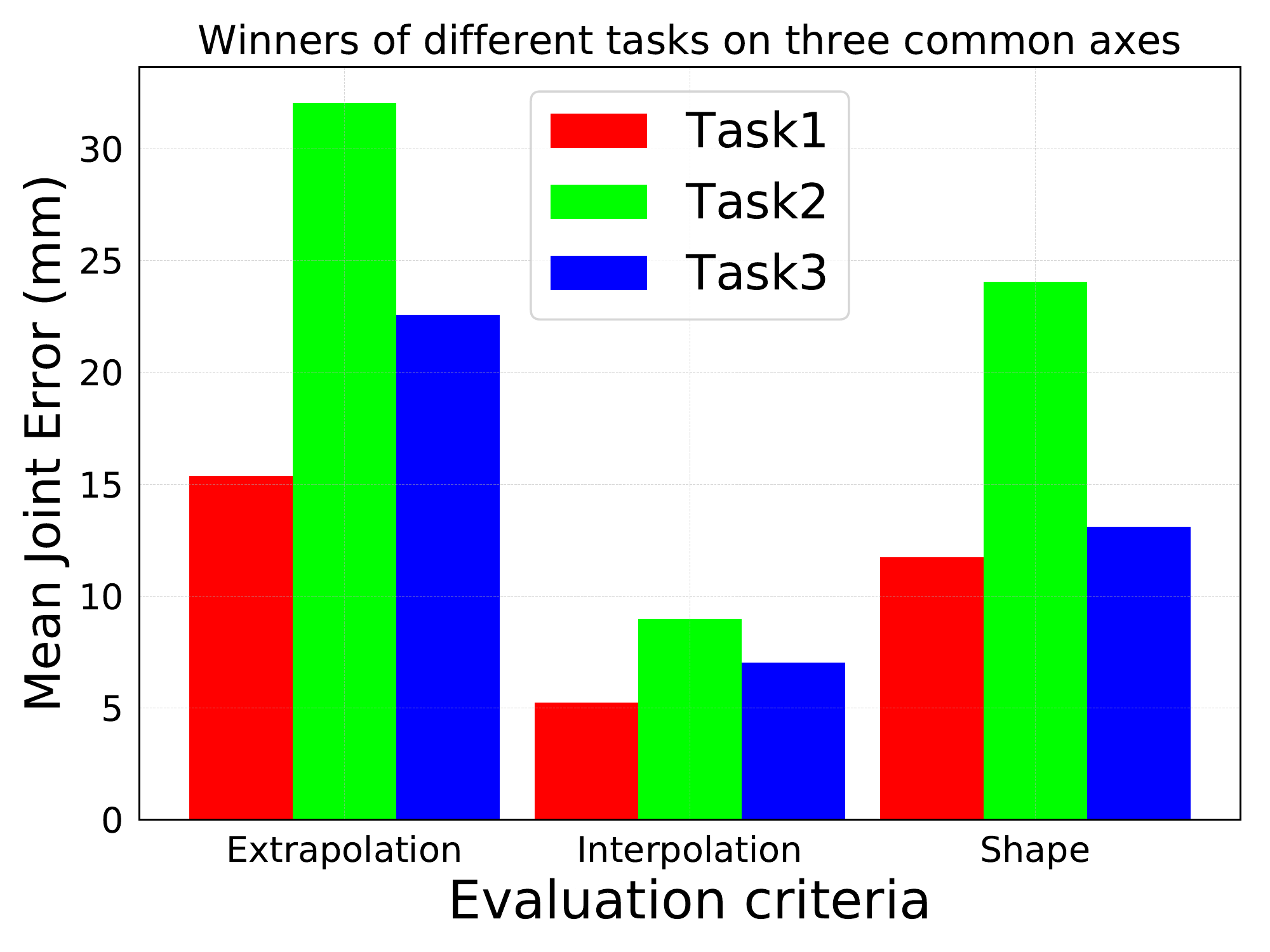}
    \end{minipage}
    \caption{Left: The six evaluation criteria used in the challenge. For each axis (Viewpoint, Articulation, Shape, Object), we indicate if hand poses in an evaluation criterion are also available (green) in the training set or not (red). Right: MJE comparison of the best methods for the Extrapolation, Interpolation and Shape criteria on each task.} 
    \label{fig:evalstrategy}
\end{figure}

We evaluate the generalisation capability of HPEs in terms of four ``axes": Viewpoint, Articulation, Shape, and Object. For each axis, frames within a dataset are automatically annotated by using the ground-truth 3D joint locations and the object information to annotate each frame in each axis. The annotation distribution of the dataset for each axis are used are used to create a training and a test set. Using the frame annotations on each axis, the sets are sampled in a structured way to have the test frames that are similar to the frames in the training data (for interpolation) and also the test frames where axes' annotations are never seen in the training data (for extrapolation). More details on the dataset are given in Section~\ref{sec:datasets}. To measure the generalisation of HPEs, six evaluation criteria are further defined with the four main axes: 

\textbf{Viewpoint}, \textbf{Articulation}, \textbf{Shape} and \textbf{Object} are respectively used for measuring the extrapolation performance of HPEs on the frames with articulation cluster, viewpoint angle, hand shape and object type (axis annotations) that are not present in the training set. \textbf{Extrapolation} is used to measure the performance on the frames with axis annotations that do not overlap/present in the training set. Lastly, \textbf{Interpolation} is defined to measure the performance on the frames with the axis annotations present in the training set.
    
The challenge uses the mean joint error (MJE)~\cite{forth} as the main evaluation metric. Results are ranked according to the \textbf{Extrapolation} criterion which measures the total extrapolation power of the approaches with MJE on all axes. We also consider success rates based on maximum allowed distance errors for each frame and each joint for further analysis. 

Fig.~\ref{fig:evalstrategy}~(left) summarises the six evaluation strategies, and Fig.~\ref{fig:evalstrategy}~(right) shows the accuracies obtained by the best approaches, measured for the three  evaluation criteria that could be evaluated for all three tasks. Articulation and viewpoint criteria are only considered for Task 1 since the joint angles are mostly fixed during object interaction and hence the Articulation criteria is not as meaningful as in Task 1 for the other tasks. The Viewpoint criteria is not meaningful for Task 2 which is for egocentric views since the task's dataset constrains the relative palm-camera angle to a small range. For Task 3, the data scarcity is not helping to sample enough diverse viewpoints. The extrapolation errors tend to be three times larger than the interpolation errors while the shape is a bottleneck among the other attributes. Lower errors on Task 3 compared to Task 2 are likely due to the fact that the ground truth wrist position is provided for Task 3.

%% file: input_Dataset_details.tex
\section{Datasets}
\label{sec:datasets}
\begin{figure*}[!t]
{\includegraphics[width=0.245\linewidth]{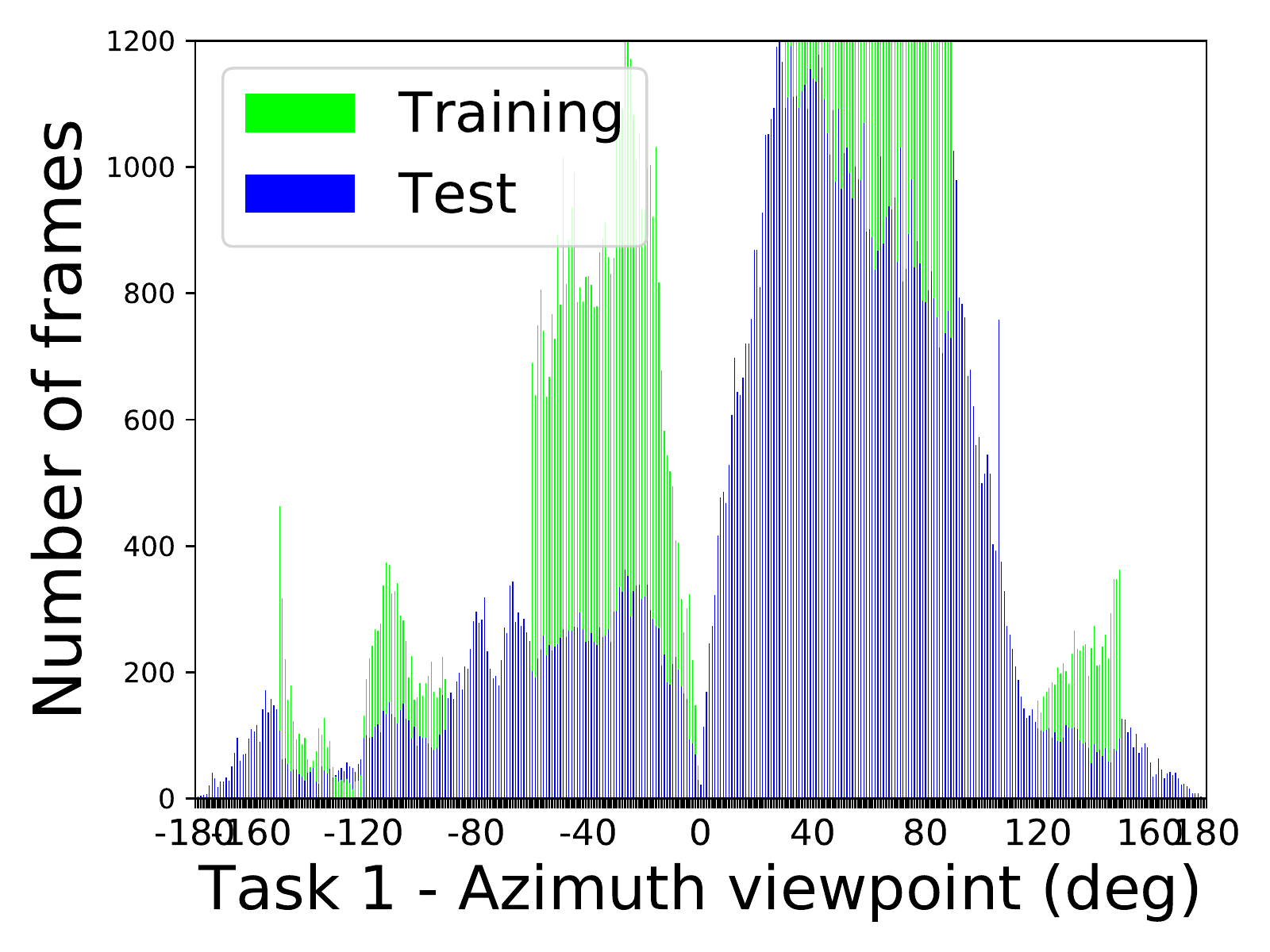}}
{\includegraphics[width=0.245\linewidth]{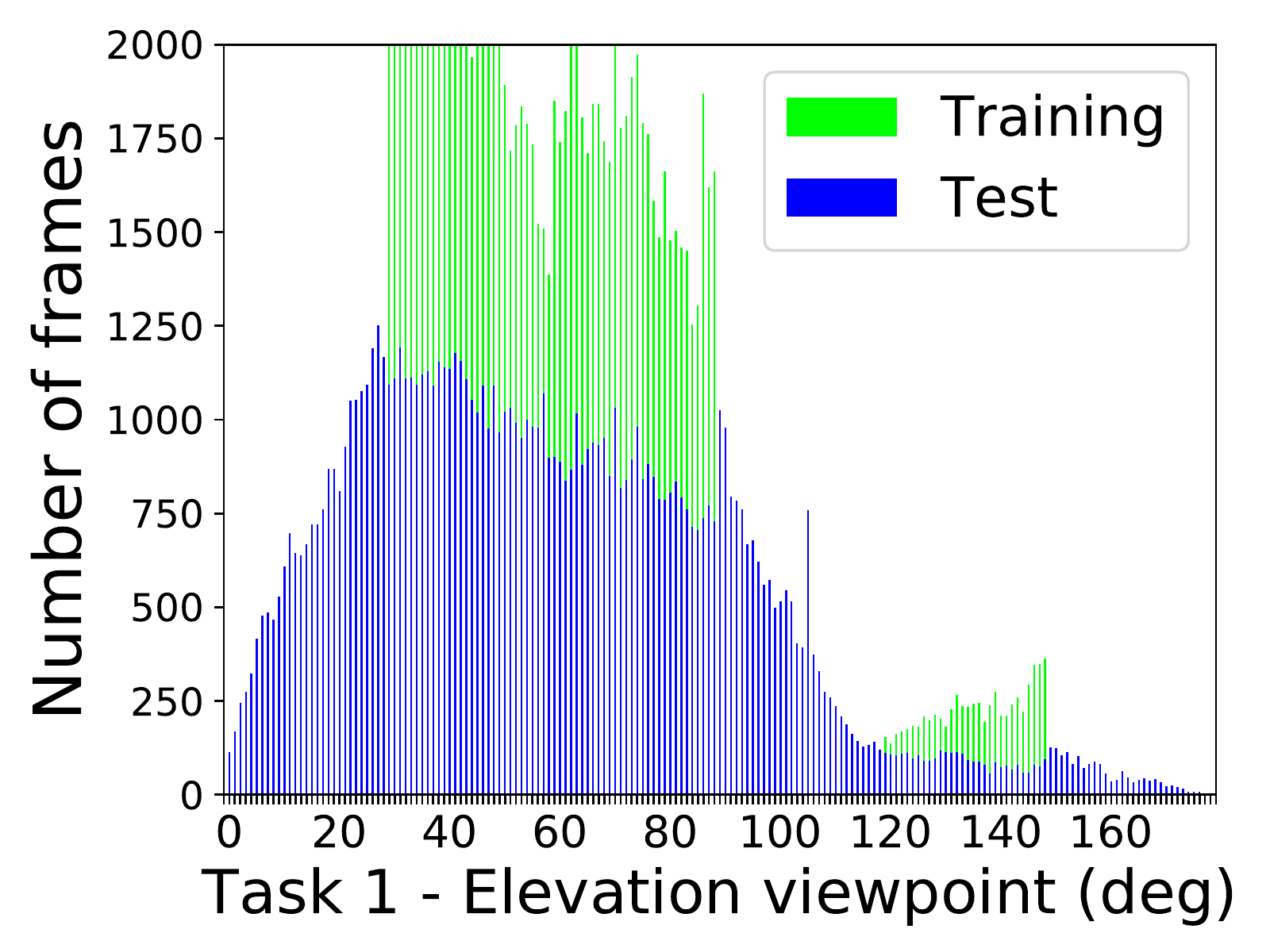}}
{\includegraphics[width=0.245\linewidth]{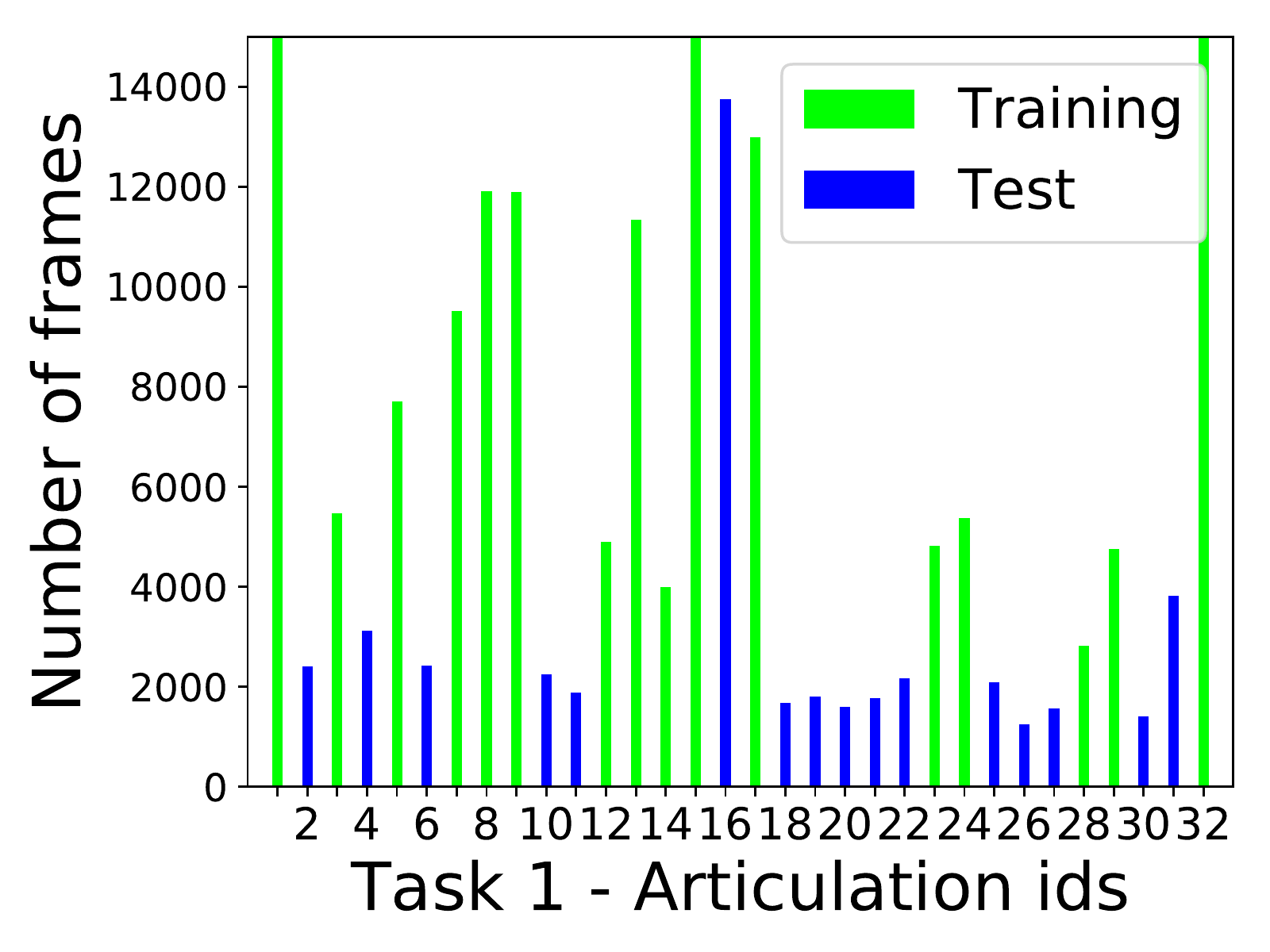}}
{\includegraphics[width=0.245\linewidth]{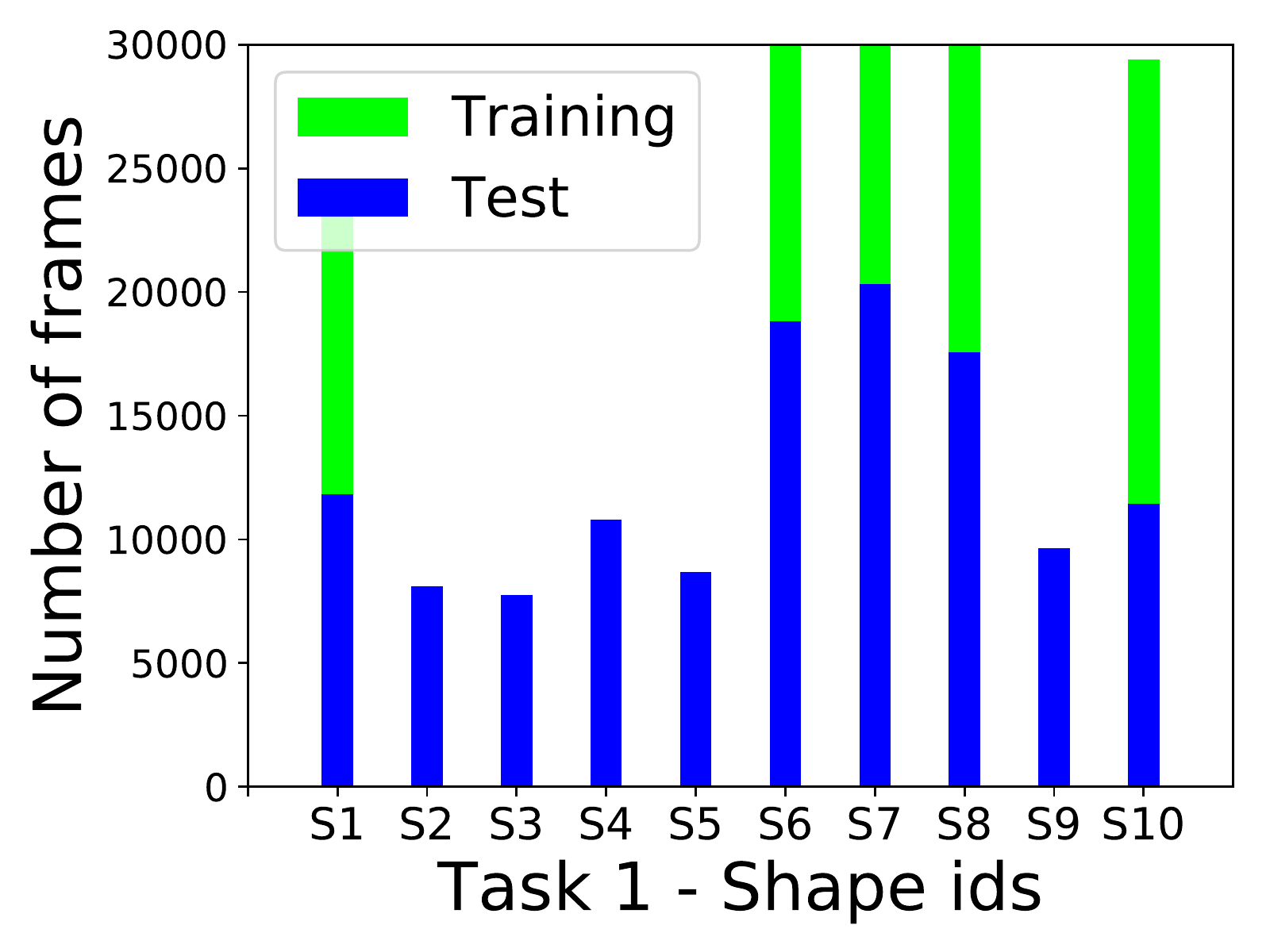}}
\\
{\includegraphics[width=0.245\linewidth]{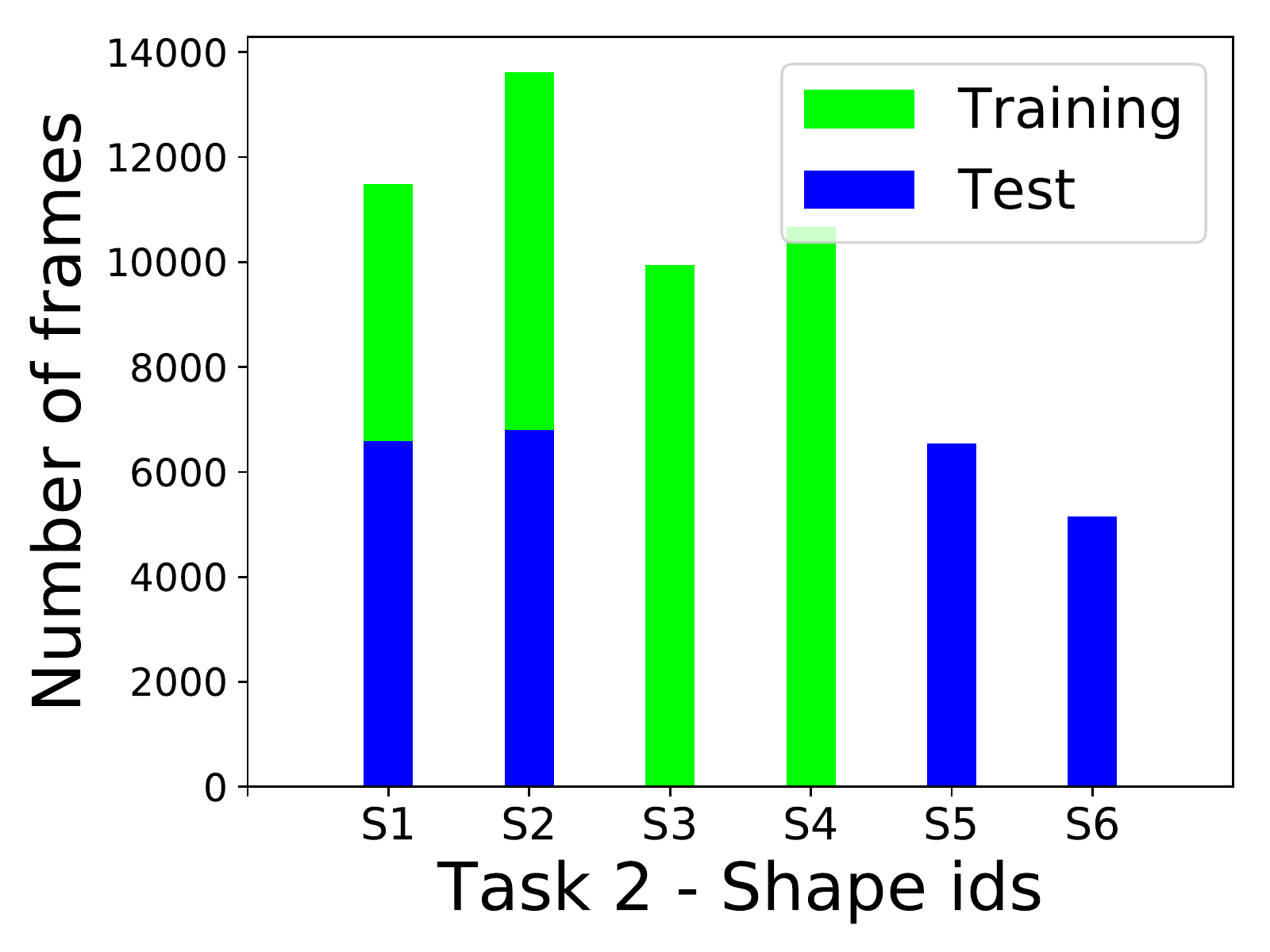}}
{\includegraphics[width=0.245\linewidth]{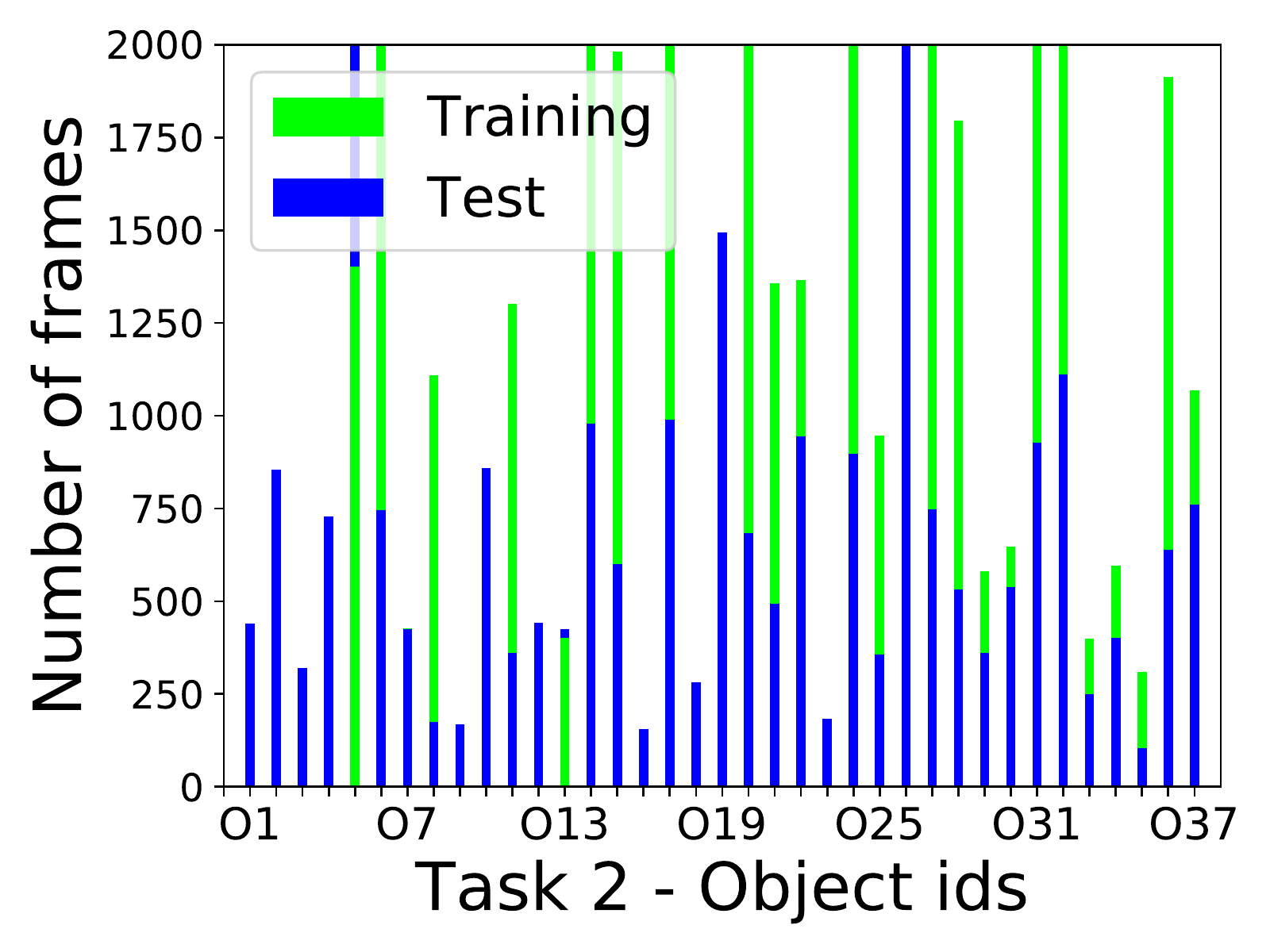}}
{\includegraphics[width=0.245\linewidth]{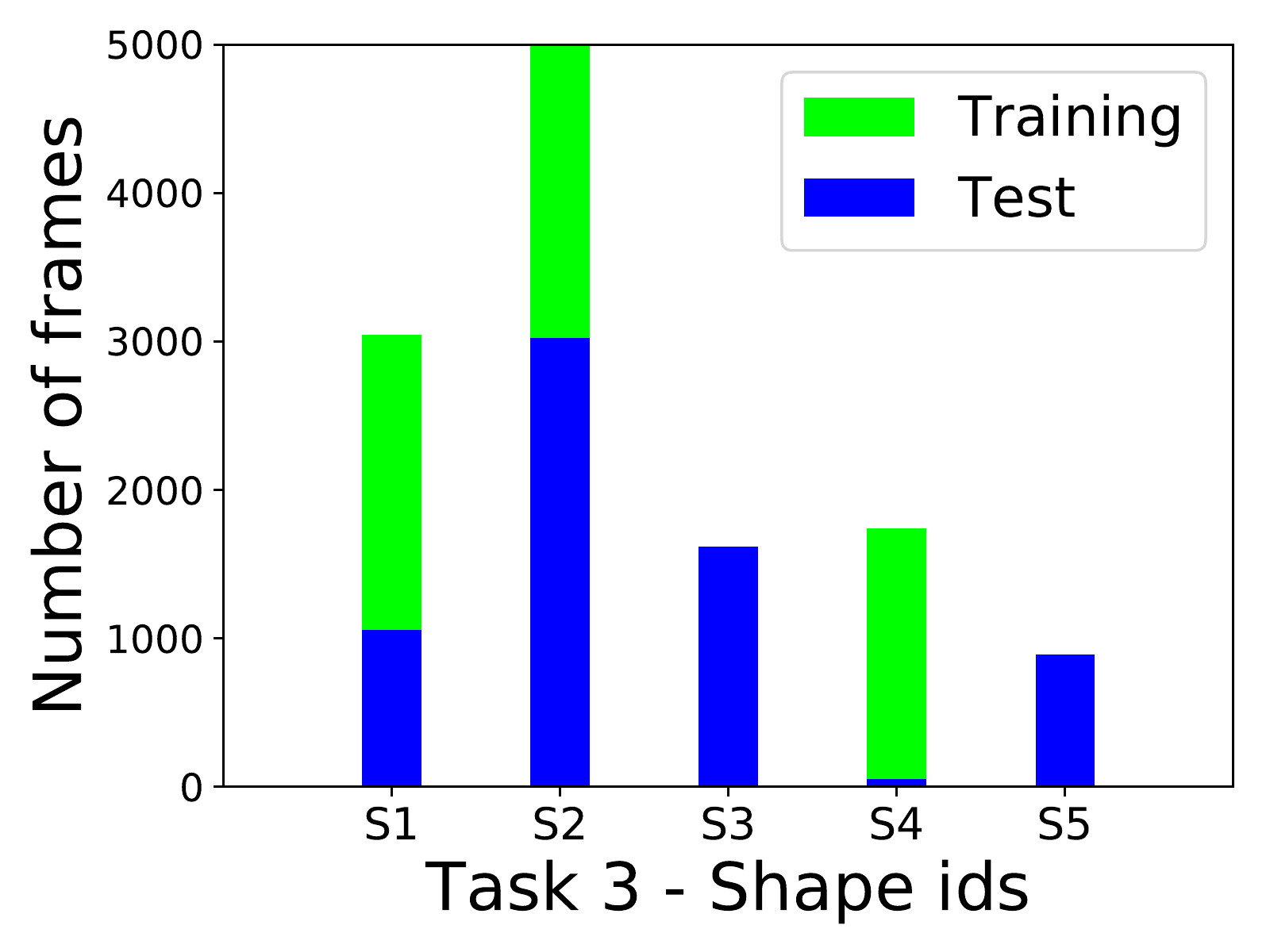}}
{\includegraphics[width=0.245\linewidth]{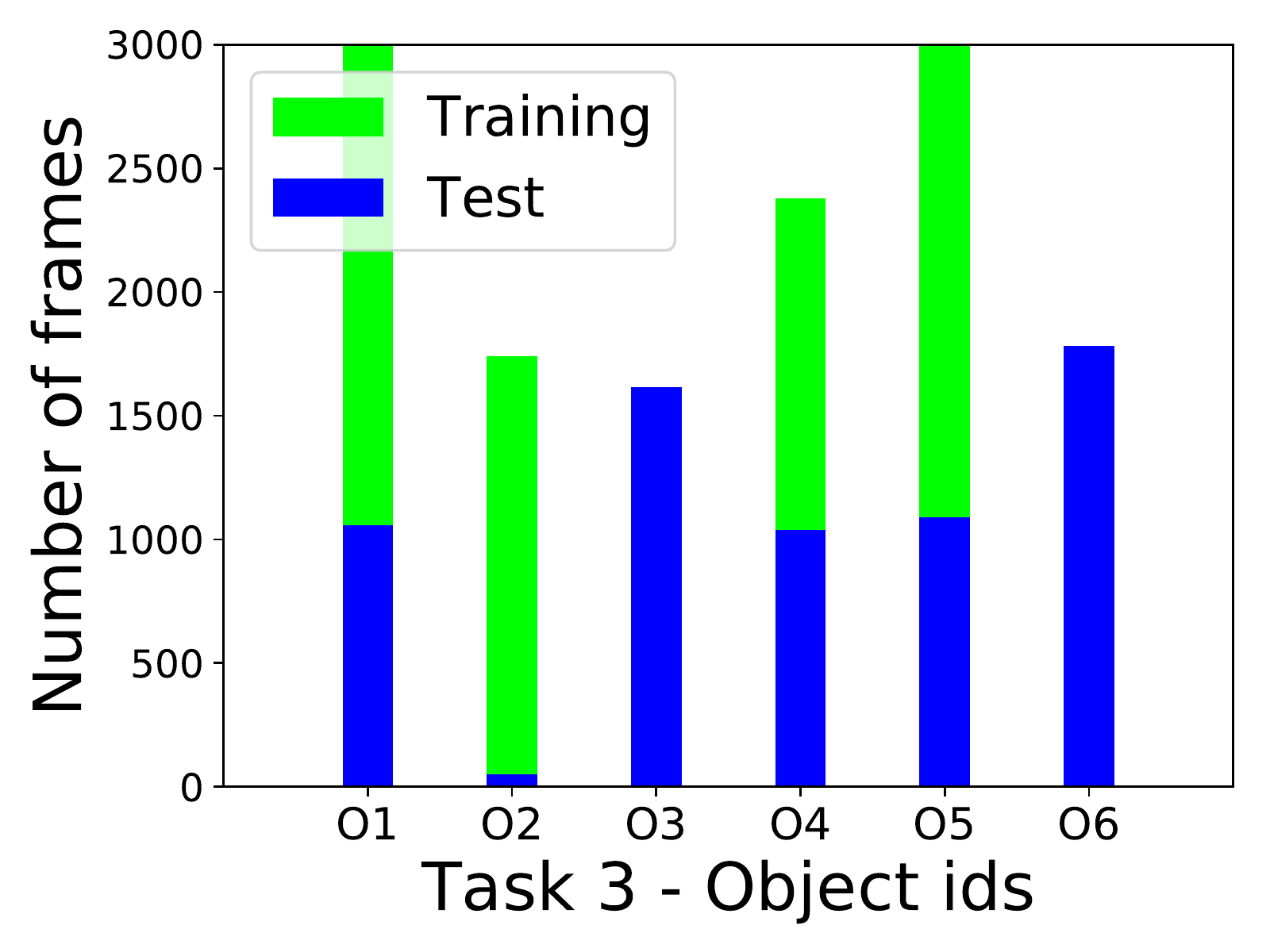}}
\\
\caption{Distributions of the training and test datasets for Task 1 (top), Task 2 (bottom left), and Task 3 (bottom right). The splits are used to evaluate the extrapolation power of the approaches and decided based on the viewpoints, the articulation clusters of the hand pose, the hand shape, and the type of the object present.
}
\label{fig:dataset_dist}
\end{figure*}

Given a task, the training set is the same and the test frames used to evaluate each criterion can be different or overlapped. The number of training frames are 175K, 45K and 10K for Task 1, 2 and 3 respectively. The sizes of the test sets for each evaluation criterion are shown in Table~\ref{tab:dataset_details}.

\newlength{\oldintextsep}
\setlength{\oldintextsep}{\intextsep}
\setlength\intextsep{0pt}
\begin{table}
\centering
\caption{\small Detailed analytics on the number of frames provided on the training and test sets for the different tasks.}
\resizebox{.95\textwidth}{!}{
\begin{tabular}{@{}ccccccccccccc@{}}
\multicolumn{1}{l}{}                           & \multicolumn{1}{l}{}   & \multicolumn{7}{c}{\#Frames}                                                                                            & \multicolumn{1}{l}{} & \multicolumn{1}{l}{}  & \multicolumn{1}{l}{}  & \multicolumn{1}{l}{}  \\ \cmidrule(lr){3-9}
Dataset                                        & Task id                & Total & Ext. & Int. & Art.                  & View.                 & Sha. & Obj.                                       & \#Subjects           & \#Objects             & \#Actions             & \#Seq.                \\ \midrule
\multicolumn{1}{c|}{\multirow{3}{*}{Test}}     & \multicolumn{1}{c|}{1} & 125K  & 20\% & 16\% & 16\%                  & 32\%                    & 16\% & \multicolumn{1}{c|}{\xmark} & 10                   & \xmark & \xmark & \xmark \\
\multicolumn{1}{c|}{}                          & \multicolumn{1}{c|}{2} & 25K   & 14\% & 32\% & \xmark & \xmark & 37\% & \multicolumn{1}{c|}{17\%}                  & 4                    & 37                    & 71                    & 292                   \\
\multicolumn{1}{c|}{}                          & \multicolumn{1}{c|}{3} & 6.6K  & 24\% & 35\% & \xmark & \xmark & 14   & \multicolumn{1}{c|}{27\%}                  & 5                    & 5                     & 1                     & 5                     \\ \midrule
\multicolumn{1}{c|}{\multirow{3}{*}{Training}} & \multicolumn{1}{c|}{1} & \multicolumn{7}{c|}{175951}                                                                                              & 5                    & \xmark & \xmark & \xmark \\
\multicolumn{1}{c|}{}                          & \multicolumn{1}{c|}{2} & \multicolumn{7}{c|}{45713}                                                                                               & 4                    & 26                    & 45                    & 539                   \\
\multicolumn{1}{c|}{}                          & \multicolumn{1}{c|}{3} & \multicolumn{7}{c|}{10505}                                                                                               & 3                    & 4                     & 1                     & 12                    \\ \bottomrule
\end{tabular}
}
\label{tab:dataset_details}
\end{table}

Fig.~\ref{fig:dataset_dist} shows the distributions of the training and test data for each task. The viewpoints are defined as elevation and azimuth angles of the hand \wrt the camera using the ground-truth joint annotations. The articulation of the hand is defined and obtained by clustering on the ground-truth joint angles in a fashion similar to~\cite{lin2000modeling}, by using binary representations (open/closed) of each finger \eg~`00010' represents a hand articulation cluster with frames with the index finger closed and the rest of the fingers open, which ends up with $2^5=32$ clusters. Examples from the articulation clusters are provided in Appendix~\ref{appendix:visual}. Note that the use of a low-dimensional embedding such as PCA or t-SNE is not adequate here to compare the two data distributions, because the dimensionality of the distributions is very high and a low-dimensional embedding would not be very representative. Fig.~\ref{fig:dataset_dist} further shows the splits in terms of subjects/shapes, where five seen subjects and five unseen subjects are present. Similarly, the data partition was done on objects. This way we can control the data to define the evaluation metrics. 

\noindent\textbf{Use of 3D Hand Models for HPEs.}
A series of methods~\cite{cvpr19_handmesh,cvpr19_handmesh2,cvpr19_handmesh3,cvpr19_handmesh4,iccv19_handmesh} have been proposed in the literature to make use of 3D hand models for supervision of HPEs. Ge~\etal~\cite{cvpr19_handmesh3} proposed to use Graph CNNs for mapping RGB images to infer the vertices of 3D meshes. Hasson~\etal~\cite{cvpr19_handmesh4} jointly infers both hands and object meshes and investigated the effect of the 3D contact loss penalizing the penetration of object and hand surfaces.
Others~\cite{cvpr19_handmesh,cvpr19_handmesh2,iccv19_handmesh} attempted to make use of MANO~\cite{mano}, a parametric 3D hand model by learning to estimate low-dimensional PCA parameters of the model and using it together with differentiable model renderers for 3D supervision. All the previous works on the use of 3D models in learning frameworks have shown to help improving performance on the given task. Recently, \cite{black_iccv19_inaloop} showed that fitting a 3D body model during the estimation process can be accelerated by using better initialization of the model parameters however, our goal is slightly different since we aim to explore the use of 3D models for better generalisation from the methods. Since the hand pose space is huge, we make use of a 3D hand model to fill the gaps in the training data distribution to help approaches to improve their extrapolation capabilities. In this study, we make use of the MANO~\cite{mano} hand model by providing the model's parameters for each training image. We fit the 3D model for each image in an optimization-based framework which is described in more details below.

\begingroup
\setlength{\abovedisplayskip}{4pt}
\setlength{\belowdisplayskip}{4pt}
\setlength{\abovedisplayshortskip}{4pt}
\setlength{\belowdisplayshortskip}{4pt}
\noindent\textbf{Gradient-based Optimization for Model Fitting.}
We fit the MANO~\cite{mano} models' shape $\mathbf{s}=\{s_j\}_{j=1}^{10}$, camera pose $\mathbf{c}=\{c_j\}_{j=1}^{8}$, and articulation $\mathbf{a}=\{a_j\}_{j=1}^{45}$ parameters to the $i$-th raw skeletons of selected articulations $\mathbf{z}=\{z_i\}_{i=1}^{K}$, by solving the following equation:
\begin{eqnarray}
\label{eq:obtainrealpose}
(\mathbf{s}^{i*},\mathbf{c}^{i*},\mathbf{a}^{i*}) = \arg\min_{(\mathbf{s},\mathbf{c},\mathbf{a})} O(\mathbf{s}, \mathbf{c}, \mathbf{a}, \mathbf{z}^i)), \forall i \in [1,K] \> ,
\end{eqnarray}
where our proposed objective function $O(\mathbf{s}, \mathbf{c}, \mathbf{a}, \mathbf{z}^i)$ for the sample $i$ is defined as follows:
\begin{eqnarray}
\label{eq:objective}
O(\mathbf{s}, \mathbf{c}, \mathbf{a}, \mathbf{z}^i)=||f^{reg}(V(\mathbf{s},\mathbf{c},\mathbf{a})) - \mathbf{z}^i||^2_2 + \sum_{j=1}^{10} \|\mathbf{s}_j\|^2_2 + R_{Lap}(V(\mathbf{s},\mathbf{c},\mathbf{a})) \> .
\end{eqnarray}
$V(\mathbf{s},\mathbf{c},\mathbf{a})$ denotes the 3D mesh as a function of the three parameters $\mathbf{s},\mathbf{c},\mathbf{a}$. Eq.~\eqref{eq:objective} is composed of the following terms: $i)$ the Euclidean distance between 3D skeleton ground-truths $\mathbf{z}^i$ and the current MANO mesh model's 3D skeleton values $f^{reg}(V(\mathbf{s},\mathbf{c},\mathbf{a}))$\footnote{$f^{reg}$ geometrically regresses the skeleton from the mesh vertex coordinates. It is provided with the MANO model and the weights are fixed during the process.};
$ii)$ A shape regularizer enforcing the shape parameters $\mathbf{s}$ to be close to their MANO model's mean values, normalized to $0$ as in~\cite{mano}, to maximize the shape likelihood; and $iii)$ A Laplacian regularizer $R_{Lap}(V(\mathbf{s},\mathbf{c},\mathbf{a}))$ to obtain the smooth mesh surfaces as in~\cite{eccv_2018_mesh}. Eq.~\eqref{eq:obtainrealpose} is solved iteratively by using the gradients from Eq.~\eqref{eq:objective} as follows:
\begin{align}
(\mathbf{s}_{t+1}, \mathbf{c}_{t+1}, \mathbf{a}_{t+1}) &= (\mathbf{s}_{t}, \mathbf{c}_{t}, \mathbf{a}_{t})-\gamma\cdot\nabla O(\mathbf{s}_t, \mathbf{c}_t, \mathbf{a}_t, \mathbf{z}^i), \forall t\in [1,T] \> ,
\label{eq:refinesvp}
\end{align}
where $\gamma=10^{-3}$ and $T=3000$ are empirically set. This process is similar to the refinement step of~\cite{nips_2017_motioncapture,cvpr19_handmesh}, which refines estimated meshes by using the gradients from the loss.
\endgroup
In Fig.~\ref{fig:fitting}, both the target and the fitted depth images during the process described by Eq.~\eqref{eq:refinesvp} are depicted. Minor errors of the fitting are not a problem for our purpose given that we will generate input and output pairs of the fitted model by exploiting fitted meshes' self-data generation capability while ignoring original depth and skeletons. Here the aim of fitting the hand model is to obtain a plausible and a complete articulation space. The model is fitted without optimizing over depth information from the model and the input depth image since we did not observe an improvement on the parameter estimation. Moreover, the optimization needs to be constrained to produce plausible hand shapes and noise and other inconsistencies may appear in the depth image.

\begingroup
\setlength{\tabcolsep}{0pt}
\begin{figure}[t!]
\centering
\setlength{\tabcolsep}{0pt}
\begin{tabular}{c c c}
{\includegraphics[width=0.125\linewidth]{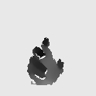}} &
{\includegraphics[width=0.125\linewidth]{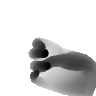}} 
{\includegraphics[width=0.125\linewidth]{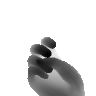}} 
{\includegraphics[width=0.125\linewidth]{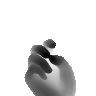}} 
{\includegraphics[width=0.125\linewidth]{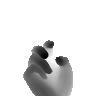}} 
{\includegraphics[width=0.125\linewidth]{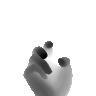}} 
{\includegraphics[width=0.125\linewidth]{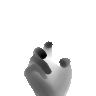}} &
{\includegraphics[width=0.125\linewidth]{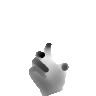}}
\\
{\includegraphics[width=0.125\linewidth]{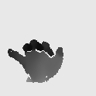}} &
{\includegraphics[width=0.125\linewidth]{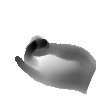}} 
{\includegraphics[width=0.125\linewidth]{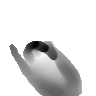}} 
{\includegraphics[width=0.125\linewidth]{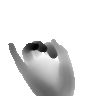}} 
{\includegraphics[width=0.125\linewidth]{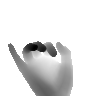}} 
{\includegraphics[width=0.125\linewidth]{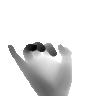}} 
{\includegraphics[width=0.125\linewidth]{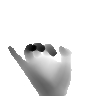}} &
{\includegraphics[width=0.125\linewidth]{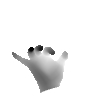}}
\\
{\smaller{(a)}} & {\smaller{(b)}} & {\smaller{(c)}}
\end{tabular}
\caption{\small Depth renderings of the hand model for different iterations in gradient-based optimization fitting. Target image (joints) (a), optimization iterations 0, 100, 300, 400, 600, 700 (b), final fitted hand pose at iteration 3000 (c).}
\label{fig:fitting}
\vspace{-20pt}
\end{figure}
\endgroup

%% file: input_Evaluated_methods.tex
\section{Evaluated Methods}
\label{sec:evaluated_methods}
In this section, we present the gist of  selected 14 methods among 36 participants (17 for Task 1, 10 for Task 2, 9 for Task 3) to further analyze their results in Section~\ref{sec:analysis}. Methods are categorized based on their main components and properties. See Tables~\ref{tab:task1_methods},\ref{tab:task2_methods} and \ref{tab:task3_methods} for a glance of the properties of the methods in~\hands.

\noindent\textbf{2D and 3D supervision for HPEs.} Approaches that embed and process 3D data obtain high accuracies but less efficient~\cite{shanxin_cvpr2018} in terms of their complexity compared to 2D-based approaches. 3D-based methods use 3D convolutional layers for point-clouds input similar to \taskonefour which uses an efficient voxel-based representation V2V-PoseNet~\cite{v2vposenet} with a deeper architecture and weighted sub-voxel predictions on quarter of each voxel representations for robustness. Some other approaches adopts 3D as a way of supervision similar to
\taskonefive~\cite{strawberyfg} which employs a render-and-compare stage to enforce voxel-wise supervision for model training and adopts a 3D skeleton volume renderer to re-parameterize an initial pose estimate obtained similar to~\cite{integralpose}. \taskonesix uses a permutation invariant feature extraction layer~\cite{respel} to extract point-cloud features and uses a two branch framework for point-to-pose voting and point-to-latent voting. 3D supervision is employed by point-cloud reconstruction from a latent embedding in Task 1 whereas 3D hand model parameters are estimated and used in a differentiable model renderer for 3D supervision for the other tasks. 

2D CNN-based approaches has been a standard way for learning regression models as used by \taskoneone~\cite{rokid_arxiv} where they adopt a two stage regression models. The first regression model is used to predict an initial pose and the second model built on top of the first model. \taskonetwo~\cite{a2j} uses a 2D supervised method based on 2D offset and depth estimations with anchor points. Anchor points are densely set on the input image to behave as local regressors for the joints and able to capture global-local spatial context information. \taskonethree~\cite{awr_aaai20} adopts a learnable and adaptive weighting operation that is used to aggregate spatial information of different regions in dense representations with 2D convolutional CNNs. The weighting operation adds direct supervision on joint coordinates and draw consensus between the training and inference as well as enhancing the model's accuracy and generalisation ability by adaptively aggregating spatial information from related regions. \tasktwothree uses a hierarchically structured regression network by following the joints' distribution on the hand morphology. \taskthreeone~\cite{ethnvidia_arxiv} adopts the latent 2.5D heatmap regression~\cite{iqbal_25d_eccv18}; additionally an MLP is adopted for denoising the absolute root depth. Absolute 3D pose in scale-normalized space is obtained with the pinhole camera equations. \taskthreetwo~\cite{lcrnetpp} first performs a classification of the hand into a set of canonical hand poses (obtained by clustering on the poses in the training set), followed by a fine class-specific regression of the hand joints in 2D and 3D. \taskthreetwo~adopts the only approach proposing multiple hand poses in a single stage with a Region Proposal Network~(RPN)~\cite{fasterrcnn} integration.

\begingroup
\setlength{\tabcolsep}{1pt} 
\def\arraystretch{1.1}
\begin{table*}[!t]
\centering
\caption{Task 1 - Methods' Overview}
\label{tab:task1_methods}
\resizebox{\textwidth}{!}{
\begin{tabular}{c c c c c c c c c}
\toprule
 Username & Description & Input & Pre-processing & Post-processing & Synthetic Data & Backbone & Loss & Optimizer\\
\hline

\taskoneone~\cite{rokid_arxiv} & 
\begin{tabular}{@{}c@{}}2D CNN\\joint regression\end{tabular}&
\begin{tabular}{@{}c@{}}Depth\\$224\times224$\end{tabular}&
\begin{tabular}{@{}c@{}}Initial pose\\est. to crop\end{tabular}&
\xmark&
\begin{tabular}{@{}c@{}}570K Synthetic\\+ Mixed Synthetic \end{tabular}&
\begin{tabular}{@{}c@{}}EfficientNet-b0~\cite{efficientnet} \end{tabular}&
\begin{tabular}{@{}c@{}}Wing~\cite{wingloss}\end{tabular}&
\begin{tabular}{@{}c@{}}Adamax\end{tabular}\\
\hline

\taskonetwo~\cite{a2j} & 
\begin{tabular}{@{}c@{}}2D CNN, offset + depth\\ regression with anchor\\points and weighting\end{tabular}&
\begin{tabular}{@{}c@{}}Depth\\$384\times384$ \end{tabular}&
\begin{tabular}{@{}c@{}}Bbox crop\end{tabular}&
\begin{tabular}{@{}c@{}}Scale+rotation,\\10 backbone\\models ensemble\end{tabular}&
\xmark&
\begin{tabular}{@{}c@{}}ResNet-152\end{tabular}&
\begin{tabular}{@{}c@{}}Smooth L1\end{tabular}&
\begin{tabular}{@{}c@{}}Adam \end{tabular}\\
\hline

\taskonethree~\cite{awr_aaai20} & 
\begin{tabular}{@{}c@{}}2D CNN, dense direction\\\& offset rep.\\Learnable adaptive weighting\end{tabular}&
\begin{tabular}{@{}c@{}}
Depth\\$256\times256$ segm.\\$128\times128$ pose est.\end{tabular}&
\begin{tabular}{@{}c@{}}
Bbox crop\\ESPNet-v2~\cite{espnet}\\for binary segm.\\iter. refinement of CoM\end{tabular}&
\begin{tabular}{@{}c@{}}Ensemble\\from 5 models\end{tabular}&
\xmark&
\begin{tabular}{@{}c@{}}
ResNet-50\&101\\SRN~\cite{srn}\\HRNet~\cite{hrnet}\end{tabular}&
\begin{tabular}{@{}c@{}}Smooth L1 \end{tabular}&
\begin{tabular}{@{}c@{}}Adam \end{tabular}\\
\hline

\taskonefour~\cite{v2vposenet} & 
\begin{tabular}{@{}c@{}}3D CNN\\Deeper V2V-PoseNet~\cite{v2vposenet}\\Weighted sub-voxel\\prediction\end{tabular}&
\begin{tabular}{@{}c@{}}Voxels\\$88\times88\times88$\end{tabular}&
\begin{tabular}{@{}c@{}}Multi-scale\\CoM refinement\\hand cropping\end{tabular}&
\begin{tabular}{@{}c@{}}Models from\\6 training epochs\\ N confident\\sub-voxel pred.\\Truncated SVD refinement\end{tabular}&
\xmark &
\begin{tabular}{@{}c@{}}V2V-PoseNet\end{tabular}&
\begin{tabular}{@{}c@{}}L2\end{tabular}&
\begin{tabular}{@{}c@{}}RMSProp\end{tabular}\\
\hline

\taskonefive~\cite{strawberyfg} & 
\begin{tabular}{@{}c@{}}Integral Pose\\Regression~\cite{integralpose}\\3D supervision\\voxels +volume rendering\end{tabular} & 
\begin{tabular}{@{}c@{}}Depth image $256\times256$\\ 3D point proj. \\ Multi-layer depth\\ Voxels\end{tabular}&
\begin{tabular}{@{}c@{}}Coarse-to-fine\\hand cropping\\by thresholding\end{tabular}&
\xmark&
\xmark&
ResNet-50&
L1&
RMSProb\\
\hline

\taskonesix~\cite{respel}& 
\begin{tabular}{@{}c@{}}3D supervision\\with cloud reconst.\\Permutation invariant~\cite{respel}\\Point-to-pose +\\point-to-latent voting.\end{tabular}&
\begin{tabular}{@{}c@{}}Point cloud\\ 512 3D vectors\end{tabular} &
\begin{tabular}{@{}c@{}}View correction~\cite{respel}\end{tabular}&
\xmark & \xmark &
\begin{tabular}{@{}c@{}}ResPel~\cite{respel}\\for feat. extract\\FoldingNet~\cite{foldingnet}\\for reconstruction\end{tabular}&
\begin{tabular}{@{}c@{}}L2 \\ Chamfer and EMD\\ KL constraint\end{tabular}&
Adam\\
\bottomrule
\end{tabular}
}
\end{table*}
\endgroup

\begingroup
\setlength{\tabcolsep}{1pt} 
\def\arraystretch{1.1}

\begin{table*}[!t]
\centering
\caption{Task 2 - Methods' Overview}
\label{tab:task2_methods}

\resizebox{\textwidth}{!}{%
\begin{tabular}{c c c c c c c c c}
\toprule
 Username & Description & Input & Pre-processing & Post-processing & Synthetic Data & Backbone & Loss & Optimizer\\
\hline

\tasktwoone~\cite{v2vposenet}& 
\begin{tabular}{@{}c@{}}3D CNN\\Deeper V2V-PoseNet~\cite{v2vposenet}\\Weighted sub-voxel\\prediction\end{tabular}&
\begin{tabular}{@{}c@{}}Voxels\\$88\times88\times88$\end{tabular}&
\begin{tabular}{@{}c@{}}Multi-scale\\com-ref-net\\for hand cropping\end{tabular}&
\begin{tabular}{@{}c@{}}Models from\\6 training epochs\\N sub-voxel pred.,\\Truncated SVD and\\temporal smoothing refinement\end{tabular}&
\xmark&
\begin{tabular}{@{}c@{}}V2V-PoseNet\end{tabular}&
\begin{tabular}{@{}c@{}}L2\end{tabular}&
\begin{tabular}{@{}c@{}}RMSProp\end{tabular}\\
\hline

\tasktwotwo~\cite{a2j} & 
\begin{tabular}{@{}c@{}}2D CNN offset and\\depth regression\\with anchor points\\and weighting\end{tabular}&
\begin{tabular}{@{}c@{}}Depth\\$256\times256$ \end{tabular}&
\begin{tabular}{@{}c@{}}Bbox crop \end{tabular}&
\begin{tabular}{@{}c@{}}Ensemble predictions\\from 3\\ training epochs\end{tabular}&
\xmark&
\begin{tabular}{@{}c@{}}SEResNet-101~\cite{senet}\end{tabular}&
\begin{tabular}{@{}c@{}}Smooth L1\end{tabular}&
\begin{tabular}{@{}c@{}}Adam\end{tabular}\\
\hline

\tasktwothree & 
\begin{tabular}{@{}c@{}}2D CNN\\tree-like branch\\structure regression\\with hand morphology\end{tabular}&
\begin{tabular}{@{}c@{}}Depth\\$128\times128$ \end{tabular}&
\begin{tabular}{@{}c@{}}Iterative CoM\end{tabular}&
\xmark&
\xmark&
ResNet-50&
L2&
-\\
\hline

\tasktwofour~\cite{respel} & 
\begin{tabular}{@{}c@{}}Differentiable\\Mano~\cite{mano} layer\\Permutation invariant~\cite{respel}\\Point-to-pose+\\point-to-latent voting\end{tabular}&
\begin{tabular}{@{}c@{}}Point cloud\\512 3D points \end{tabular}&
\begin{tabular}{@{}c@{}}View correction~\cite{respel}\end{tabular}&
\xmark&
\begin{tabular}{@{}c@{}}32K synthetic\\+ random objects\\from HO-3D~\cite{ho3d_2019}\end{tabular}&
ResPel~\cite{respel}&
\begin{tabular}{@{}c@{}}L2 pose\\L2 Mano vertex\\KL constraint\end{tabular}&
Adam\\
\bottomrule
\end{tabular}
}
\end{table*}
\endgroup

\begingroup
\setlength{\tabcolsep}{1pt} 
\def\arraystretch{1.1}
\begin{table*}[!h]
\centering
\caption{Task 3 - Methods' Overview}
\label{tab:task3_methods}
\resizebox{\textwidth}{!}{
\begin{tabular}{c c c c c c c c c}
\toprule
 Username & Description & Input & Pre-processing & Post-processing & Synthetic Data & Backbone & Loss & Optimizer\\
\hline

\begin{tabular}{@{}c@{}}\taskthreeone\\\cite{ethnvidia_arxiv}\end{tabular}& 
\begin{tabular}{@{}c@{}}2D CNN, 2D location +\\relative depth\\Heatmap-regression + an MLP for\\denoising absolute root depth\end{tabular}&
\begin{tabular}{@{}c@{}}RGB\\ $128\times128$ \end{tabular}&
\begin{tabular}{@{}c@{}}Bbox crop \end{tabular}&
\xmark&
\xmark&
\begin{tabular}{@{}c@{}}ResNet-50\end{tabular}&
L1&
\begin{tabular}{@{}c@{}}SGD \end{tabular}\\
\hline

\taskthreetwo~\cite{lcrnetpp} & 
\begin{tabular}{@{}c@{}}2D hand proposals + classification of\\multiple anchor poses + regression of\\2D-3D keypoint offsets \wrt the anchors\end{tabular}&
\begin{tabular}{@{}c@{}}RGB\\$640\times480$\end{tabular}&
\xmark&
\begin{tabular}{@{}c@{}}Ensemble multiple\\pose proposals and\\ensemble over\\rotated images\end{tabular}&
\xmark&
\begin{tabular}{@{}c@{}}ResNet-101 \end{tabular}&
\begin{tabular}{@{}c@{}}Smooth L1 for reg.\\Log loss for classif.\\RPN~\cite{fasterrcnn} for\\localization loss\end{tabular}&
\begin{tabular}{@{}c@{}}SGD\end{tabular}\\
\hline

\taskthreethree~\cite{yang2019aligning} &
\begin{tabular}{@{}c@{}}Multi-modal input\\with latent space\\alignment~\cite{yang2019aligning}\\ Differentiable Mano~\cite{mano} layer\end{tabular}&
\begin{tabular}{@{}c@{}}RGB $256\times256$\\Point cloud - 356 \end{tabular}&
\begin{tabular}{@{}c@{}}Bbox cropping \end{tabular}&
\xmark&
\begin{tabular}{@{}c@{}}100K synthetic +\\random objects\\ from HO-3D~\cite{ho3d_2019}\end{tabular}&
\begin{tabular}{@{}c@{}}EncoderCloud: ResPEL~\cite{respel}\\EncoderRGB: ResNet-18\\DecoderMano: 6 fully-connected \end{tabular}&
\begin{tabular}{@{}c@{}}L2 pose, L2 Mano vert.\\Chamfer, Normal and\\Edge length for mesh\\KL constraint\end{tabular}&
\begin{tabular}{@{}c@{}}Adam\end{tabular}\\
\bottomrule
\end{tabular}
}
\end{table*}
\endgroup

\noindent\textbf{Detection, regression and combined HPEs.} Detection methods are based on hand key-points and producing a probability density maps for each joint. \tasktwoone uses a 3D CNN~\cite{v2vposenet} to estimate per-voxel likelihood of each joint. Regression-based methods estimate the joint locations by learning a direct mapping from the input image to hand joint locations or the joint angles of a hand model~\cite{hand_cvpr_2016_2,hand_ijcai_2016}. \taskoneone uses joint regression models within two stages to estimate an initial hand pose for hand cropping and estimates the final pose from the cleaned hand image. \taskonetwo adopts regression framework by regressing offsets from anchors to final joint location. \taskonesix's point-wise features are used in a voting scheme which behaves as a regressor to estimate the pose. 

Some approaches take advantage of both detection-based and regression-based methods. Similarly, \taskonethree, \taskonefive estimates probability maps to estimate joint locations with a differentiable \textit{soft-argmax} operation~\cite{integralpose}. A hierarchical approach proposed by \tasktwothree regresses the joint locations from joint probability maps. \taskthreeone estimates 2D joint locations from estimated probability maps and regresses relative depth distance of the hand joints \wrt a root joint. \taskthreetwo first localizes the hands and classifies them to anchor poses and the final pose is regressed from the anchors.

\noindent\textbf{Method-wise ensembles.} 
\taskonetwo uses densely set anchor points in a voting stage which helps to predict location of the joints in an ensemble way for better generalisation leveraging the uncertainty in reference point detection. In a similar essence, \taskonethree adaptively aggregates the predictions from different regions and \taskonefive adopts local patch refinement~\cite{patch_hpe} where refinement models are adopted to refine bone orientations. \taskonesix uses the permutation equivariant features extracted from the point-cloud in a point-to-pose voting scheme where the votes are ensembled to estimate the pose. \taskthreetwo ensembles anchor poses to estimate the final pose.

\noindent\textbf{Ensembles in post-processing.} 
Rather than a single pose estimator, an ensemble approach was adopted by multiple entries by randomly replicating the methods and fusing the predictions in the post-prediction stage, \eg~\taskonetwo, \taskonethree, \taskonefour, \taskthreetwo and \taskonefive.

\taskonetwo ensembles predictions from ten different backbone architectures in Task 1 like \taskonethree (5 backbones) and augments test images to ensemble the predictions with different scales and rotations as similar to rotation augmentation adopted by \taskthreetwo. \taskonefour uses predictions obtained from the same model at 6 different training epochs. A similar ensembling is also adopted by \tasktwotwo in Task 2. \tasktwoone adopts a different strategy where $N$ most confident sub-voxel predictions are ensembled to further use them in a refinement stage with Truncated SVDs together with temporal smoothing (Task 2). \taskthreetwo takes advantage of ensembles from multiple pose proposals~\cite{lcrnetpp}. \taskonefive employs a different strategy and ensembles the predictions from models that are trained with various input modalities.

\setlength{\oldintextsep}{\intextsep}
\setlength\intextsep{0pt}
\begin{wrapfigure}{r}{0.45\textwidth}
\centering
\begingroup
\setlength{\tabcolsep}{1pt} 
\renewcommand{\arraystretch}{0.1} 
\begin{tabular}{c c c}
{\includegraphics[width=0.3\linewidth]{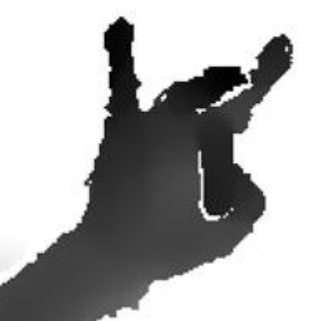}}  &
{\includegraphics[width=0.3\linewidth]{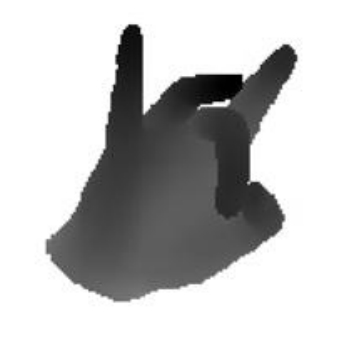}}  &
{\includegraphics[width=0.3\linewidth]{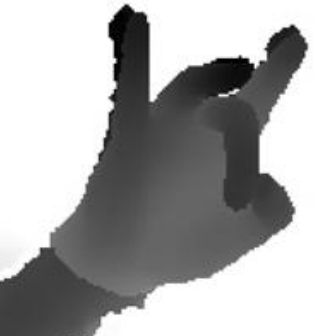}}
\\
{\footnotesize{\begin{tabular}{@{}c@{}}(a) Real\\Cropped\\Hand\end{tabular}}} &
{\footnotesize{\begin{tabular}{@{}c@{}}(b) Synthetic\\Depth\\Rendering\end{tabular}}} &
{\footnotesize{\begin{tabular}{@{}c@{}}(c) Real +\\Synthetic\\Mixed Hand\end{tabular}}} 
\end{tabular}
\endgroup
\caption{{Visualization of synthetic depth images by \taskoneone~\cite{rokid_arxiv}: (a) input depth image, (b) rendered depth image using 3D hand model, (c) the mixed by using the pixels with the closest depth values from real and synthetic images.}
}
\label{fig:zzzh_mixed_depth}
\end{wrapfigure} 

\noindent\textbf{Real + synthetic data usage.}
The methods \taskoneone in Task 1 and \tasktwofour in Tasks 2 and 3 make use of the provided MANO~\cite{mano} model parameters to synthesize more training samples. \taskoneone leverages the synthesized images and combines them the real images---see Fig.~\ref{fig:zzzh_mixed_depth}---to train their initial pose regression network which effectively boosts accuracies---see Table~\ref{tab:zzzh_mano_ratio}.
However, the amount of synthetic data created is limited to 570K for \taskoneone and 32K in Task 2, 100K in Task 3 for \tasktwofour. Considering the continuous high-dimensional hand pose space with or without objects, if we sub-sample uniformly and at minimum, for instance, $10^2$(azimuth/elevation angles)$\times2^5$(articulation)$\times10^1$(shape)$\times10^1$(object) = 320K, the number is already very large, causing a huge compromise issue for memory and training GPU hours. Random sampling was applied without a prior on the data distribution or smart sampling techniques~\cite{cubuk2019autoaugment,bhattarai2019sampling}.
\taskthreethree generates synthetic images with objects and hands similar to~\cite{GANeratedHands_CVPR2018} by randomly placing the objects from~\cite{ho3d_2019} to nearby hand locations without taking into account the hand and object interaction. The rest of the methods use the provided real training data only.
\begin{table}[!b]
\begingroup
\setlength{\tabcolsep}{1.5pt} 
\begin{minipage}{0.55\textwidth}
\centering
    \caption{Task 1 - MJE (mm) and ranking of the methods on five evaluation criteria. Best results on each evaluation criteria are highlighted.}
    \label{tab:task1_results}
    \resizebox{1.\textwidth}{!}{
    \begin{tabular}{c c c c c c}
    \toprule
    Username & Extrapolation &	Interpolation & Shape & Articulation & Viewpoint\\
    \hline
    \taskoneone & \textbf{13.66} (1) & 4.10 (2) & \textbf{10.27} (1) & 4.74 (3)	& \textbf{7.44} (1)\\
    \taskonetwo & 13.74 (2)	& 6.33 (6)	& 11.23 (4)	& 6.05 (6)	& 8.78 (6)\\
    \taskonethree & 13.76 (4)	& \textbf{3.93} (1) &	11.75 (5)	& \textbf{3.65} (1) &	7.50 (2)\\
    \taskonefour & 15.57 (7) &	4.54 (3) &	12.05 (6)	& 4.21 (2)	& 8.47 (4)\\
    \taskonefive	& 19.63 (12)	& 8.42 (10) &	14.21 (10)	& 7.50 (9)	& 14.16 (12)\\
    \taskonesix & 23.62 (14) &	18.78 (16)	& 21.84 (16)	& 16.73 (16) &	19.48 (14)\\
    \bottomrule
    \end{tabular}
}
\end{minipage}
\endgroup
\begingroup
\setlength{\tabcolsep}{1pt} 
\begin{minipage}{0.43\linewidth}
    \caption{Task 2 - MJE (mm) and ranking of the methods on four evaluation criteria.}
    \label{tab:task2_results}
    \resizebox{1.\linewidth}{!}{
    \begin{tabular}{c c c c c}
        \toprule
        Username & Extrapolation &	Interpolation & Object & Shape\\
        \hline
        \tasktwoone & \textbf{33.48 (1)} &	\textbf{17.42 (1)} &	29.07 (2) &	23.62 (2)\\
        \tasktwotwo	& 33.66 (2) &	17.45 (2) &	\textbf{27.76 (1)} &	\textbf{23.39} (1)\\
        \tasktwothree	& 38.33 (4) &	19.71 (4)	& 32.60 (4)	& 26.26 (4)\\
        \tasktwofour	& 47.18 (5) &	24.95 (6) &	38.76 (5)	& 32.36 (5)\\
        \bottomrule
    \end{tabular}
}
\end{minipage}
\endgroup
\end{table}

\noindent\textbf{Multi-modal inputs for HPEs.}
\taskthreethree adopts~\cite{yang2019aligning} in Task 3 to align latent spaces from depth and RGB input modalities and to embed the inherit depth information in depth images during learning. \taskonefive makes use of multi-inputs where each is obtained from different representations of the depth image, \eg~point-cloud, 3D point projection~\cite{ge2016robust}, multi-layer depth map~\cite{shin2019multi}, depth voxel~\cite{v2vposenet}.

\noindent\textbf{Dominating HPE backbones.} ResNet~\cite{resnet50} architectures with residual connections have been a popular backbone choice among many HPEs \eg~\taskonetwo, \taskonethree, \taskonefive, \tasktwothree, \taskthreeone, \taskthreetwo or implicitly by \taskonesix within the ResPEL~\cite{respel} architecture. \taskoneone adopts EfficientNet-b0~\cite{efficientnet} as a backbone which uniformly scales the architecture's depth, width, and resolution.

%% file: input_Results_and_analysis.tex
\definecolor{mypurple}{rgb}{0.6,0.,0.6}
\section{Results and Discussion}
\label{sec:analysis}
We share our insights and analysis of the results obtained by the participants' approaches: 6 in Task 1, 4 in Task 2, and 3 in Task 3. Our analyses highlight the impacts of data pre-processing, the use of an ensemble approach, the use of MANO model, different HPE methods, and backbones and post-processing strategies for the pose refinement.


\begingroup
\setlength{\tabcolsep}{1pt} 

\begin{figure*}[!ht]
\centering
\centering
\begin{tabular}{c c c}
{\includegraphics[width=0.33\linewidth]{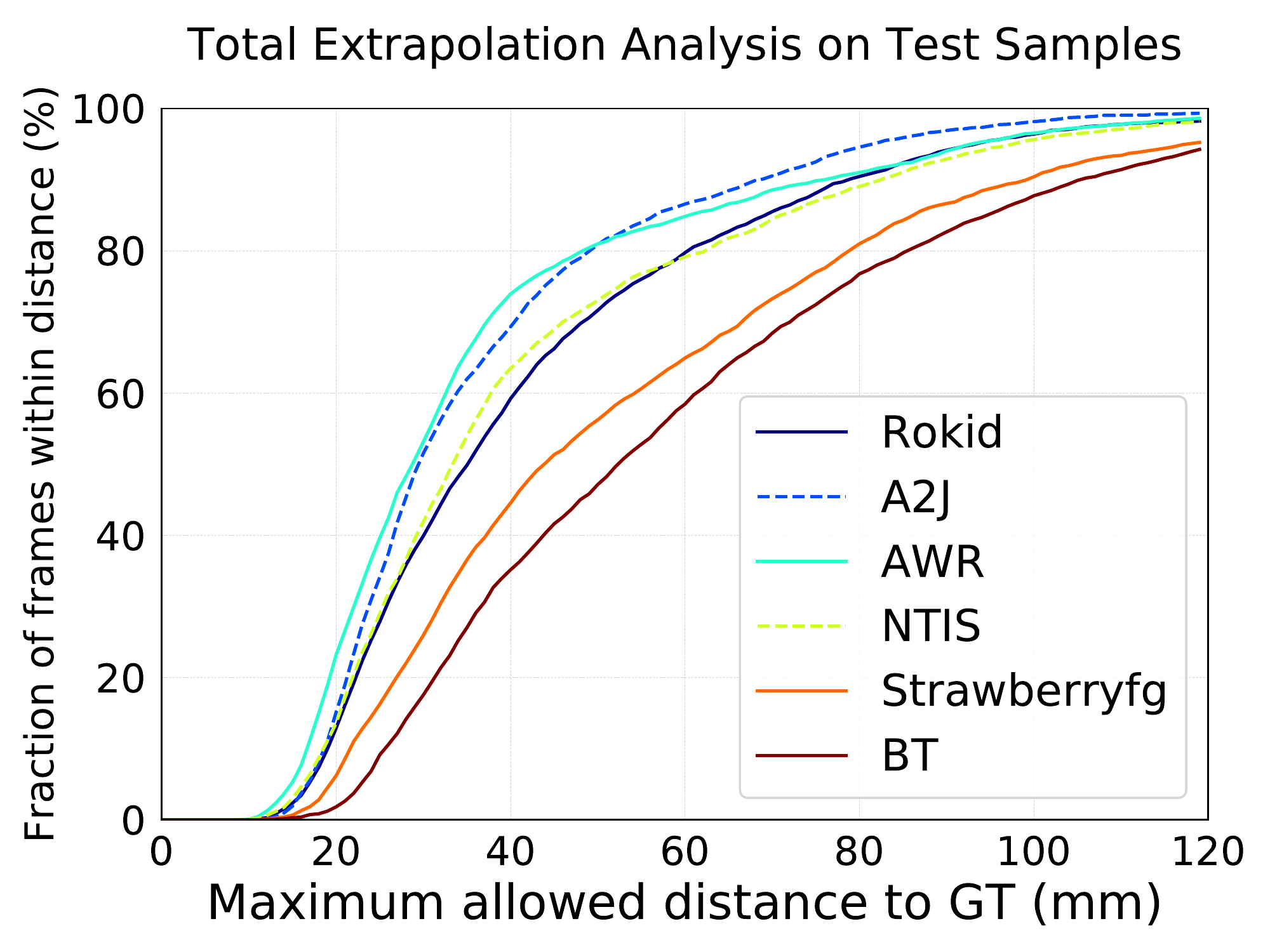}}& 
{\includegraphics[width=0.33\linewidth]{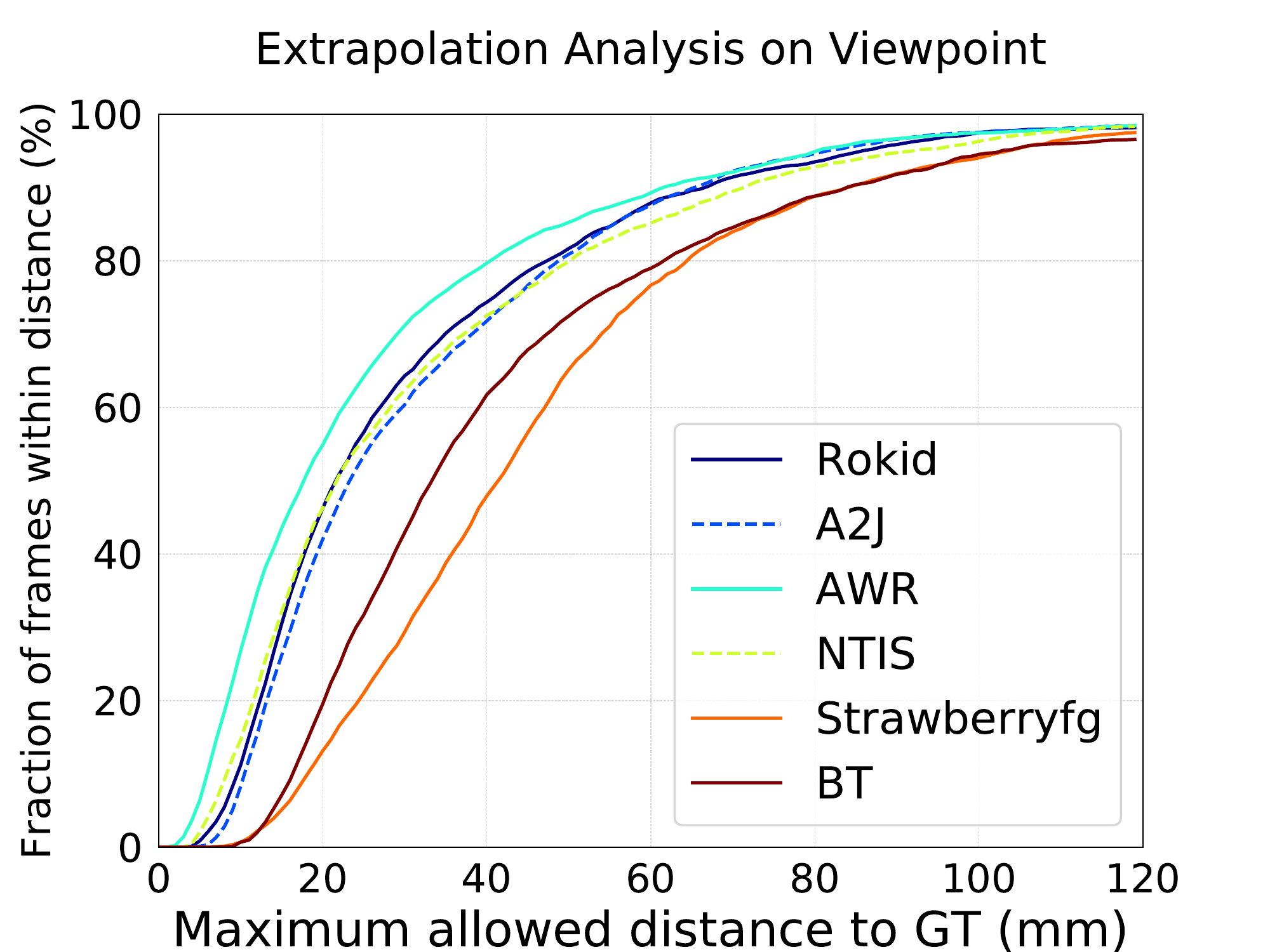}} &
{\includegraphics[width=0.33\linewidth]{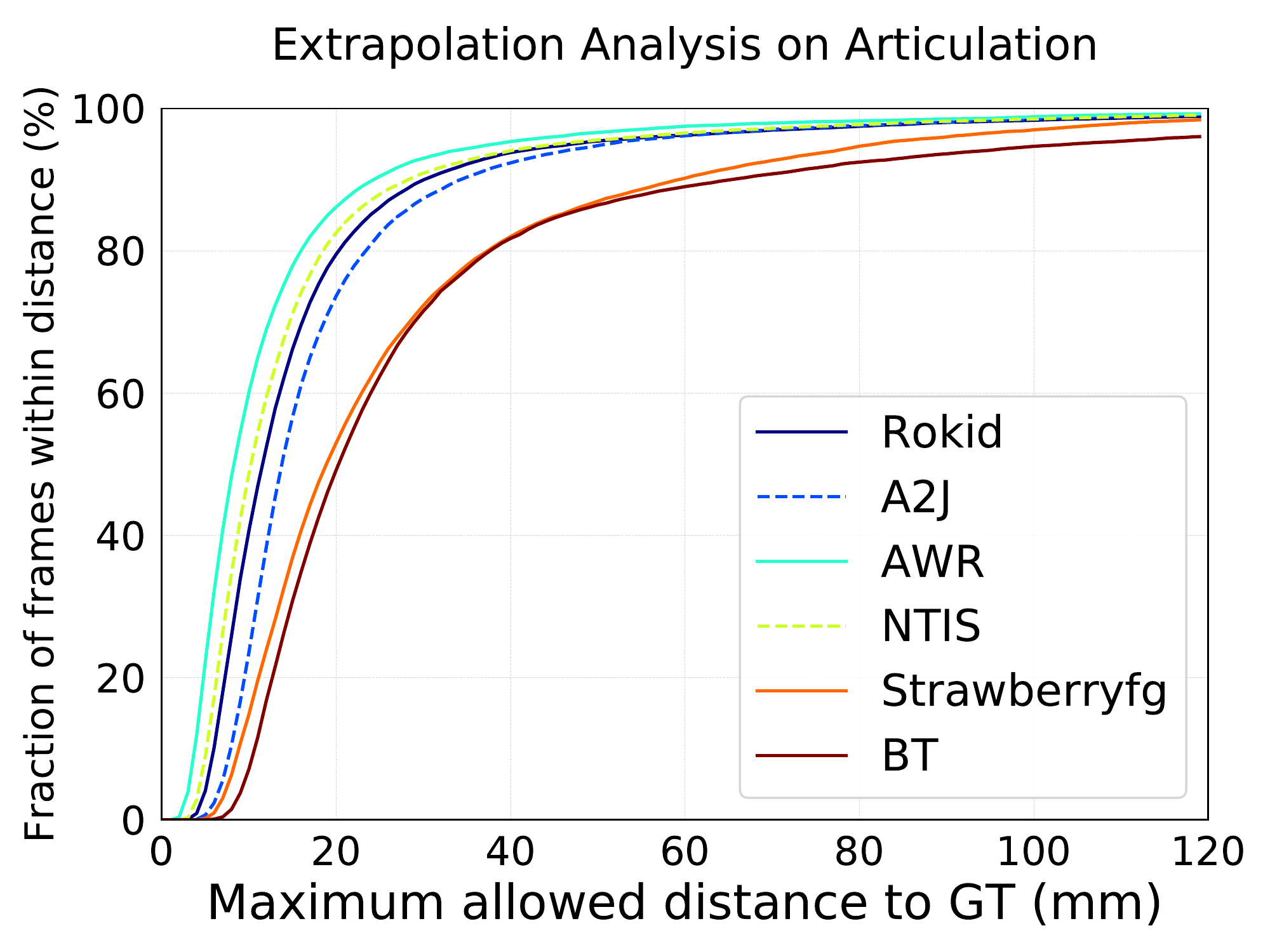}}
\\
{\small{(a) Extrapolation}} & {\small{(b) Viewpoint}} & {\small{(c) Articulation}} 
\end{tabular}
\begin{tabular}{c c c}
\centering
{\includegraphics[width=0.33\linewidth]{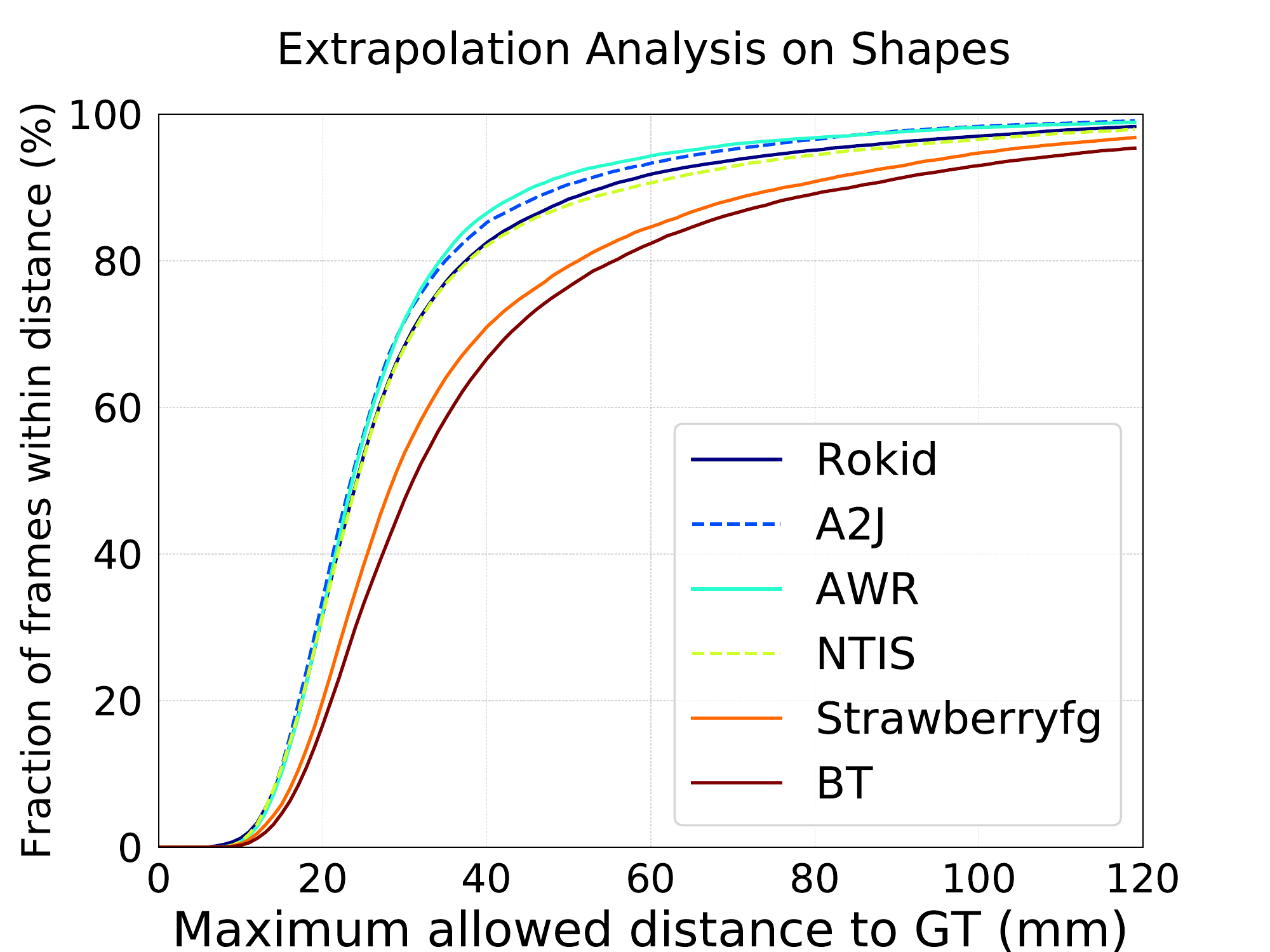}}&
{\includegraphics[width=0.33\linewidth]{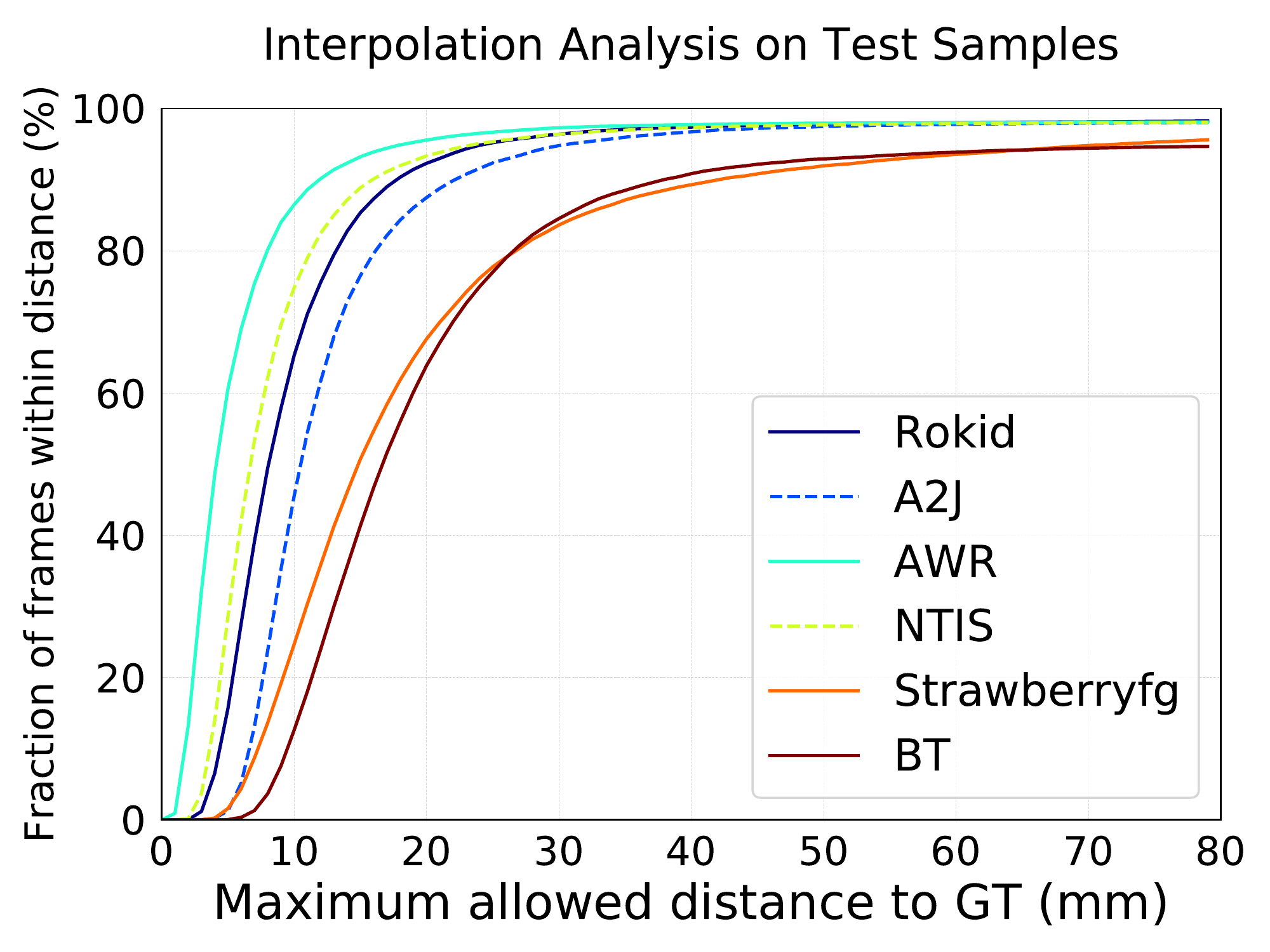}}&
{\includegraphics[width=0.33\linewidth]{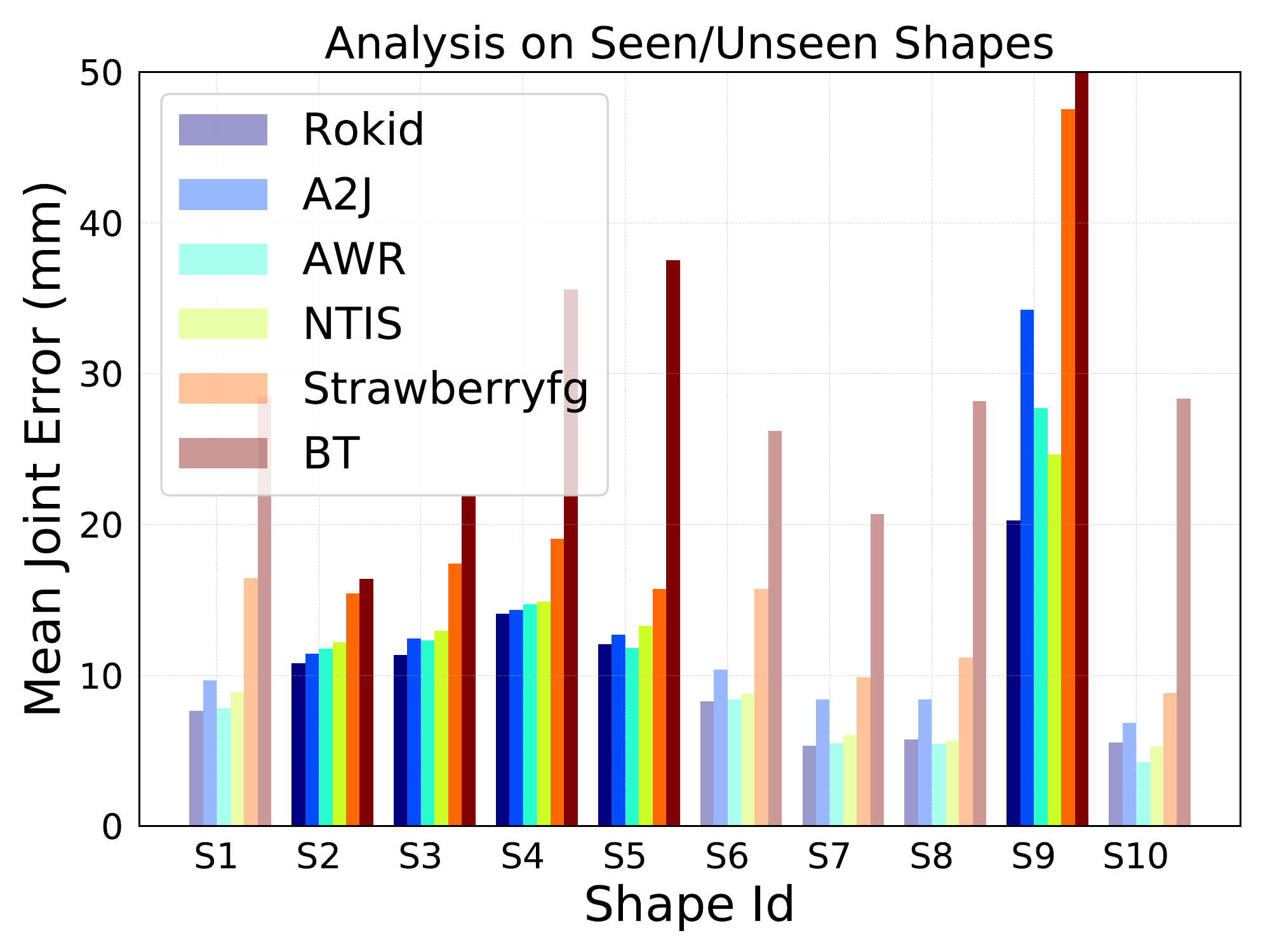}}
\\
{\small{(d) Shape}} & {\small{(e) Interpolation}} & {\small{(f) Shape MJE}}
\end{tabular}

\begin{tabular}{c c c}
\centering
{\includegraphics[width=0.33\linewidth]{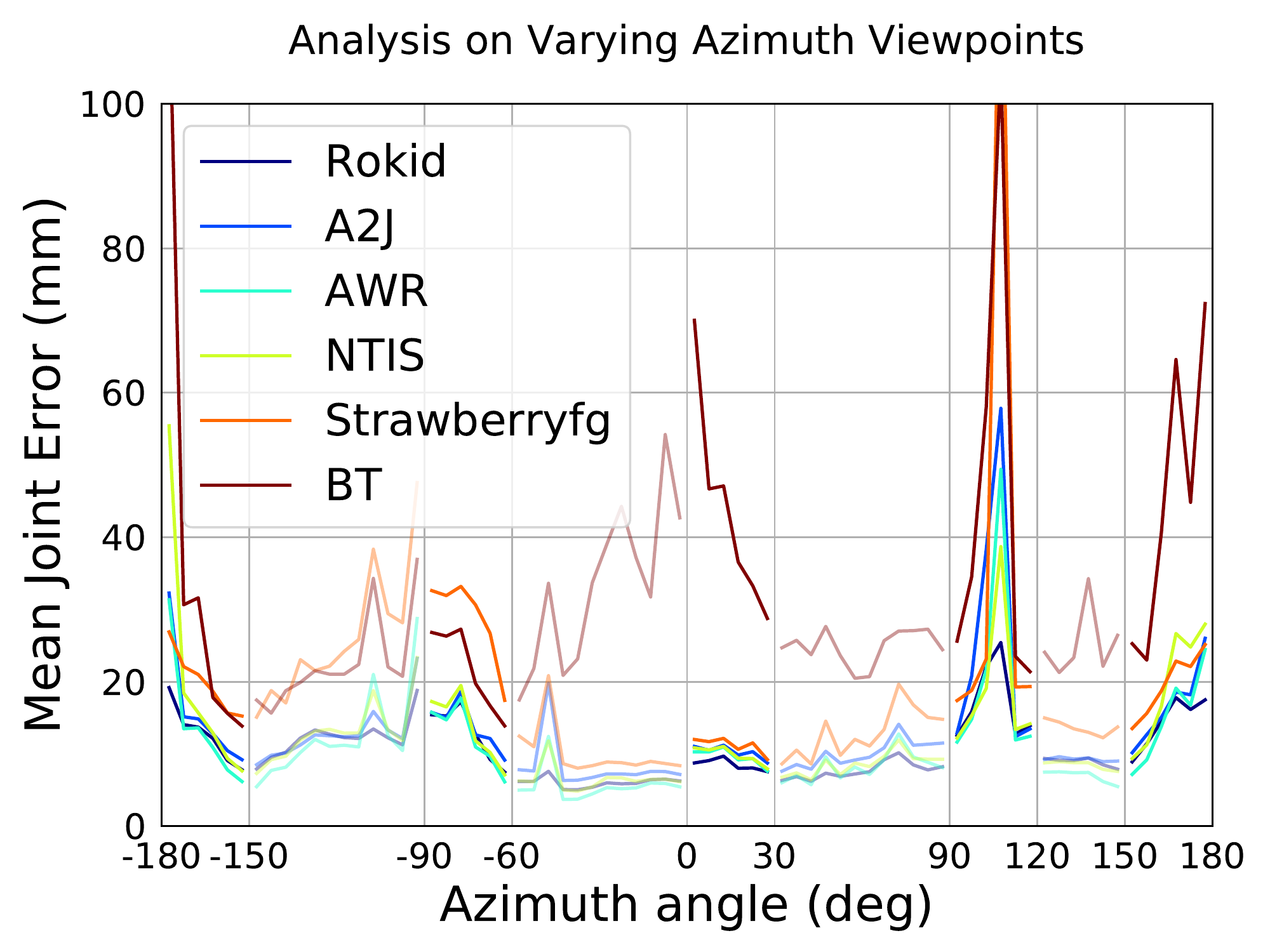}} &
{\includegraphics[width=0.33\linewidth]{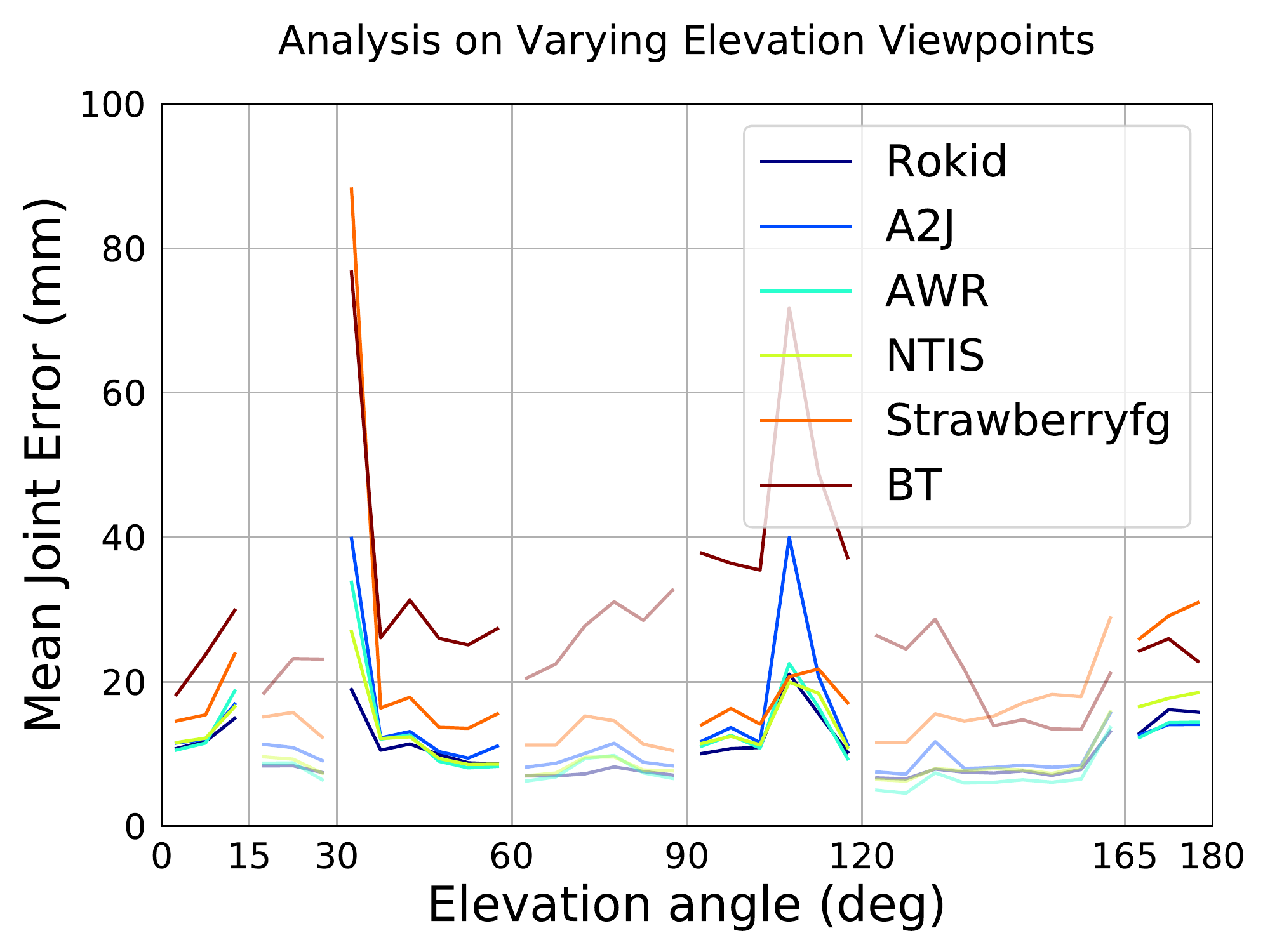}} &
{\includegraphics[width=0.33\linewidth]{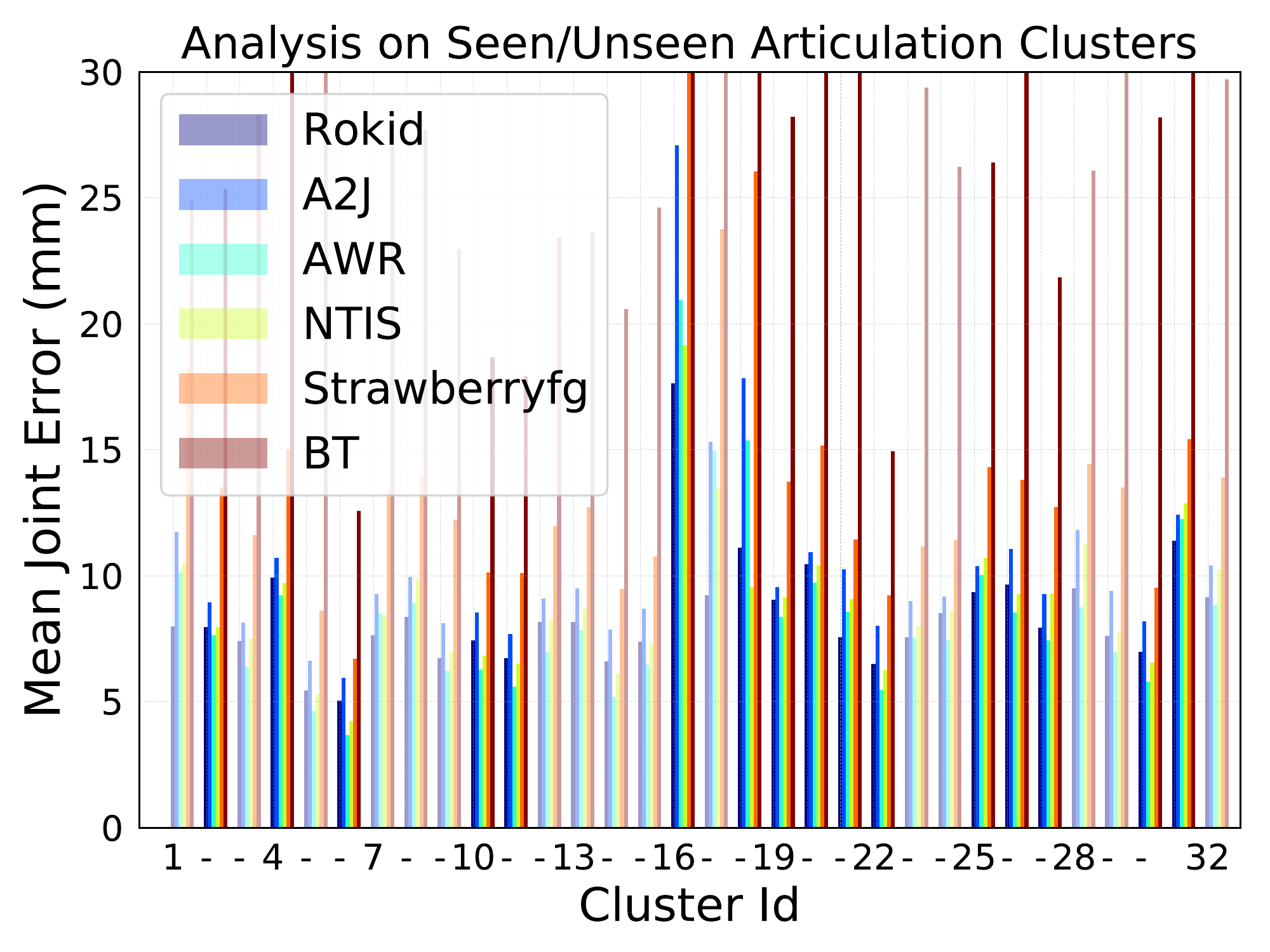}}
\\
{\small{(g) Azimuth Viewpoint}} & {\small{(h) Elevation Viewpoint}} & {\small{(i) Articulation MJE}}
\end{tabular}
\vspace{-15pt}
\caption{Task 1 - Success rate analysis (a-e) and MJE analysis on  extrapolation and interpolation using shapes (f), viewpoints (g, h) and articulations (i). Solid colors depict samples of extrapolation and transparent colors depict interpolation samples.}
\vspace{-25pt}
\label{fig:task1_extrapolation_plots}

\end{figure*} 
\endgroup

\begin{figure}[!hb]
\centering
\begin{tabular}{c c c c c}
\includegraphics[width=0.17\linewidth,height=0.17\linewidth]{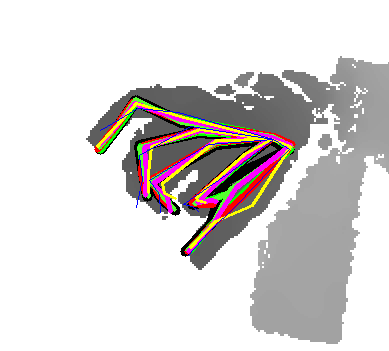} &
\includegraphics[width=0.17\linewidth,height=0.17\linewidth]{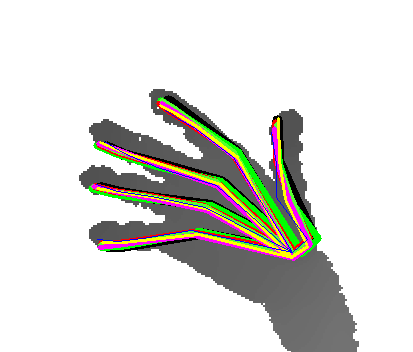} &
\includegraphics[width=0.17\linewidth,height=0.17\linewidth]{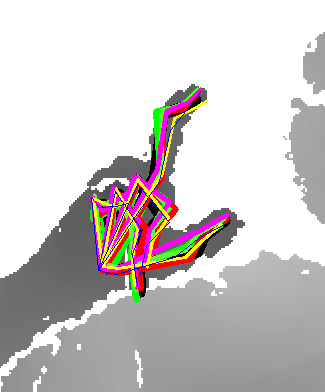} &
\includegraphics[width=0.17\linewidth,height=0.17\linewidth]{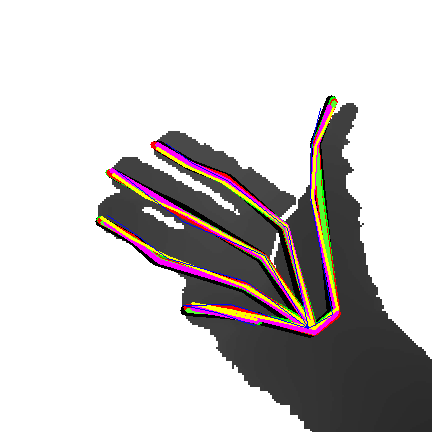} &
\includegraphics[width=0.17\linewidth,height=0.17\linewidth]{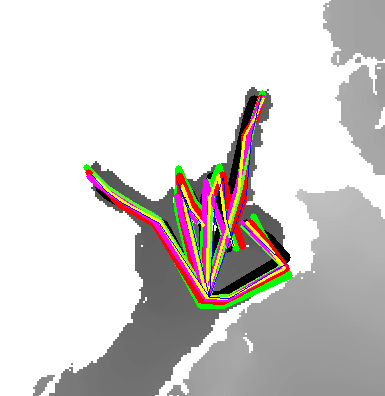}\\
\includegraphics[width=0.17\linewidth,height=0.17\linewidth]{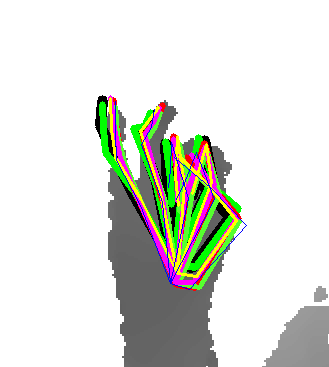} &
\includegraphics[width=0.17\linewidth,height=0.17\linewidth]{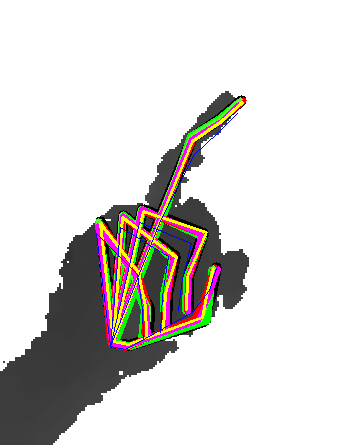} &
\includegraphics[width=0.17\linewidth,height=0.17\linewidth]{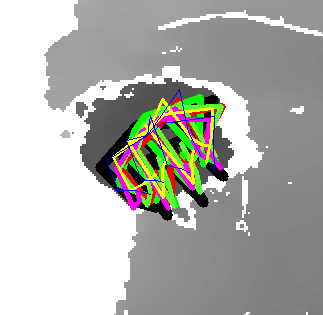} &
\includegraphics[width=0.17\linewidth,height=0.17\linewidth]{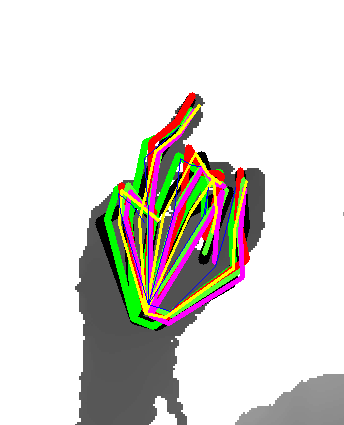} &
\includegraphics[width=0.17\linewidth,height=0.17\linewidth]{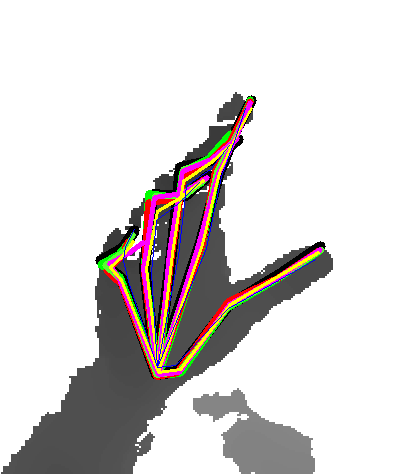}\\
\includegraphics[width=0.17\linewidth,height=0.17\linewidth]{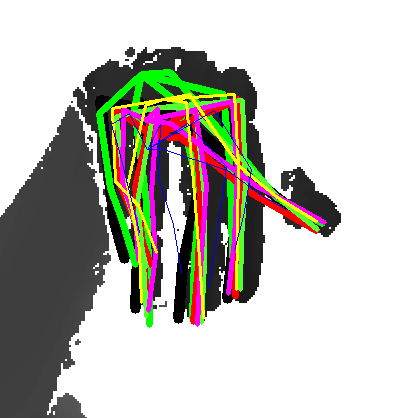} &
\includegraphics[width=0.17\linewidth,height=0.17\linewidth]{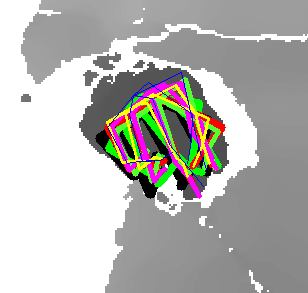} &
\includegraphics[width=0.17\linewidth,height=0.17\linewidth]{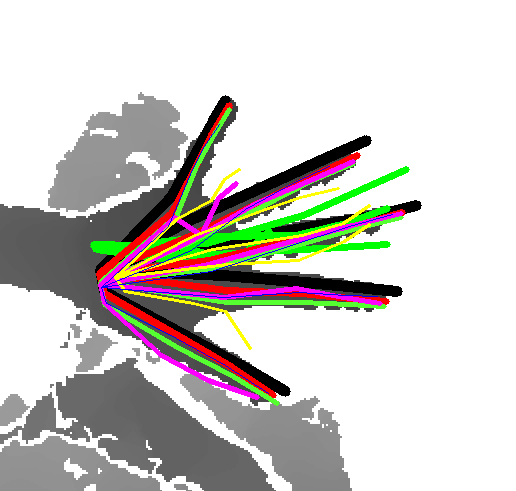} &
\includegraphics[width=0.17\linewidth,height=0.17\linewidth]{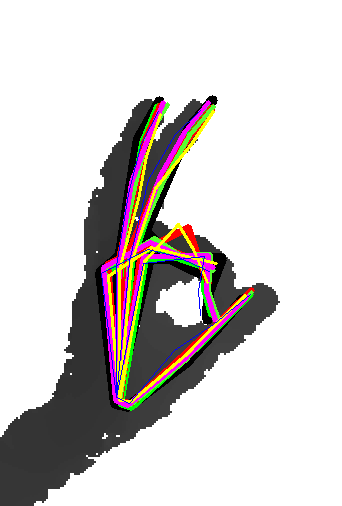} &
\includegraphics[width=0.17\linewidth,height=0.17\linewidth]{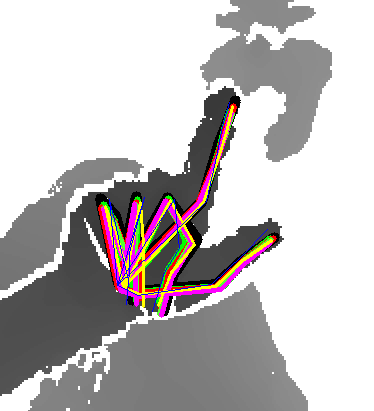}
\end{tabular}
\vspace{-5pt}
\caption{Task 1 - Visualization of the \textbf{\color{black}{ground-truth}} annotations and estimations of \textbf{~\color{green}{\taskoneone}, ~\color{red}{\taskonetwo}, ~\color{mypurple}{\taskonethree}, ~\color{magenta}{\taskonefour}, ~\color{yellow}{\taskonefive}, ~\color{blue}{\taskonesix}}.}
\vspace{-20pt}
\label{fig:task1_qualitatives}
\end{figure}

\noindent\textbf{Analysis of Submitted Methods for Task 1.}
\label{sec:task1_analysis}
We consider the main properties of the selected methods and the evaluation criteria for comparisons. Table~\ref{tab:task1_results} provides the errors for the MJE metric and Fig.~\ref{fig:task1_extrapolation_plots} show that high success rates are easier to achieve in absence of an object for low distance $d$ thresholds. 2D-based approaches such as \taskoneone, with the advantage of additional data synthesizing, or \taskonetwo, with cleverly designed local regressors, can be considered to be best when the MJE score is evaluated for the Extrapolation criterion. \taskonethree performs comparable to the other 2D-based approaches by obtaining the lowest MJE errors on the Interpolation and Articulation criteria. \taskonethree  performs best for the distances less than $50mm$ on Extrapolation as well as showing better generalisation to unseen Viewpoints and Articulations, while excelling to interpolate well. A similar trend is observed with the 3D-voxel-based approach \taskonefour. However, the other 3D supervised methods, \taskonefive and \taskonesix show lower generalisation capability compared to other approaches while performing reasonably well on the Articulation, Shape, and Interpolation criteria but not being able to show a similar performance for the Extrapolation and Viewpoint criteria.

\noindent\textbf{Analysis of Submitted Methods for Task 2.}
\label{sec:task2_analysis}
We selected four submitted methods to compare on Task 2, where a hand interacts with an object in an egocentric viewpoint. 
Success rates illustrated in Fig.~\ref{fig:task2_extrapolation_plots} highlight the difficulty of extrapolation. All methods struggle to show good performance on estimating frames with joint errors less than $15mm$. On the other hand, all methods can estimate $20\%$ to $30\%$ of the joints correctly with less than $15mm$ error for the other criteria in this task.

\tasktwoone (a voxel-based) and \tasktwotwo (weighted local regressors with anchor points) perform similarly when MJEs for all joints are considered. However, \tasktwoone obtains higher success rates on the frame-based evaluation for all evaluation criteria with low distance error thresholds ($d$), see Fig.~\ref{fig:task2_extrapolation_plots}. Its performance is relatively much higher when Extrapolation is considered, especially for the frames with unseen objects, see Fig.~\ref{fig:task2_extrapolation_plots}. This can be explained by having a better embedding of the occluded hand structure with the voxels in the existence of seen/unseen objects. \tasktwoone interpolates well under low distance thresholds. 

Note that the first three methods, \tasktwoone, \tasktwotwo, and \tasktwothree perform very similar for high error thresholds \eg~$d>30mm$. \tasktwothree uses a structured detection-regression-based HPE where a heatmap regression is employed for the joints from palm to tips in a sequential manner which is highly valuable for egocentric viewpoints, helps to obtain comparable results with \tasktwotwo where the structure is implicitly refined by the local anchor regressors.

\begingroup
\begin{figure*}[!ht]
\centering
\begin{tabular}{c c c}
\centering
{\includegraphics[width=0.33\textwidth]{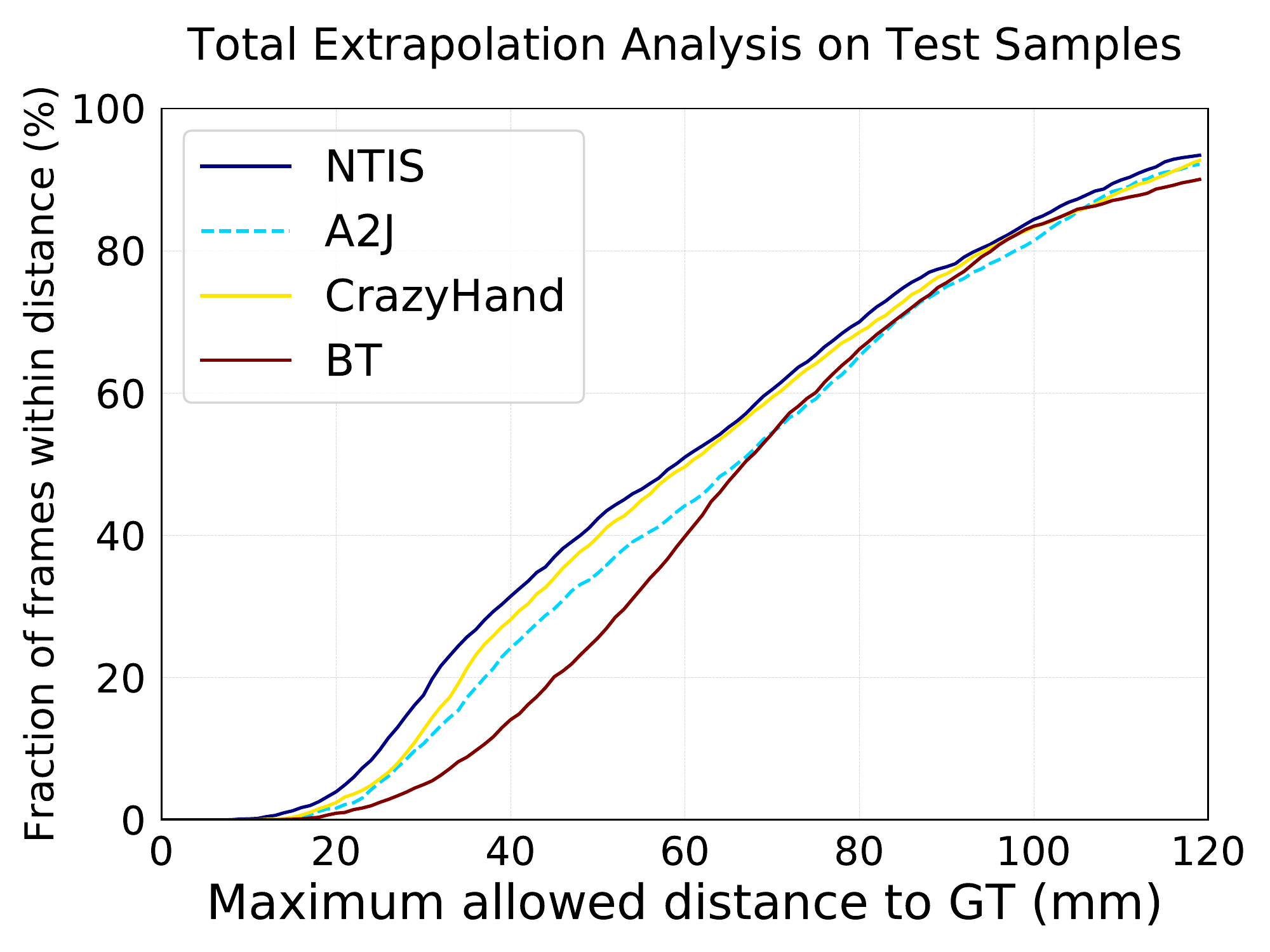}} &
{\includegraphics[width=0.33\textwidth]{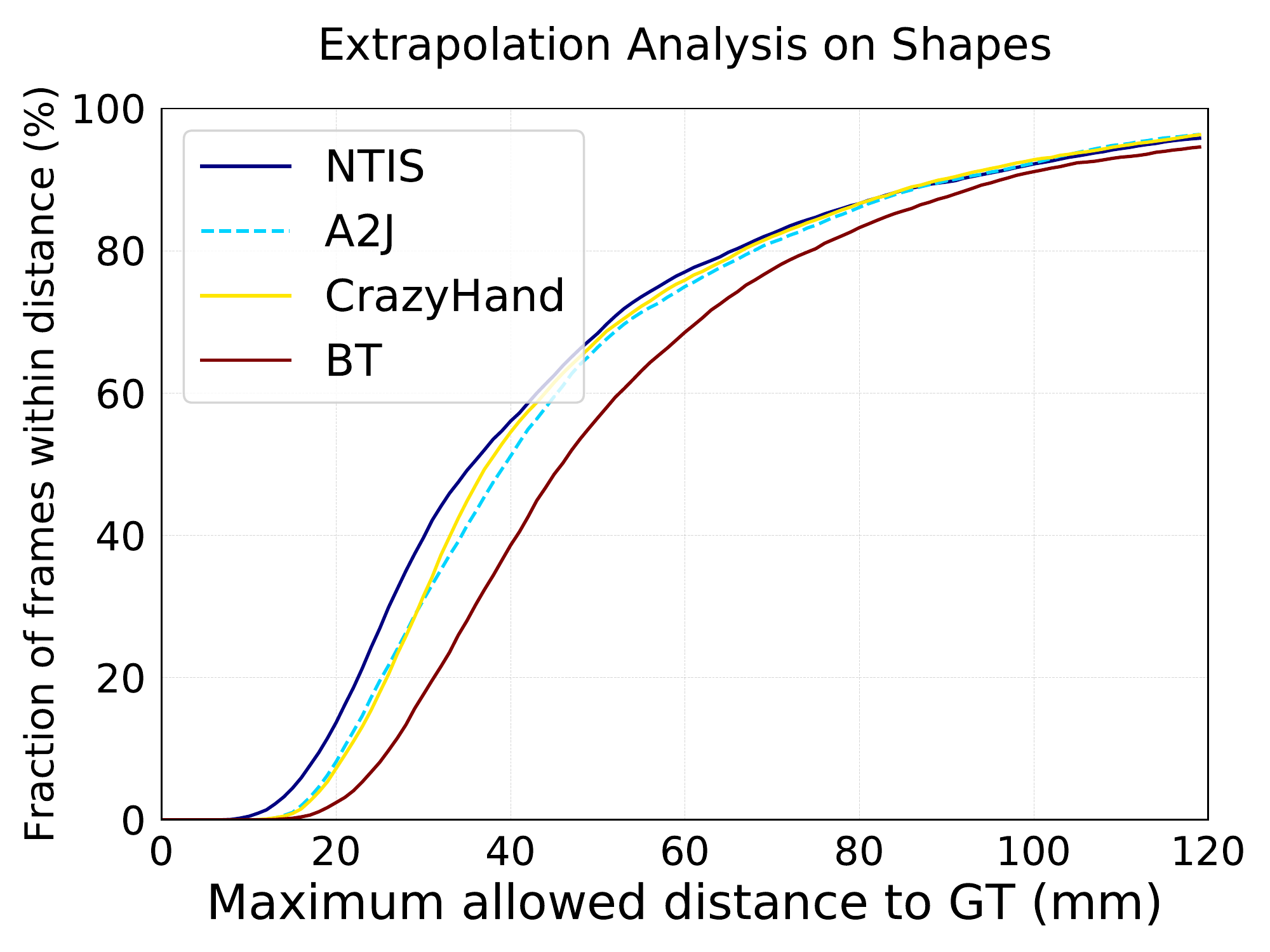}} &
{\includegraphics[width=0.33\textwidth]{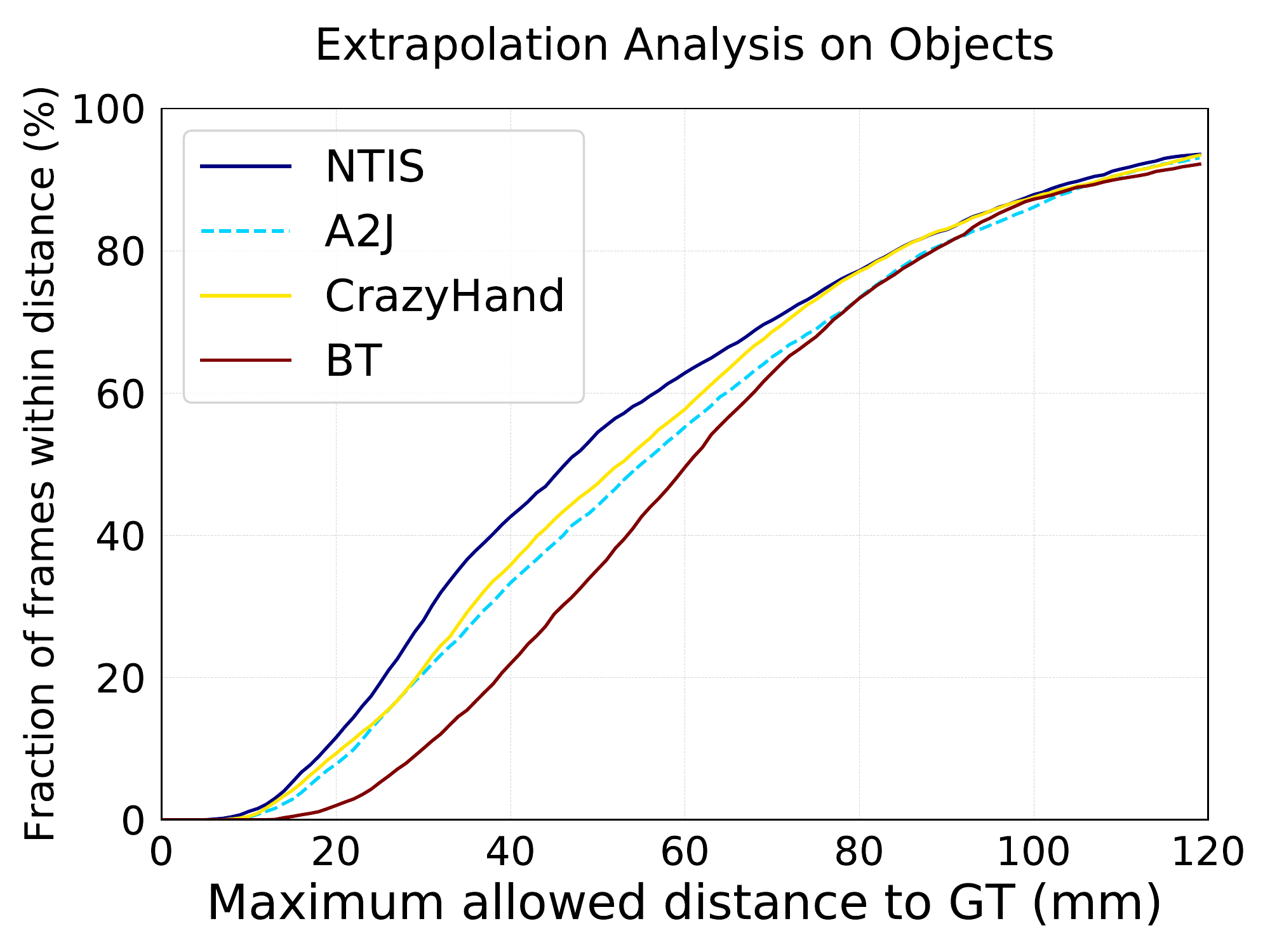}}
\\
{\small{(a) Extrapolation}} &
{\small{(b) Shape}} &
{\small{(c) Object}} 
\end{tabular}
\\
\begin{tabular}{c c}
\centering
{\includegraphics[width=0.4\textwidth]{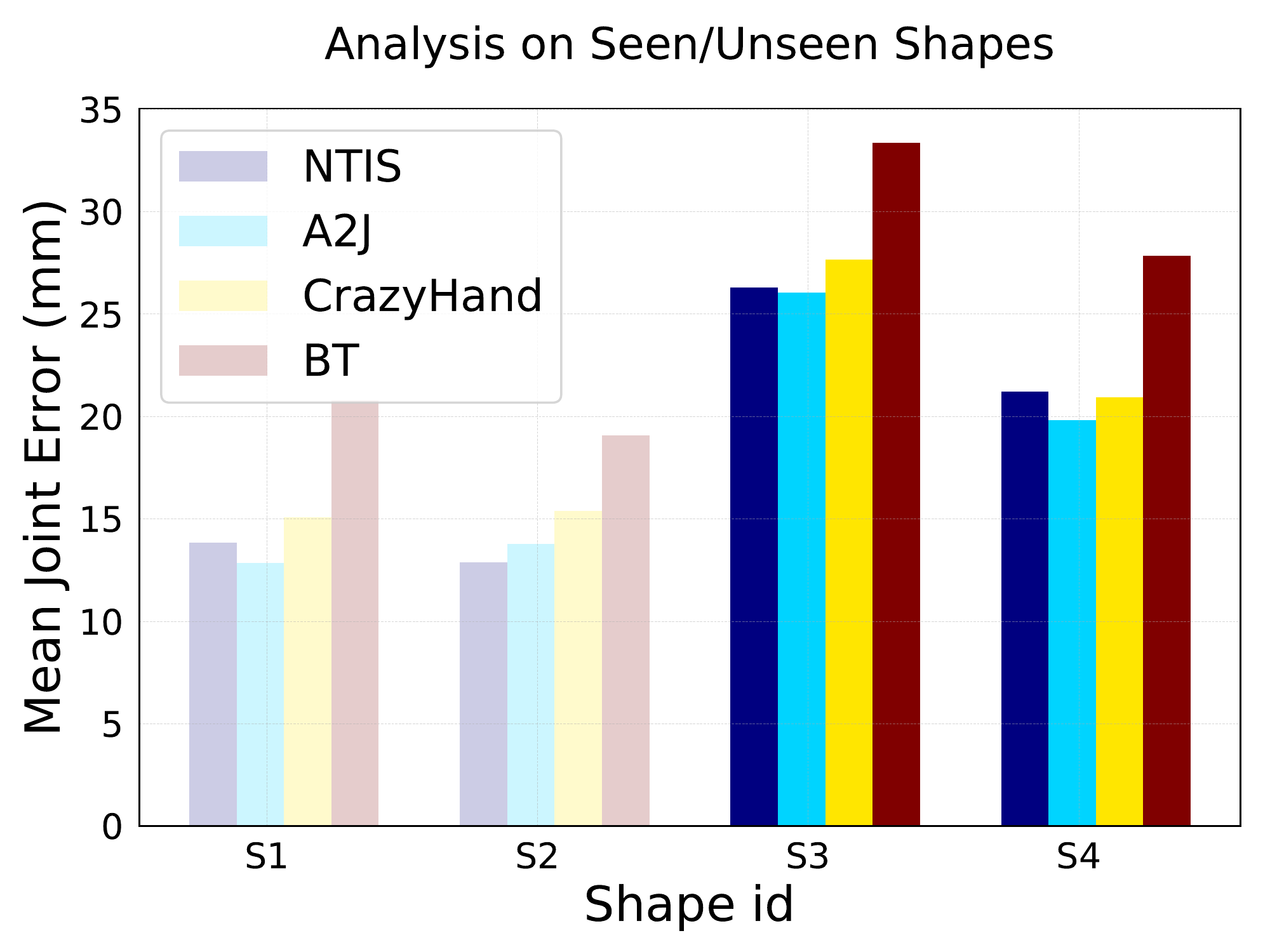}} &
{\includegraphics[width=0.4\textwidth]{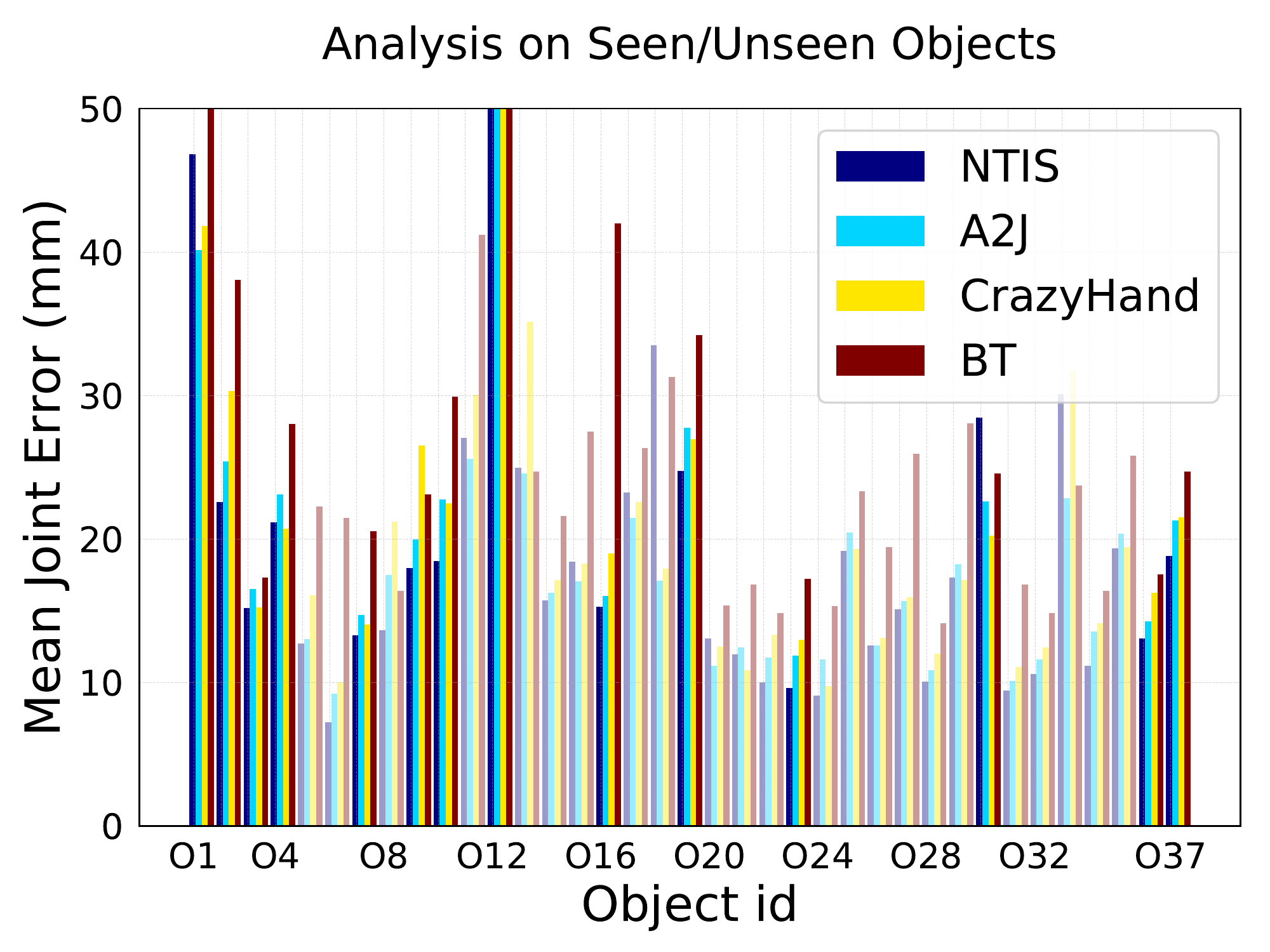}} 
\\
{\small{(d) Shape MJE}} & {\small{(e) Object MJE}}
\end{tabular}
\caption{Task 2 - Success rate analysis (a,b,c) and interpolation (seen, transparent) and extrapolation (unseen, solid) errors for subject (d) and object (e).}
\label{fig:task2_extrapolation_plots}
\end{figure*} 
\endgroup

\begin{figure}[!hb]
\centering
\begin{tabular}{c c c c c}
\includegraphics[width=0.19\linewidth,height=0.19\linewidth]{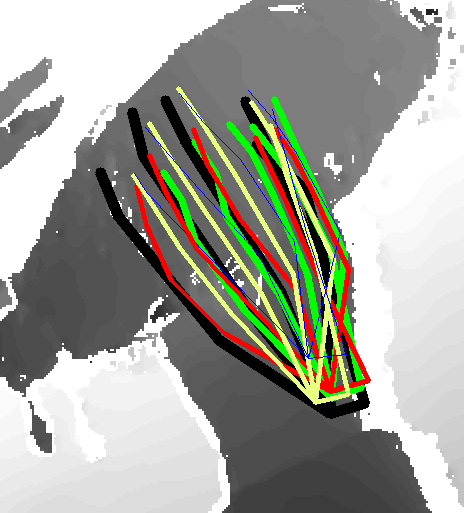} &
\includegraphics[width=0.19\linewidth,height=0.19\linewidth]{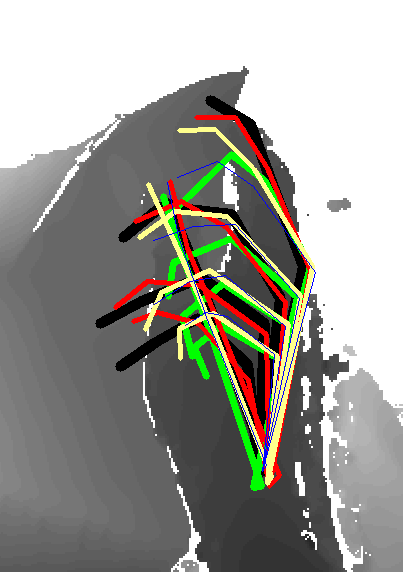} &
\includegraphics[width=0.19\linewidth,height=0.19\linewidth]{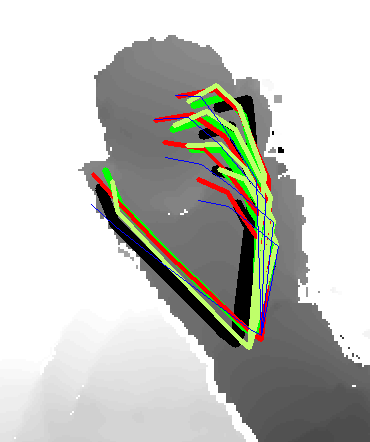} &
\includegraphics[width=0.19\linewidth,height=0.19\linewidth]{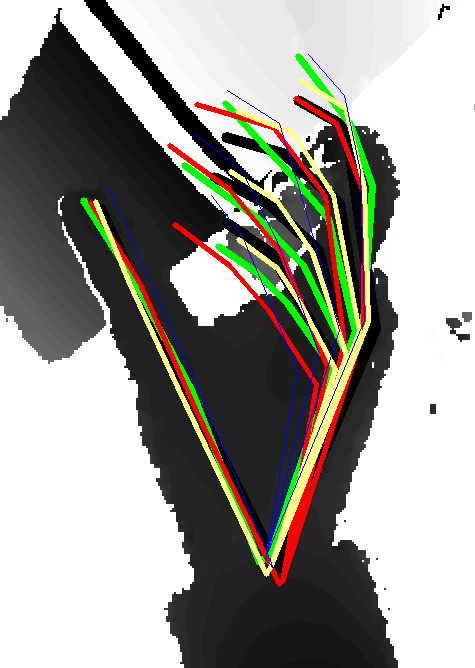} &
\includegraphics[width=0.19\linewidth,height=0.19\linewidth]{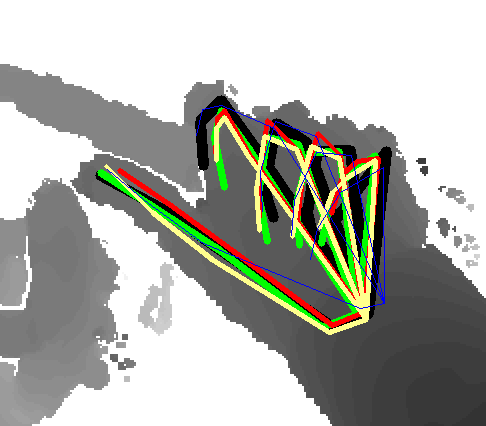}\\
\includegraphics[width=0.19\linewidth,height=0.19\linewidth]{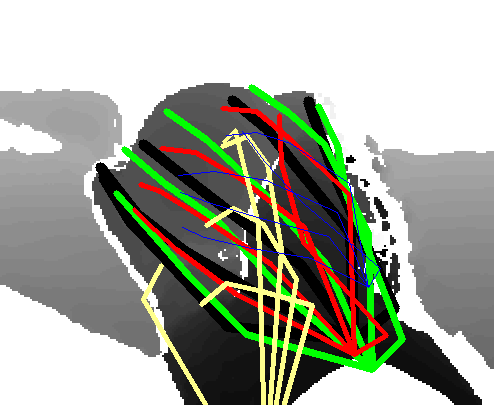} &
\includegraphics[width=0.19\linewidth,height=0.19\linewidth]{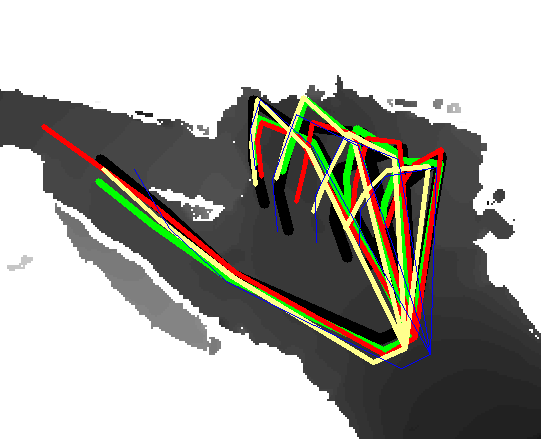} &
\includegraphics[width=0.19\linewidth,height=0.19\linewidth]{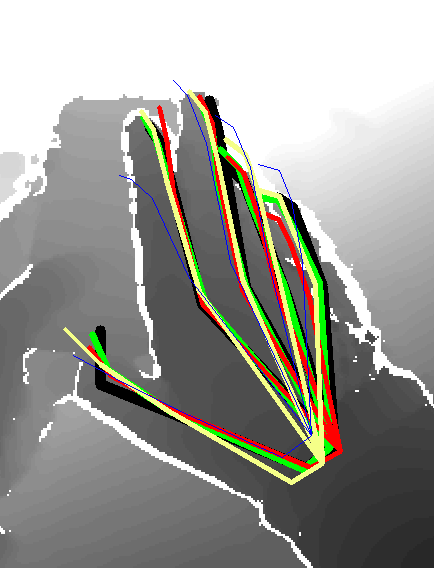} &
\includegraphics[width=0.19\linewidth,height=0.19\linewidth]{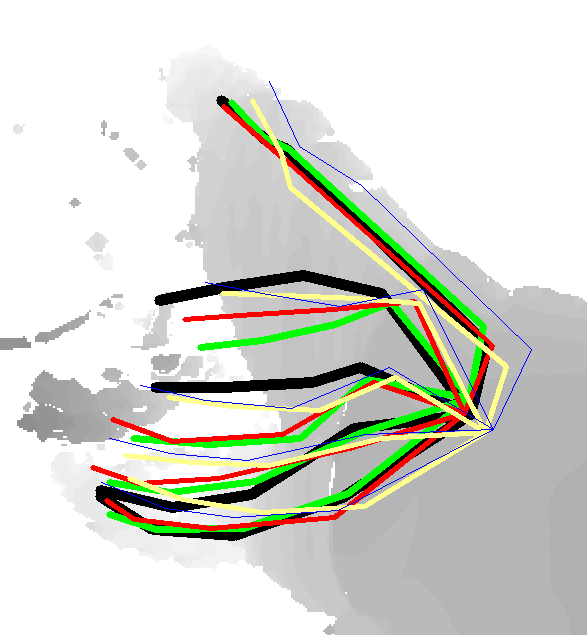} &
\includegraphics[width=0.19\linewidth,height=0.19\linewidth]{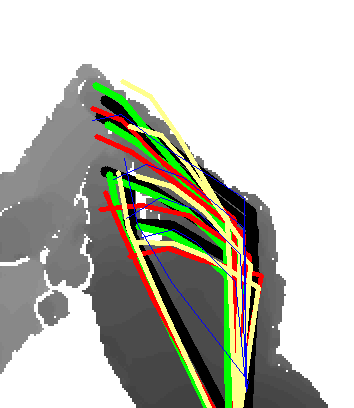}\\
\includegraphics[width=0.19\linewidth,height=0.19\linewidth]{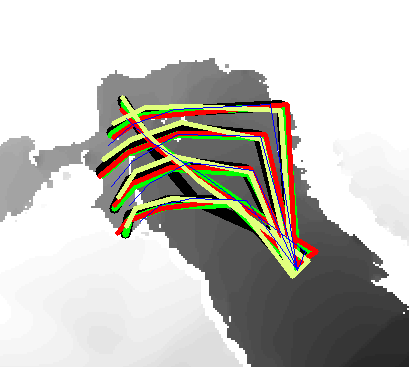} &
\includegraphics[width=0.19\linewidth,height=0.19\linewidth]{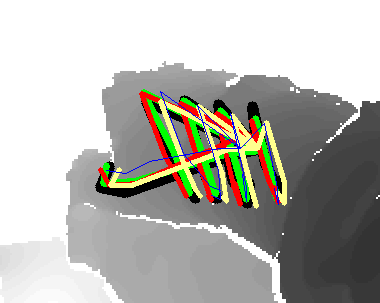} &
\includegraphics[width=0.19\linewidth,height=0.19\linewidth]{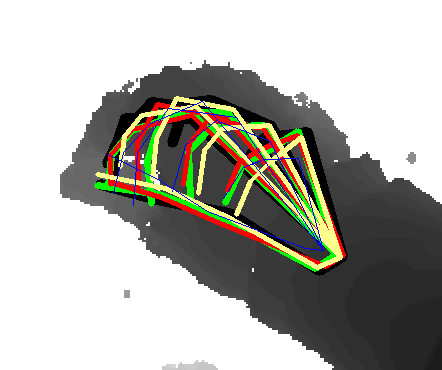} &
\includegraphics[width=0.19\linewidth,height=0.19\linewidth]{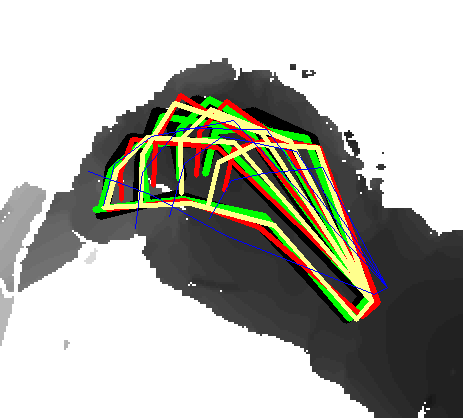} &
\includegraphics[width=0.19\linewidth,height=0.19\linewidth]{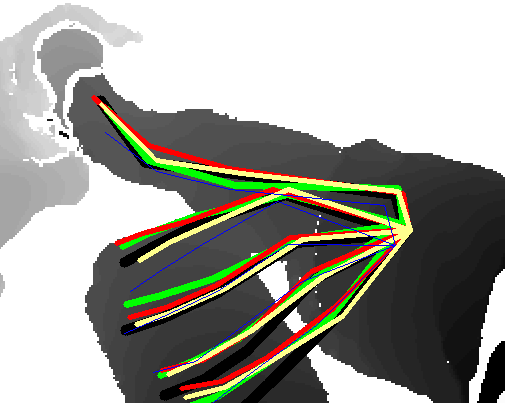}
\end{tabular}
\caption{Task 2 - Visualization of the \textbf{\color{black}{ground-truth}} annotations and estimations of \textbf{~\color{green}{\tasktwoone}, ~\color{red}{\tasktwotwo}, ~\color{yellow}{\tasktwothree}, ~\color{blue}{\tasktwofour}}.}
\label{fig:task2_qualitatives}
\end{figure}

\begingroup
\begin{table}[b]
\centering
\setlength{\tabcolsep}{2pt} 
\begin{minipage}{0.47\linewidth}
\centering
\caption{Task 3 - MJE (mm) and ranking of the methods on four evaluation criteria.}
\label{tab:task3_results}
\resizebox{1.\linewidth}{!}{
\begin{tabular}{c c c c c}
    \toprule
    Username & Extrapolation &	Interpolation & Object & Shape\\
    \hline
    \taskthreeone~ & \textbf{24.74} (1) & 6.70 (3) &	27.36 (2) &	\textbf{13.21} (1)\\
    \taskthreetwo~ & 29.19 (2)	& \textbf{4.06} (1) &	\textbf{18.39} (1) &	15.79 (3)\\
    \taskthreethree~ & 31.51 (3) & 19.15 (5)	& 30.59 (3) & 23.47 (4)\\
    \bottomrule
\end{tabular}
}
\end{minipage}
\hfill
\begin{minipage}{0.50\linewidth}
\centering
\caption{Impact of synthetic data reported by \taskoneone~\cite{rokid_arxiv} with learning from different ratios of synthetic data and the Task 1 training set. $100\%=570K$.}
    \label{tab:zzzh_mano_ratio}
\resizebox{1.\linewidth}{!}{
\begin{tabular}{c | c c c c c}
        \toprule
         Synthetic Data \% & - & 10\% & 30\% & 70\% & 100\% \\
         \hline
         Extrapolation MJE (mm) & 30.11 & 16.70 & 16.11 & 15.81 & 15.73\\
         \bottomrule
\end{tabular}
}
\end{minipage}
\end{table}
\endgroup

\begin{figure*}[t]
\centering
\begingroup
\renewcommand{\arraystretch}{0.5} 
\begin{tabular}{c c c}
{\includegraphics[width=0.33\linewidth]{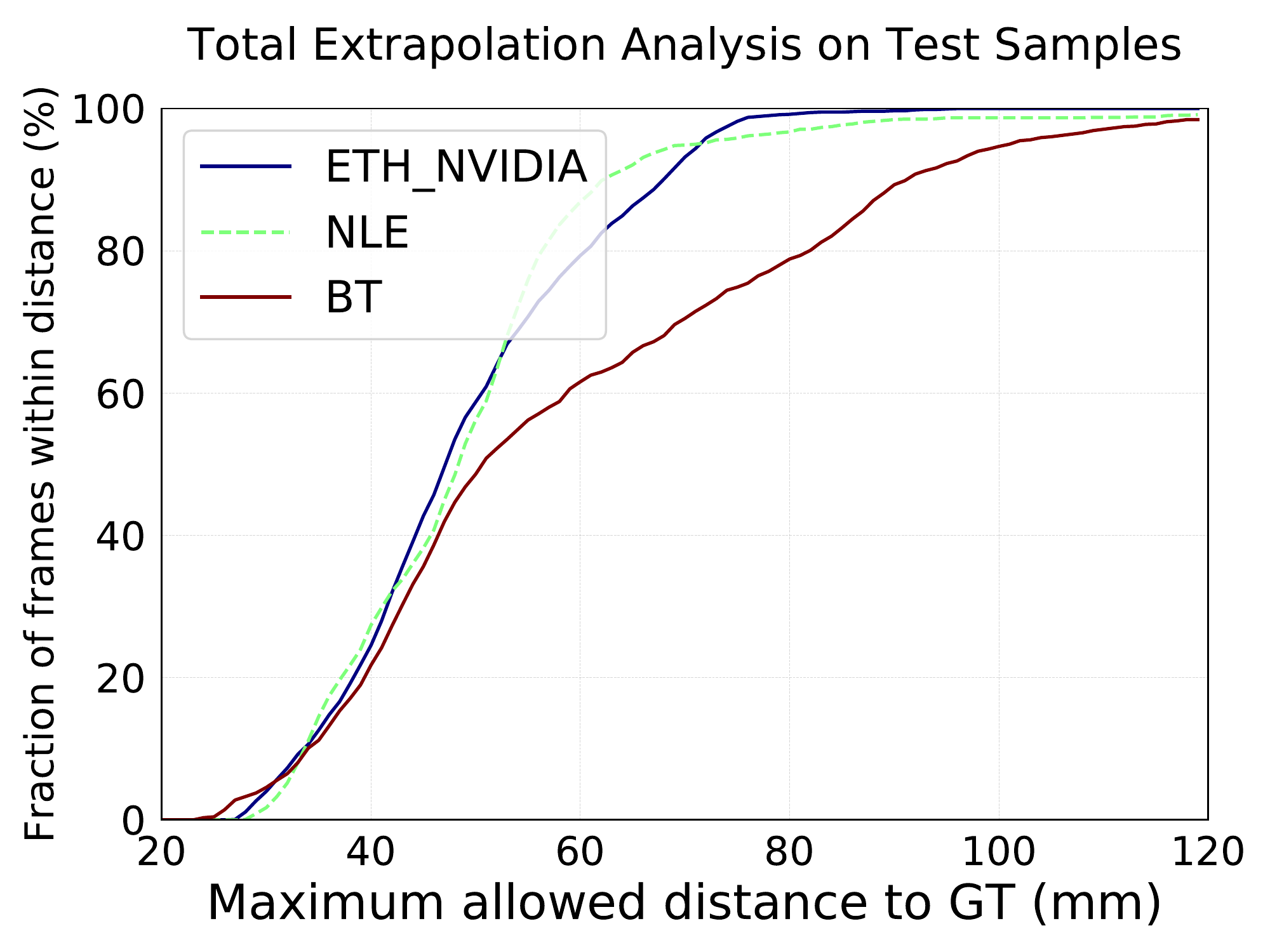}} &
{\includegraphics[width=0.33\linewidth]{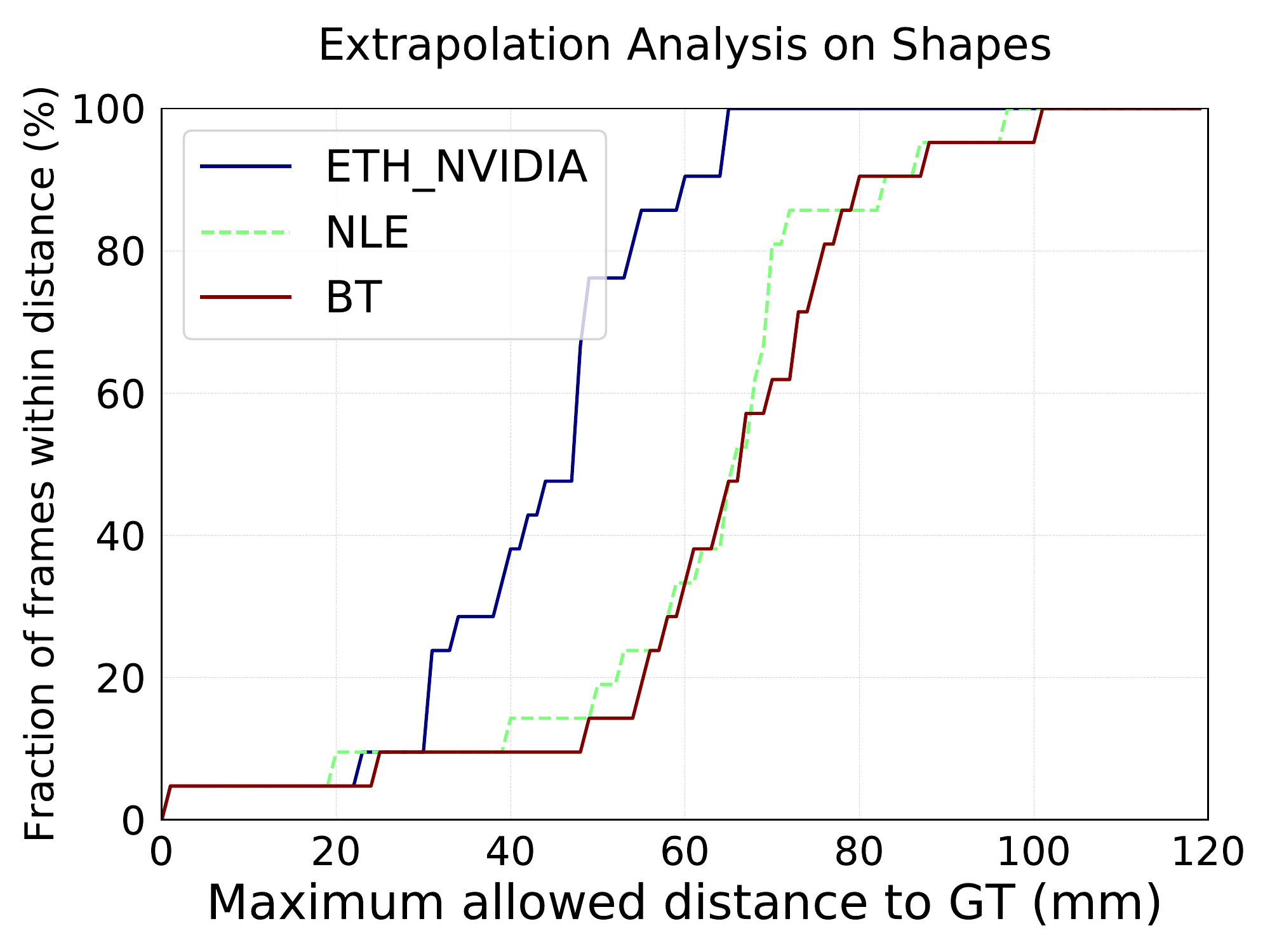}} & {\includegraphics[width=0.33\linewidth]{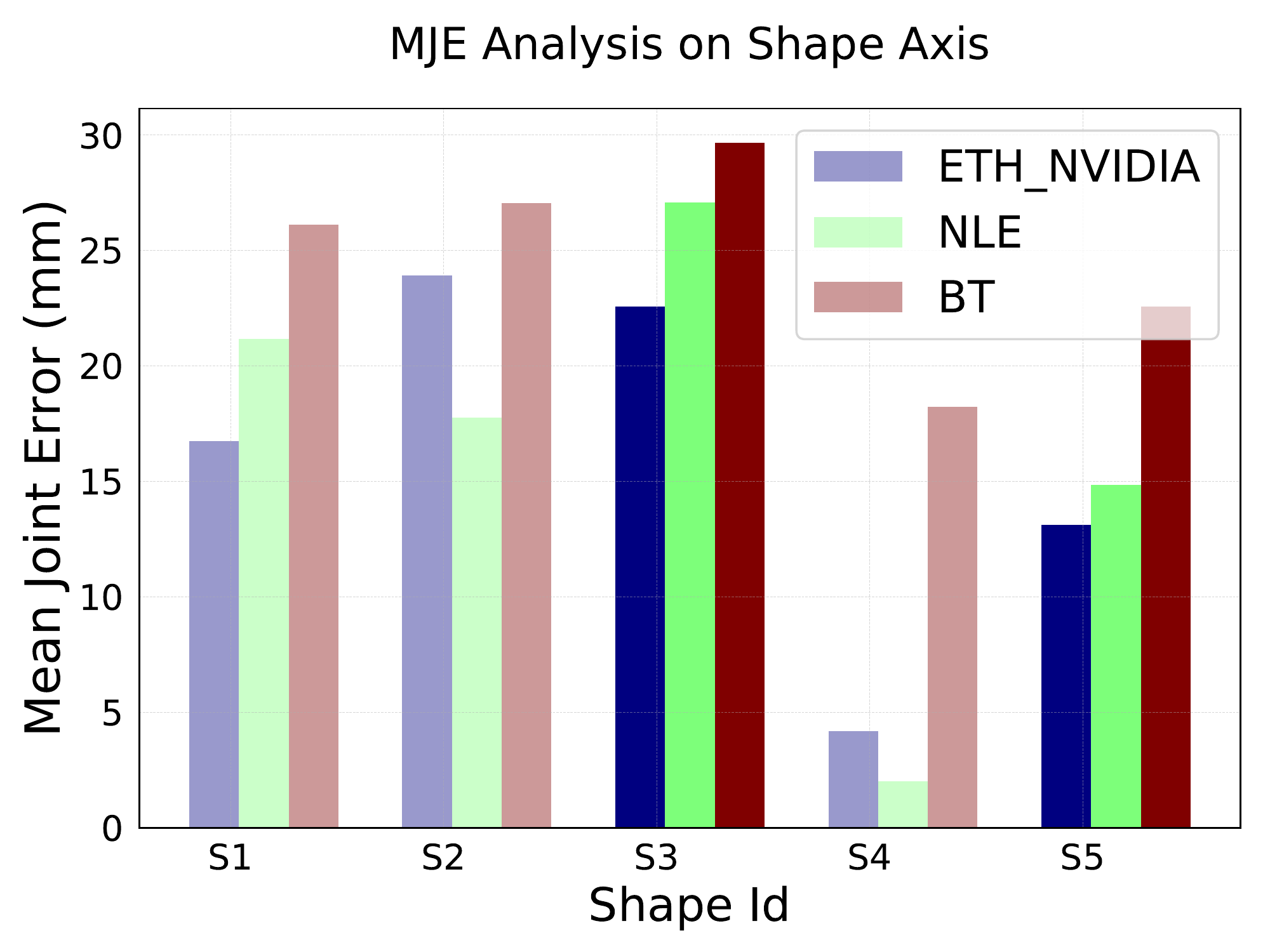}}
\\
{\small{(a) Extrapolation}} & {\small{(b) Shape}} & {(c) Shape MJE}
\\
{\includegraphics[width=0.33\linewidth]{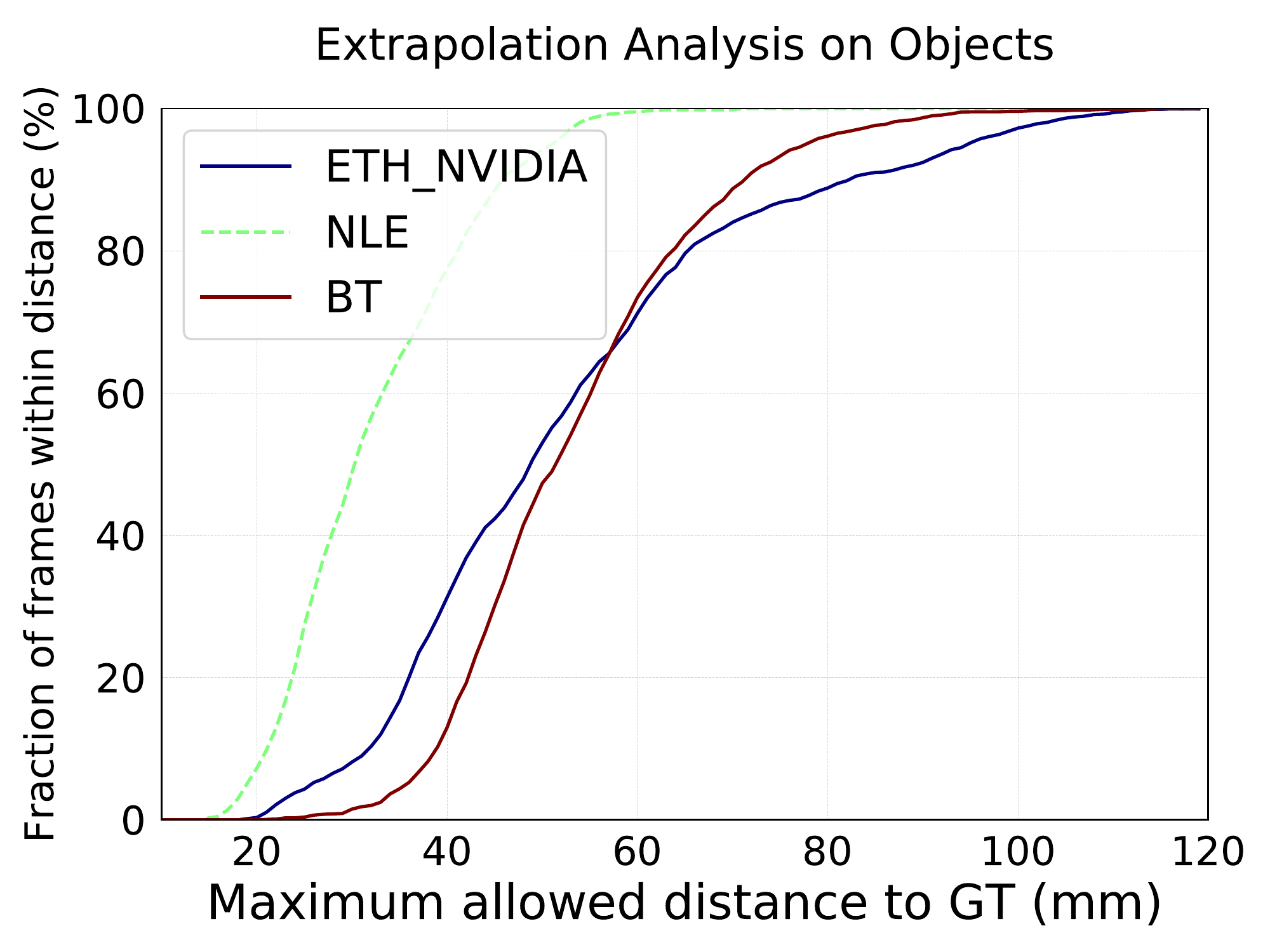}} &
{\includegraphics[width=0.33\linewidth]{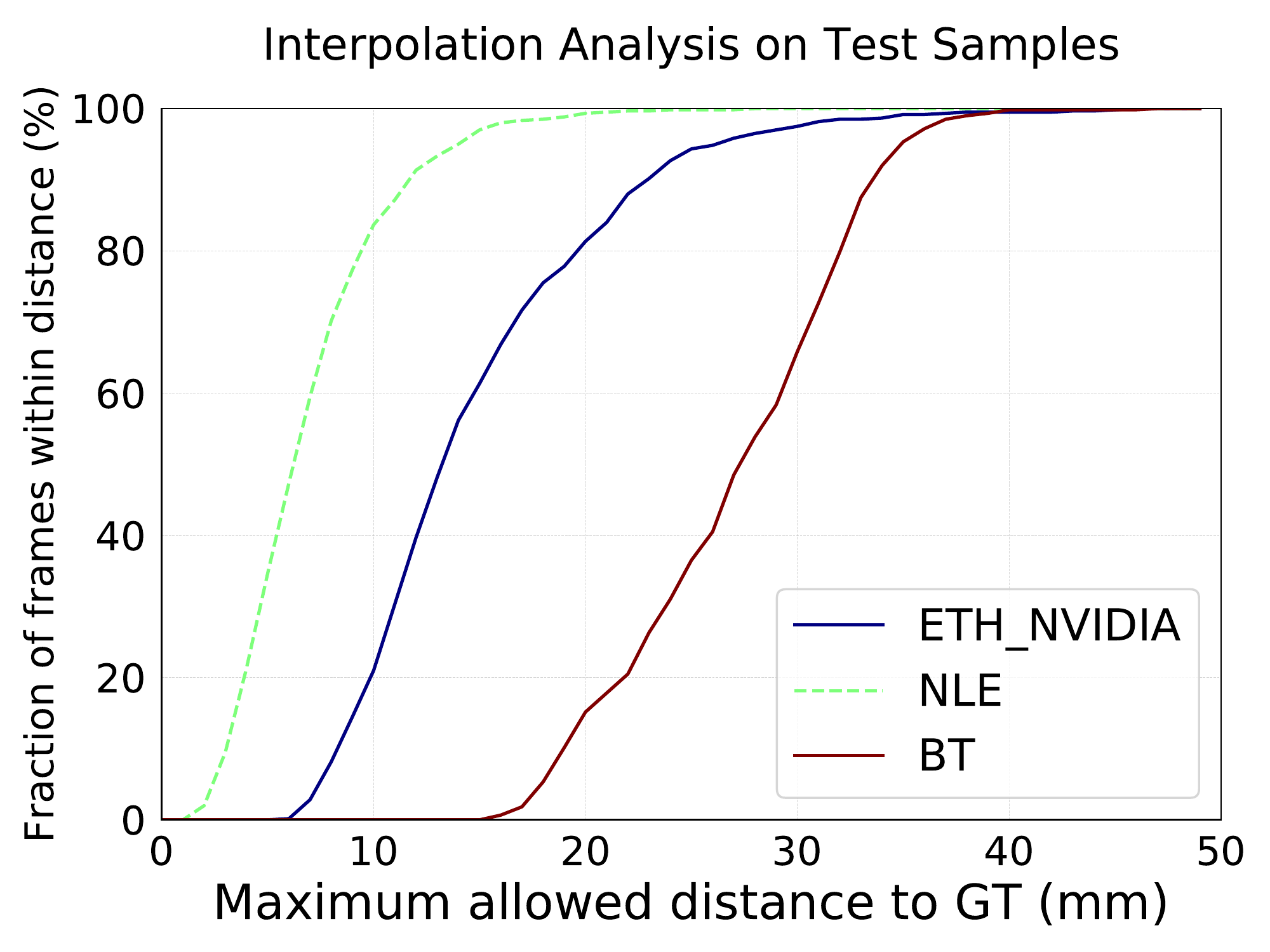}} & 
{\includegraphics[width=0.33\linewidth]{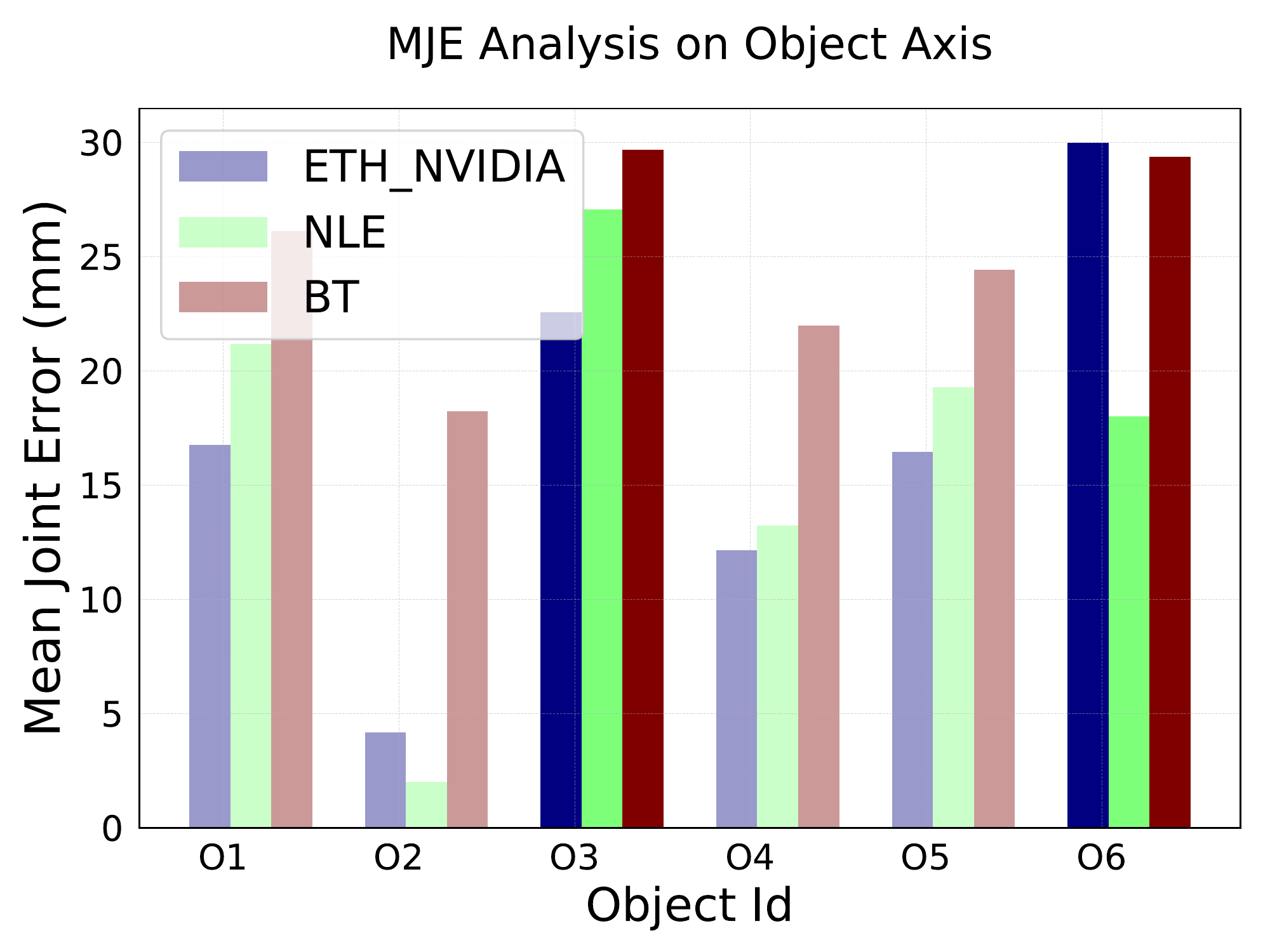}}
\\
{\small{(d) Object}} & {\small{(e) Interpolation}} & {\small{(f) Object MJE}}
\end{tabular}
\endgroup
\caption{Task 3 - Success rate analysis on the evaluation criteria (a,b,d,e) and MJE error analysis on the seen/unseen subjects (c) and objects (f). For (c) and (f), solid and transparent colors are used to depict extrapolation and interpolation.}
\label{fig:task3_extrapolation_plots}
\end{figure*}

\begin{figure}[h!]
\centering
\setlength{\tabcolsep}{4pt} 
\begin{tabular}{c c c c c}
\includegraphics[width=0.20\linewidth,height=0.20\linewidth]{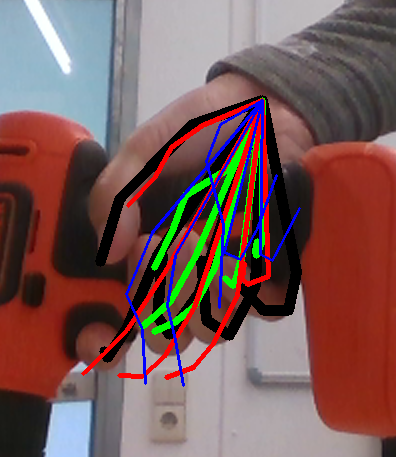} &
\includegraphics[width=0.20\linewidth,height=0.20\linewidth]{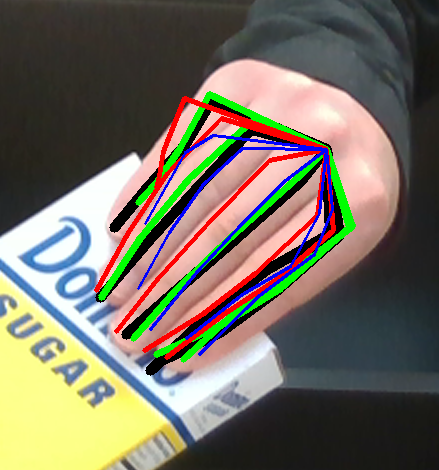} &
\includegraphics[width=0.20\linewidth,height=0.20\linewidth]{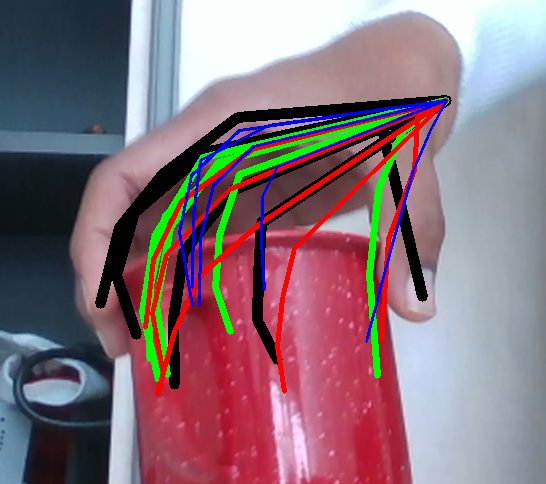} &
\includegraphics[width=0.20\linewidth,height=0.20\linewidth]{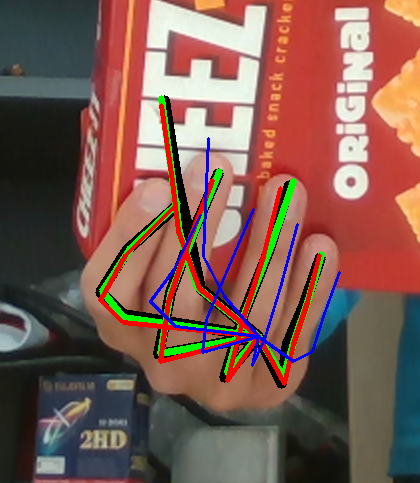}\\
\includegraphics[width=0.20\linewidth,height=0.20\linewidth]{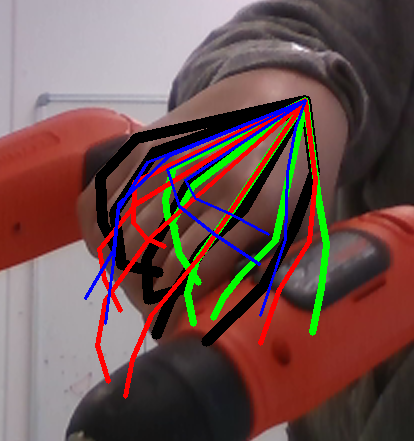} &
\includegraphics[width=0.20\linewidth,height=0.20\linewidth]{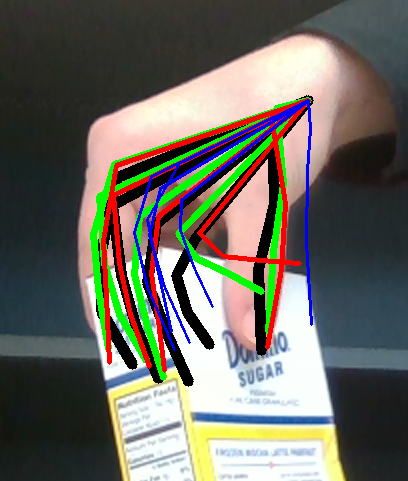} &
\includegraphics[width=0.20\linewidth,height=0.20\linewidth]{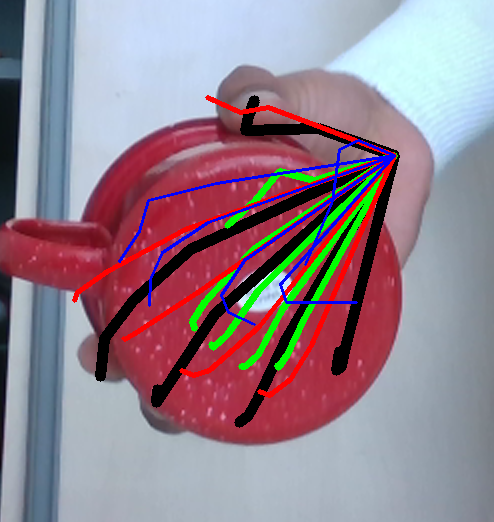} &
\includegraphics[width=0.20\linewidth,height=0.20\linewidth]{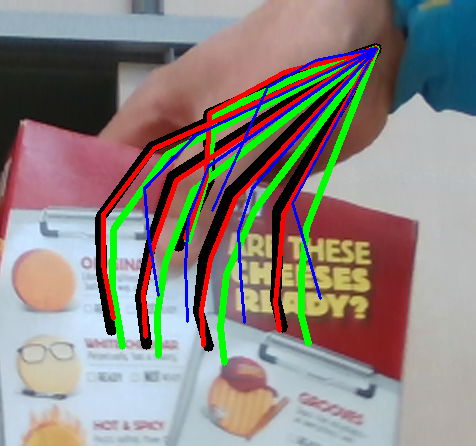}\\
{(a) Extrapolation} & {(b) Shape} & {(c) Object} & {(d) Interpolation}
\end{tabular}
\caption{Task 3 - Visualization of the \textbf{\color{black}{ground-truth}} annotations and estimations of \textbf{~\color{green}{\taskthreeone},~\color{red}{\taskthreetwo},~\color{blue}{\taskthreethree}}. Each column shows different examples used in our evaluation criteria.}
\label{fig:task3_qualitatives}
\end{figure} 

\noindent\textbf{Analysis of Submitted Methods for Task 3.}
\label{sec:task3_analysis}
We selected 3 entries, with different key properties for this analysis. It is definitively harder for the participants to provide accurate poses compared to the previous tasks. None of the methods can estimate frames that have all joints estimated with less than $25mm$ error, see Fig.~\ref{fig:task3_extrapolation_plots}. The $25mm$ distance threshold shows the difficulty of estimating a hand pose accurately from RGB input modality even though the participants of this task were provided with the ground-truth wrist joint location.

The task is based on hand-object interaction in RGB modality. Therefore, the problem raises the importance of multi-modal data and learn from different modalities. Only \taskthreethree uses the MANO parameters provided by the organizers to synthesize 100K images and adds random objects near the hand. This approach supports the claim on the importance of multi-modality and filling the real data gaps with synthetic data with its close performance to the two higher ranked methods in MJE.

The generalisation performance of \taskthreethree in Task 3 compared to the team's approaches with similar gist in Tasks 1 and 2 supports the importance of multi-model learning and synthetic data augmentation. The close performance of the method to generalise to unseen objects compared to \taskthreeone and to generalise to unseen shapes compared to \taskthreetwo also supports the argument with the data augmentation. The approach is still outperformed in MJE for this task although it performs close to the other methods.

\taskthreetwo's approach shows the impact of learning to estimate 2D joints+3D joints ($28.45mm$) compared to learning 3D joints alone ($37.31mm$) on the Object as well as improvements for the Interpolation. Object performance is further improved to $23.22mm$ with PPI integration. Further insights put by \taskthreetwo's own experiments on the number rotation augmentations ($n$) in post-processing helps to better extrapolate for unseen shapes ($17.35mm$, $16.77mm$, $15.79mm$ where $n=1, 4, 12$, respectively).

\noindent\textbf{Analysis on the Usage of Synthetic Images.}
The best performing method of Task 1 (\taskoneone) in MJE uses the 3D hand model parameters to create 570K synthetic images by either perturbing (first stage) the model parameters or not (second stage). Synthetic data usage significantly helps in training the initial model (see Fig.~\ref{fig:zzzh_mixed_depth}). Table~\ref{tab:zzzh_mano_ratio} shows the impact of different proportions of the 570K synthetic data usage to train the model together with the real training images. Using synthetic data can boost such a simple 3D joint regressor's performance from MJE of $30.11mm$ to $15.73mm$, a $\sim50\%$ improvement. Moreover, \taskoneone's experiments with a regression model trained for 10 epochs shows the impact of the mixed depth  inputs, Fig.~\ref{fig:zzzh_mixed_depth}, to lower the total extrapolation error ($26.16mm$) compared to the use of raw depth renderings ($30.13mm$) or the renderings averaged ($31.92mm$) with the real input images. \tasktwofour uses synthetic images in a very small amount of $32K$ and $100K$ in Tasks 2 and 3 since 3D reconstruction is difficult to train at a larger scale. However, favorable impact the data can be observed by comparing performances in Tasks 1 and 2.


%% file: input_supp_analysis.tex
\definecolor{mypurple}{rgb}{0.6,0.,0.6}
\noindent\textbf{Analysis on Evaluation Criteria}
We discuss the generalisation power of the methods based on our evaluation criteria below. Fig.~\ref{fig:task1_extrapolation_plots} (f-i) shows the average errors obtained on the different evaluation axis based on if the evaluation criterion has seen in the training set or not. Overall, while unseen shapes and viewpoints are harder to extrapolate in most of the cases, some unseen articulations are easier to extrapolate than some seen articulations which are hard to estimate the hand pose from. 

\noindent\textit{Viewpoint extrapolation.} 
HPEs tend to have larger errors on extreme angles like $[-180, -150]$ or $[150, 180]$ for azimuth viewpoint or similarly in elevation viewpoint and it's harder to extrapolate to unseen viewpoints in the training. While the approach by~\taskoneone~fills those unseen gaps with the generated synthetic data, other approaches mostly rely on their ensemble-based methodologies or their 3D properties.

Both Fig.~\ref{fig:task1_extrapolation_plots} (g) for azimuth angles and (h) for elevation angles show the analysis for the viewpoints. Most of the extrapolation intervals (except the edges since both edges used to evaluate extrapolation) show distributions similar to a Gaussian which is expected since the mid-intervals are most far away viewpoints from a seen viewpoint from the training set. While both elevation and azimuth extrapolation errors are always higher than the interpolation error obtained with the corresponding methods, however the azimuth extrapolation tends to be varying more than the elevation extrapolation for some angles. 

\noindent\textit{Articulation extrapolation.} Fig.~\ref{fig:task1_extrapolation_plots} (i) shows the average errors for 32 articulation clusters. 16 of those clusters have already seen in the training set while 16 have never seen and only available in the test set. While the samples that fall into some clusters, (\eg 16, 18, 19, 20 and 31) tend to be harder to estimate most of the time, however some articulations without depending on seen (\eg 1, 7, 8, 17) or unseen are hard to estimate as well because of the type of the articulation. Fig.~\ref{fig:task1_articulation_cluster_examples} shows the example frames for the 32 clusters. 

\noindent\textit{Shape extrapolation.} Fig.~\ref{fig:task1_extrapolation_plots} (f) shows average errors obtained for different shape types seen/unseen. All approaches have higher errors on unseen hand shapes (2, 3, 4, 5, 9) compared to errors obtained on shapes (1, 6, 7, 8, 10) seen in the training set.

Fig.~\ref{fig:task3_extrapolation_plots} (c, f) show the MJE analysis based on seen/unseen shapes (c) and objects (f). A list of objects that appear in the task test set is given in Table~\ref{fig:task3_object_list}. Although shape 'S5' refers to an unseen shape, all methods can extrapolate to this shape better than some other seen shapes in the training set. This can be explained with 'S5' being similar to some other shapes and it has the lowest number of frames (easy examples) compared to number of test frames from other shapes in the test set, see Fig.~\ref{fig:dataset_dist} (bottom right) for the distributions of the training and test set. A similar aspect has been observed in~\cite{yuan20172017} where different hand shape analysis has been provided, see Fig.~\ref{fig:hands17_shape_analysis}. However, all methods tend to have higher errors on the frames from another unseen test shape 'S3' as expected. 

\noindent\textit{Object extrapolation.} Poses for hands with unseen objects, 'O3' power drill and 'O6' mug, are harder to extrapolate by most methods since their shapes are quite different than the other seen objects in the training set. Please note that seen 'O2' object has the lowest number of frames in the test set. Some example frames for the listed objects are showed in Fig.~\ref{fig:ho3d_objects}.

\section{Ablation Studies by the Participants}
\label{sec:ablation}

Here we present the experiments and their results conducted by the participants for the challenge. Section~\ref{sec:backbones} presents experimental results conducted by the participated approaches based on different backbone architectures and similarly, Section~\ref{sec:ensemble} shows experimental evaluation on the ensembling techniques in pre-processing, post-processing and methodological level. 




\subsection{Experiments with Different Backbone Architectures}
\label{sec:backbones}
While Residual Network (ResNet)~\cite{resnet50} backbones are well adopted by many approaches and ResNet-50 or ResNet-101 architectures obtain better results compared to other backbone models as reported in experiments of~\taskonethree~and~\taskthreetwo. However, most approaches adopt ensembling predictions from models trained with different backbone architectures and this improves the performance as showed in Tables~\ref{tab:awr_backbone_results} and~\ref{tab:nplwe_backbones}. 
\\
\begin{table}[!htb]
    \caption{Extrapolation MJE obtained with different backbone architectures in~\taskonethree~experiments. 'center1' denotes using thresholds to compute hand center, 'center2 + original' denotes using semantic segmentation network to compute hand center and extract hand region from original depth images, 'center2 + segmented' denotes using semantic segmentation network to compute hand center while extract hand region from network's output mask.}
    \label{tab:awr_backbone_results}
    \centering
    \begin{tabular}{c | c}
        Backbone & Extrapolation MJE (mm) \\
        \hline
        Resnet50 (center1) & 20.70\\
        Resnet50 (center2 + original) & 14.89\\
        Resnet50 (center2 + segmented) & 14.75\\
        Resnet101 (center2 + original) & 14.57\\
        Resnet101 (center2 + segmented) & 14.44\\
        HRnet48 & 17.23\\
        SRN & 16.00\\
        SRN\_multi\_size\_ensemble & 15.20\\
        HRNet\_Resnet50\_shape\_ensemble & 14.68\\
        model\_ensemble & \textbf{13.67}\\
    \end{tabular}
\end{table} 

Table~\ref{tab:awr_backbone_results} shows the experiments for impact of different network backbones and different ways of obtaining the hand center by~\taskonethree. Changing the way of attaining hand center from 'center1' to 'center2 + original' yields an improvement of $5.81mm$, 'center2 + segmented' further improves by $0.14mm$. The best result is obtained with a backbone of ResNet-101, $14.44mm$.

At the final stage, multiple models are ensembled (model\_ensemble in Table~\ref{tab:awr_backbone_results}) including ResNet-101 (center2+segmented), ResNet-101 (center2+original), ResNet-50 (center2+original), {SRN\_multi\_size\_ensemble} and HRNet\_Resnet50\-\_shape\_ensemble. Since ESPNetv2~\cite{espnet} sacrifices accuracy for speed to some extent, the segmentation results are not accurate enough and may contain wrists or lack part of the fingers, therefore cropping hand regions from original depth images sometimes yields better performance. 

Among the approaches using ensembled networks, SRN~\cite{srn} is a stacked regression network which is robust to self-occlusion and when depth values are missing. It performs the best for Shape extrapolation, but is sensitive to the cube size that are used when cropping hand region. The mean error of a single-stage SRN with cube size 200mm already reaches $16mm$. Ensembling SRN with cube size $180mm$, $200mm$ and $220mm$, the results of SRN\_multi\_\-size\_ensemble is 15.20mm. 

SRN performs the best on the shape evaluation axis. For example, single SRN can achieve $12.32mm$ and SRN\_multi\_size\_ensemble can achieve $11.85mm$.

HRNet-48 makes a major success in human pose estimation, but we do not get desired results after applying it. The mean error of single HRNet-48 is $17.23mm$. Although it converges faster and has relatively lower loss than ResNet-50 and ResNet-101 in the training stage, it performs worse during inference. HRNet-48 predicts well on some of the shapes. Therefore, the depth images are divided into 20 categories according to the proportion of hand pixels over all pixels. The prediction error in training set is used to compute the weight of each category, which is used to weight the test set results. The weighted results depicted with HRNet\_Resnet50\-\_shape\_ensemble reaches mean error of $14.68mm$. 

The model\_ensemble refers to ensembling predictions of five models including ResNet-101 ($14.44mm$), ResNet-101\_noseg ($14.57mm$), ResNet-50\_noseg ($14.89mm$), HRNet\_Resnet50\-\_shape\_ensemble ($14.68mm$), SRN\_multi\_size\-\_ensemble ($15.20mm$). Among them, the first four models are based on adaptive weighting regression (AWR) network with different backbones. 
\\
\begin{table}[!h]
    \caption{Impact of different network architectures, in~\taskthreetwo~experiments. No color jittering is applied during training in these experiments. MJE (mm) metric is used. Please note that for this experiment while ResNet-50 and ResNet-152 backbones results are obtained with 10 different anchor poses while the rest use 5 different anchor poses in~\taskthreetwo' settings for Pose Proposal Integration (PPI).}
    \label{tab:nplwe_backbones}
    \centering
    \begin{tabular}{c | c c c c}
        Backbone & Extrapolation & Interpolation & Object & Shape \\
        \hline
        ResNet-50 & 34.63 & 5.63 & 23.22 & \textbf{17.79}\\
        ResNet-101 & \textbf{32.56} & 4.49 & \textbf{18.68} & 18.50\\
        ResNet-152 & 37.56 & 4.24 & 20.11 & 18.58\\
        ResNext-50 & 33.88 & 4.99 & 25.67 & 19.70\\
        ResNext-101 & 38.09 & \textbf{3.83} & 21.65 & 20.93
    \end{tabular}
\end{table} 

Table~\ref{tab:nplwe_backbones} shows comparison of different residual based backbones. Deeper backbones can obtain lower errors on Interpolation however, the method obtains higher errors on Extrapolation criteria and ResNet-101 a medium depth seems to be a reasonable choice in most cases in~\taskthreetwo~experiments. While errors on different evaluation criteria with ResNext based arthictectures tend to vary a lot, ResNet based backbones are more solid. 
\\
\begin{table}[!h]
    \caption{Impact of widening the architecture used in V2V-PoseNet~\cite{v2vposenet} in~\tasktwoone~experiments. The number of kernels in each block in V2V-PoseNet architecture is quadrupled (wider).}
    \label{tab:ntis_v2v_deeper}
    \centering
    \begin{tabular}{c | c c}
        Architecture V2V-PoseNet~\cite{v2vposenet} & Extrapolation MJE (mm)\\
        \hline
        Original & 38.33 \\
        Wider  & \textbf{36.36}
    \end{tabular}
\end{table} 
Components of V2V-PoseNet architecture include: Volumetric Basic Block, Volumetric Residual Block, and Volumetric Downsampling and Upsampling Block.~\tasktwoone~uses the same individual blocks as in V2V-PoseNet~\cite{v2vposenet} but with a wider architecture.~\tasktwoone' experiment, see Table~\ref{tab:ntis_v2v_deeper} shows that quadrupling the number of kernels in individual blocks provides the best results.

\subsection{Impact of Ensembling Techniques}
\label{sec:ensemble}
In this section, we provide the experiments to show the importance of ensembling techniques. These techniques include ensembling in data pre-processing, methodological ensembles and ensembles as post-processing.

\noindent\textit{\taskthreetwo' experiments on methodological and post-processing ensembling techniques.}
~\taskthreetwo~adopts an approach based on LCR-Net++\cite{lcrnetpp} where poses in the training set are clustered to obtain anchor poses and during inference, the test samples are first classified to these anchors and the final hand pose estimation is regressed from the anchor poses. Table~\ref{tab:nplwe_anchor_poses} shows the impact of using different number of anchor poses. Shape extrapolation axis is heavily affected with the number anchor poses. While the number of obtained anchor poses from the training set increases from 1 to 50, the shape extrapolation error decreases from $21.08mm$ to $16.55mm$. On the other hand, the number of anchor poses does not seem to have an observable impact on the other criteria, however; this can be because of the size of Task 3 test set and also because of the low hand pose variances in Task 3.
\\
\begin{table}[!htb]
    \caption{Impact of number of anchor poses, in~\taskthreetwo~experiments, obtained with k-means clustering for Pose Proposal Integration (PPI). No color jittering is applied during training in these experiments. ResNet-101 backbone architecture and MJE (mm) metric is used.}
    \label{tab:nplwe_anchor_poses}
    \centering
    \begin{tabular}{c | c c c c}
        \#Anchor poses & Extrapolation & Interpolation & Object & Shape \\
        \hline
        1 & 37.68 & \textbf{3.99} & 28.69 & 21.08\\
        5 & \textbf{32.56} & 4.49 & 18.68 & 18.50\\
        10 & 37.57 & 4.35 & 19.38 & 18.33\\
        20 & 34.67 & 4.38 & 21.10 & 16.94\\
        50 & 35.64 & 4.86 & \textbf{17.84} & \textbf{16.55}
    \end{tabular}
\end{table} 

\begin{table}[]
    \caption{Importance of pose proposal integration~\cite{lcrnetpp} (PPI) compared to non-max suppression (NMS), and of joint 2D-3D regression in~\taskthreetwo~experiments (ResNet-50 backbone and 5 anchor poses are used). MJE (mm) metric is used.}
    \label{tab:nplwe_objectextrapolation}
    \centering
    \begin{tabular}{c | c | c c c c}
        2D-3D Estimation & Post. & Extrapolation & Interpolation & Object & Shape\\
        \hline
        3D only & NMS & 38.59 & 8.48 & 37.31 & 18.78\\
        2D+3D & NMS & 38.08 & 7.60 & 28.45 & 18.73\\
        2D+3D & PPI & \textbf{34.63} & \textbf{5.63} & \textbf{23.22} & \textbf{17.79}
    \end{tabular}
\end{table}

~\taskthreetwo's experiments later show the impact of learning and inferencing both 2D and 3D pose, and the impact of pose proposal integration~\cite{lcrnetpp} (PPI) compared to non-maximum suppression approach to obtain the poses. Learning to estimate 2D pose of a hand significantly impacts the extrapolation capability especially in Object axis. We believe this is because the objects occlude the hands and 2D information can be better obtained and help to guide estimation of the 3D hand poses. Later the pose proposal with 5 anchor poses brings a significant boost for extrapolation capabilities of the method.
\\
\begin{table}[]
    \caption{Importance of rotation data augmentation in~\taskthreetwo~experiments, conducted with a ResNet-101 backbone architecture and 5 anchor poses. MJE (mm) metric is used.}
    \label{tab:nplwe_rotation}
    \centering
    \begin{tabular}{c | c c c c}
        \#Test Rot. & Extrapolation & Interpolation & Object & Shape \\
        \hline
        1 & 29.55 & 4.85 & 18.09 & 17.35\\
        4 & \textbf{28.83} & 4.63 & \textbf{18.06} & 16.77\\
        12 & 29.19 & \textbf{4.06} & 18.39 & \textbf{15.79}
    \end{tabular}
\end{table} 

~\taskthreetwo~adopts another ensembling technique in the post-processing stage where test images are rotated by uniformly covering the space and the predictions obtained from each rotated test sample is ensembled. Experiments of~\taskthreetwo~show that rotation as a post-processing ensemble technique helps significantly on shape extrapolation as well as interpolation axis and has minor impacts on other extrapolation criteria. Table~\ref{tab:nplwe_rotation} shows the impact of different number of rotation ensembles.

\noindent\textit{\taskonefive~ensembling as data pre-processing and orientation refinement per limb.}
\taskonefive~makes use of different input types obtained from the depth input image and their combinations to use them in their approach. Different input types include 3D joints projection, multi-layer depth and voxel representations and a list of input types and their combinations adopted to train different models are listed in Table~\ref{tab:strawberryfg_multimodel_inputs}. The impact of each mentioned model is reported in Table~\ref{tab:strawberryfg_multimodel_results}. The model used with different combination of different input types obtained from the depth images has no significant impact on evaluation criteria. We believe that this is because each different input type has different characteristics for the model to learn from and it's hard for the model to adapt to each type. Maybe a kind of adaptive weighting technique as adopted by some other approaches participated in the challenge can help in this case. However, as ensembling results of different models is proven to be helpful with all the approaches adopted the technique seems to be helpful in this case as well. 'Combined' model as depicted in Table~\ref{tab:strawberryfg_multimodel_results} obtains the best results for all evaluation criteria.~\taskonefive' experiment report to have $10.6\%$ on articulation, $10\%$ on interpolation, $8.4\%$ on viewpoint, $7.2\%$ on extrapolation, $6.2\%$ on shape criteria improvements with ensembling of 4 models.

Table~\ref{tab:strawberryfg_refinement_impact} using \taskonefive~ shows the impact of patch orientation refinement networks adopted for each limb of a hand to show the impact. Orientation refinement brings a significant impact with $~1mm$ lower error on all evaluation criteria.
\\
\begin{table}[h]
    \caption{Input data types for four different models used in~\taskonefive~experiments.}
    \label{tab:strawberryfg_multimodel_inputs}
    \centering
    \begin{tabular}{c | c c c c}
        Model Id & & Input Type & & \\
        \hline
        & Depth Image & 3D Points Projection & Multi-layer Depth & Depth Voxel\\
        \hline
        1&\cmark&\xmark&\xmark&\xmark\\
        2&\cmark&\cmark&\cmark&\xmark\\
        3&\cmark&\cmark&\xmark&\cmark\\
        4&\cmark&\cmark&\cmark&\cmark\\
    \end{tabular}
\end{table} 
\\
\begin{table}[h]
    \caption{MJE (mm) obtained in~\taskonefive~experiments by using different models trained with different input types, see Table~\ref{tab:strawberryfg_multimodel_inputs}. 'Combined' model refers to ensembling predictions from all 4 models.}
    \label{tab:strawberryfg_multimodel_results}
    \centering
    \begin{tabular}{c | c c c c c}
        Model Id & Extrapolation & Viewpoint & Articulation & Shape & Interpolation \\
        \hline
        1 & 20.99 & 14.70 & 8.42 & 14.85 & 9.35\\
        2 & 21.39 & 15.34 & 8.25 & 15.21 & 9.17\\
        3 & 21.02 & 16.12 & 8.52 & 15.30 & 9.61\\
        4 & 21.19 & 15.78 & 8.36 & 15.23 & 9.32\\
        Combined & \textbf{19.63} & \textbf{14.16} & \textbf{7.50} & \textbf{14.21} & \textbf{8.42}
    \end{tabular}
\end{table} 
\\
\begin{table}[h]
    \caption{Impact of local patch refinement and volume rendering supervision adopted by~\taskonefive~. Model 4 with 4 different inputs are used in this evaluation, see Table~\ref{tab:strawberryfg_multimodel_inputs}.}
    \label{tab:strawberryfg_refinement_impact}
    \centering
    \resizebox{1.\linewidth}{!}{
    \begin{tabular}{c | c c c c c}
        Model Id - Type & Extrapolation & Viewpoint & Articulation & Shape & Interpolation \\
        \hline
        4 - w/o refinement \& volume rendering & 22.56 & 16.77 & 9.20 & 15.83 & 10.15\\
        4 - w/ refinement \& volume rendering & \textbf{21.19} & \textbf{15.78} & \textbf{8.36} & \textbf{15.23} & \textbf{9.32}\\
    \end{tabular}
    }
\end{table} 

\noindent\textit{\taskonetwo~uses ensembling in post-processing.}
At inference stage,~\taskonetwo~applies rotation and scale augmentations. More specifically,~\taskonetwo~rotates the test samples with $-90^{\circ}$/$45^{\circ}/90^{\circ}$ , and scales with factor $1/1.25/1.5$. Then these predictions are averaged. Several backbone models are trained, including ResNet-50/101/152, SE-ResNet-50/101, DenseNet-169/201, EfficientNet-B5/B6/B7. Input image sizes are $256\times256$/$288\times288$/$320\times320$/$384\times384$. The best single model is ResNet-152 with input size $384\times384$, it achieves $14.74mm$ on the extrapolation axis. Finally, these predictions are ensembled with weights to obtain a final error of $13.74mm$ on the extrapolation axis. 

\noindent\textit{\taskonefour~ensembling in post-processing with confident joint locations, Truncated SVDs and temporal smoothing.}
\taskonefour~ adopts a post-processing technique for refinement of hand poses where several inverse transformations of predicted joint positions are applied; in detail,~\taskonefour~uses truncated singular value decomposition transformations (Truncated SVDs; 9 for Task 1 and 5 for Task 2) with number of components $n \in {10, 15, 20, 25, 30, 35, 40, 45, 50}$ obtained from the training ground-truth hand pose labels and prepares nine refined pose candidates. These candidates are combined together as final estimation that is collected as weighted linear combination of pose candidates with weights $w \in {0.1, 0.1, 0.2, 0.2, 0.4, 0.8, 1.0, 1.8}/4.7$. Table~\ref{tab:ntis_svd_impact} shows the impact of ensembling confident joint predictions and refinement stage with Truncated SVDs.\\

\begin{table}[!h]
    \caption{Impact of refinement with Truncated SVDs in~\taskonefour~experiments on Task 1. Improvement is \~1\%. $N=100$ most confident joint locations are ensembled for this experiment. Results reported in MJE (mm) metric.}
    \label{tab:ntis_svd_impact}
    \centering
    \begin{tabular}{c | c c}
        SVD refinement & Extrapolation \\
        \hline
        w/ & \textbf{15.81}\\
        w/o & 15.98
    \end{tabular}
\end{table}

Since Task 2 is based on sequences and test samples are provided in order,~\tasktwoone~applies temporal smoothing on the predictions from each frame and provides experimental results in Table~\ref{tab:ntis_temporal_smoothing_impact} with different context sizes for smoothing. While temporal smoothing helps to decrease the extrapolation error, large context sizes do not impact much on the error. 
\\
\begin{table}[!h]
    \caption{Impact of temporal smoothing and the context size (k) for smoothing in~\tasktwoone~experiments on Task 2 using exact same V2V-PoseNet~\cite{v2vposenet} architecture.}
    \label{tab:ntis_temporal_smoothing_impact}
    \centering
    \begin{tabular}{c | c}
        Smoothing Context Size (k) & Extrapolation MJE (mm) \\
        \hline
        0 & 39.76\\
        3 & 38.32\\
        5 & 38.31\\
        7 & 38.33\\
    \end{tabular}
\end{table} 

\begin{figure}[!h]
\centering
\centering {\includegraphics[width=0.4\linewidth]{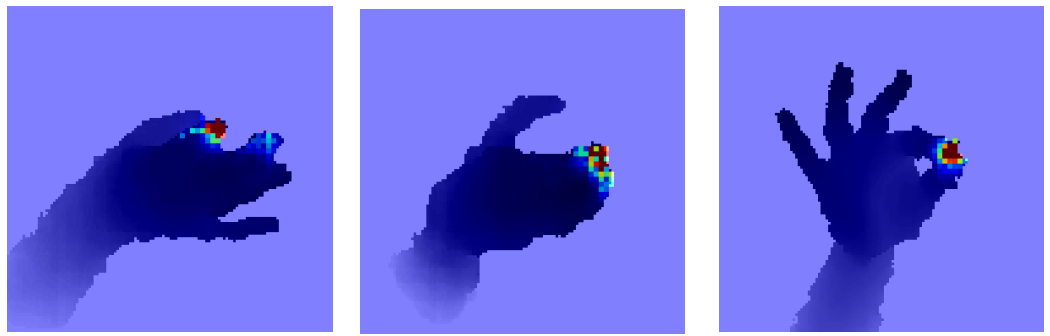}} \\
{(a) w/o AWR}\\
{\includegraphics[width=0.4\linewidth]{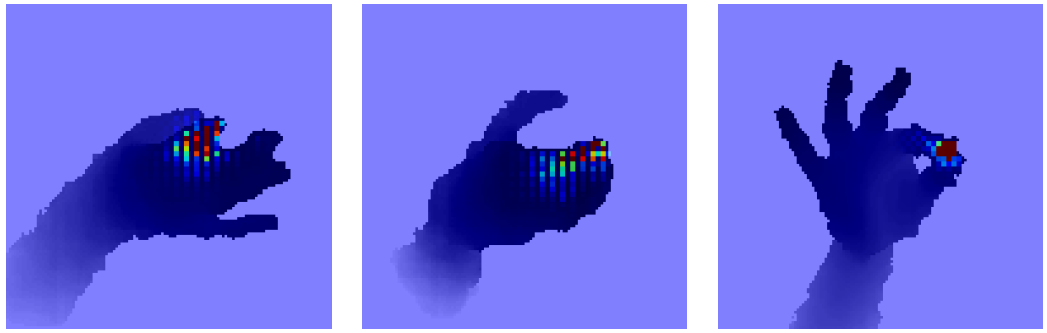}} \\
 {(b) w/ AWR} 
\caption{Impact of AWR operation on the target joints' probability maps.}
\label{fig:awr_impact}
\end{figure}

\noindent\textit{~\taskonethree methodological ensembling with AWR operation.} Fig.~\ref{fig:awr_impact} shows the impact of learnable adaptive weighting regression (AWR) approach on the probability maps of the target joints. When the target joint is visible and easy to distinguish, the weight distribution of AWR tends to focus more on pixels around it as standard detection-based methods do, which helps to make full use of local evidence. When depth values around the target joint are missing, the weight distribution spreads out to capture information of adjacent joint. Later, Table~\ref{tab:awr_impact} shows the impact of the AWR operation on two other datasets, NYU~\cite{tompson2014real} and HANDS'17~\cite{yuan20172017}.

\begin{table}[!h]
    \vspace{10pt}
    \caption{~\taskonethree~experiments for w/o adaptive weighting on NYU~\cite{tompson2014real} and HANDS'17~\cite{yuan20172017} datasets. Results reported in MJE (mm) metric.}
    \label{tab:awr_impact}
    \centering
    \begin{tabular}{c | c c}
        Dataset & w/o AWR & w/ AWR \\
        \hline
        NYU & 7.87 & \textbf{7.48}\\
        HANDS'17 & 7.98 & \textbf{7.48}
    \end{tabular}
\end{table}


%% file: appendix.tex
\onecolumn
\section{Appendix}
\label{sec:appendix}
\subsection{Frame Success Rates for All Participated Users in the Challenge}
\label{sec:appendix_all_methods}
\begin{figure*}[!htbp]
\centering
\begin{minipage}{1.\linewidth}
\centering
\begin{tabular}{c c}
{\includegraphics[width=0.4\linewidth]{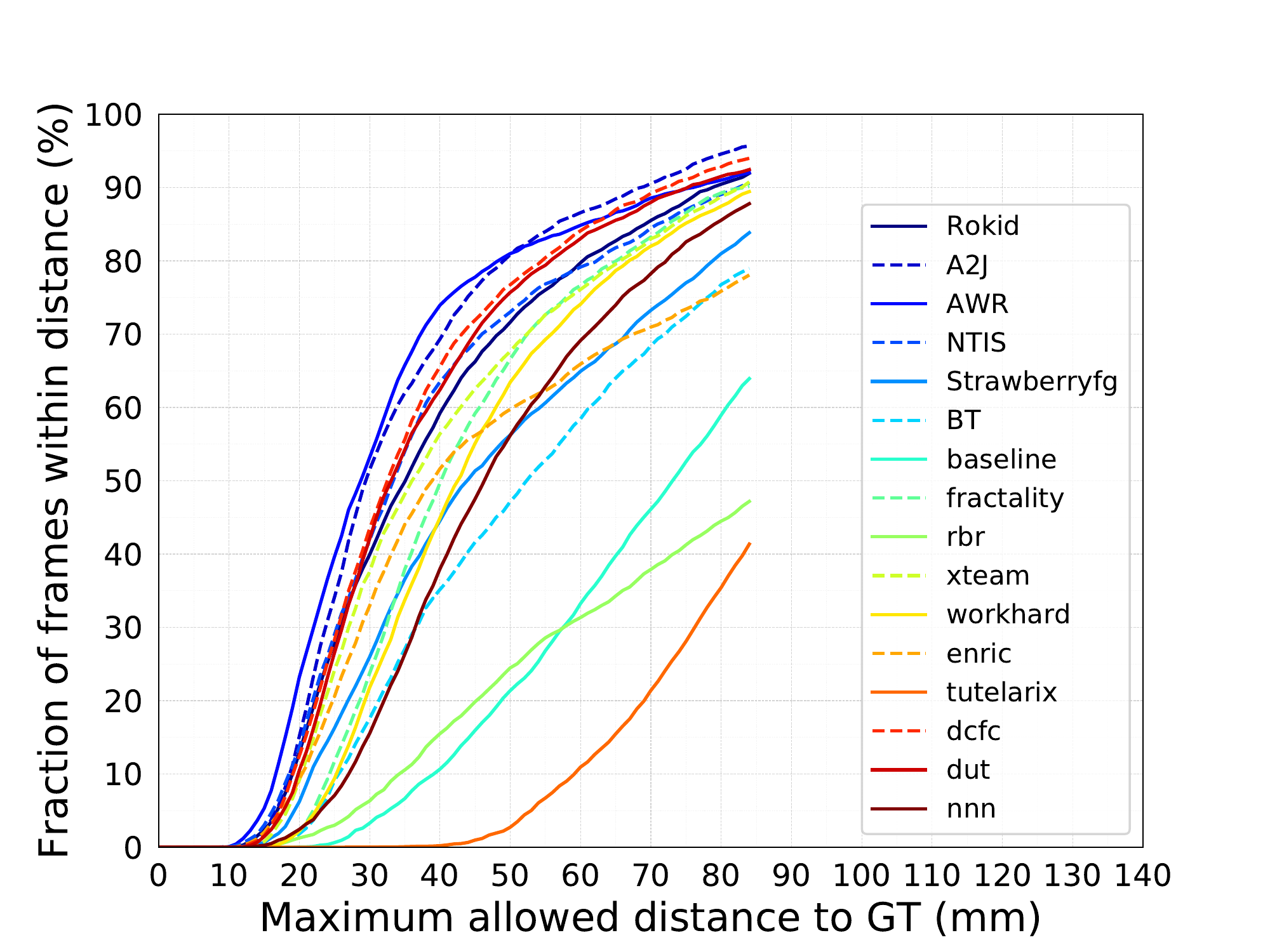}} & {\includegraphics[width=0.4\linewidth]{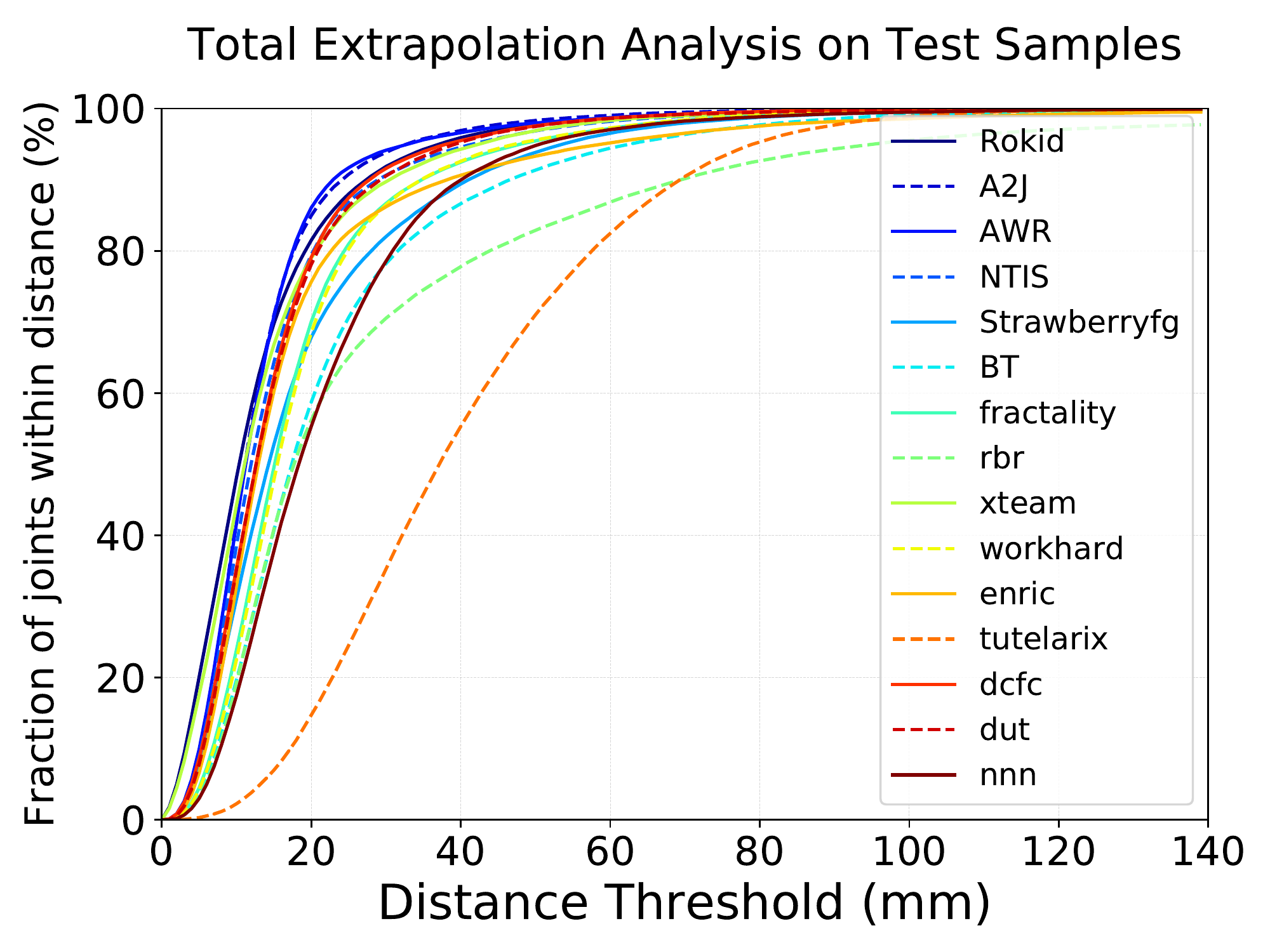}} \vspace{-0pt}\\
(a) Task 1 - Extrapolation frame &
(b) Task 1 - Extrapolation joint \vspace{-0pt}
\end{tabular}
\end{minipage} 
\\
\begin{minipage}{1.\linewidth}
\centering
\begin{tabular}{c c}
{\includegraphics[width=0.4\linewidth]{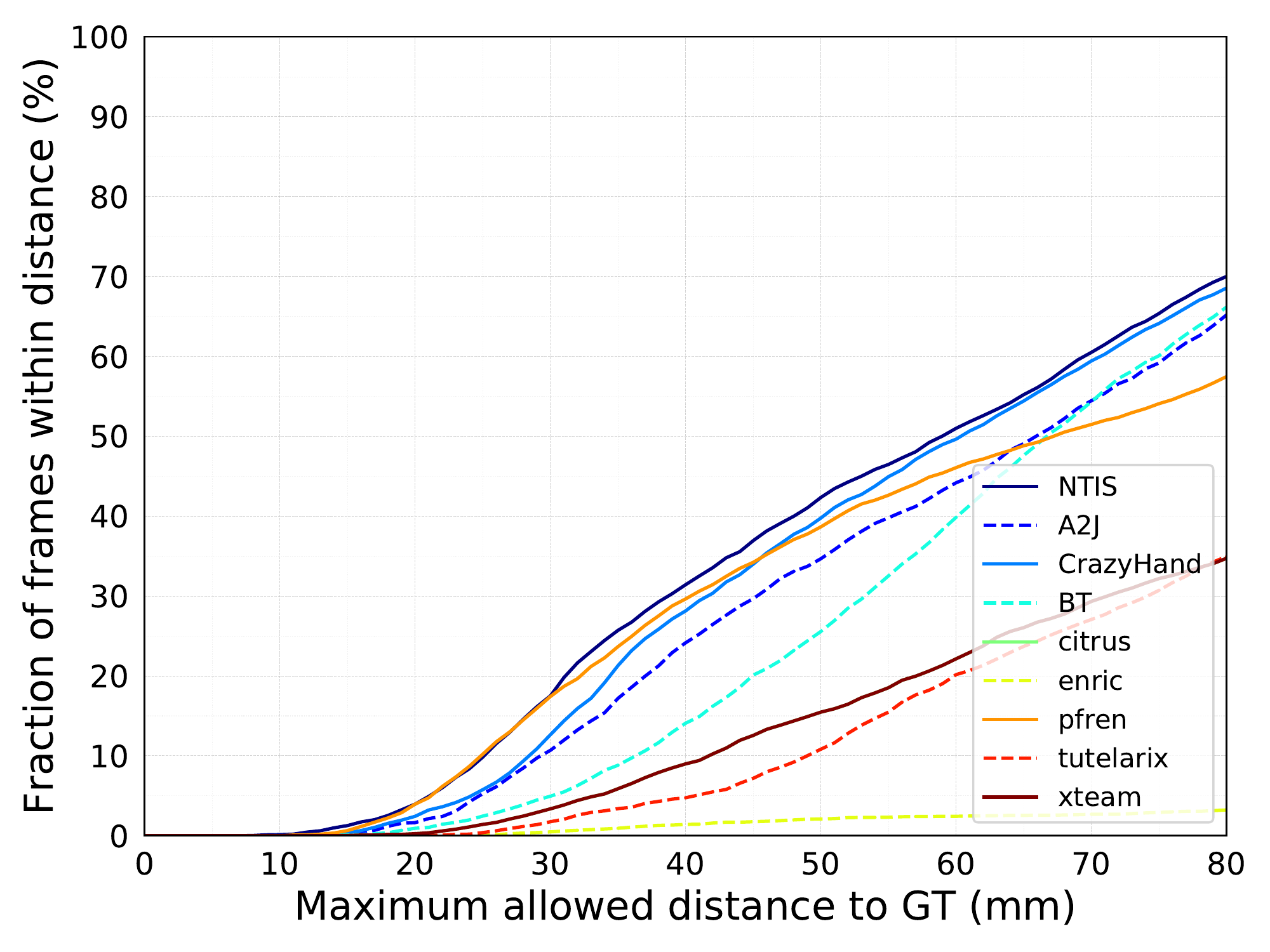}} & {\includegraphics[width=0.4\linewidth]{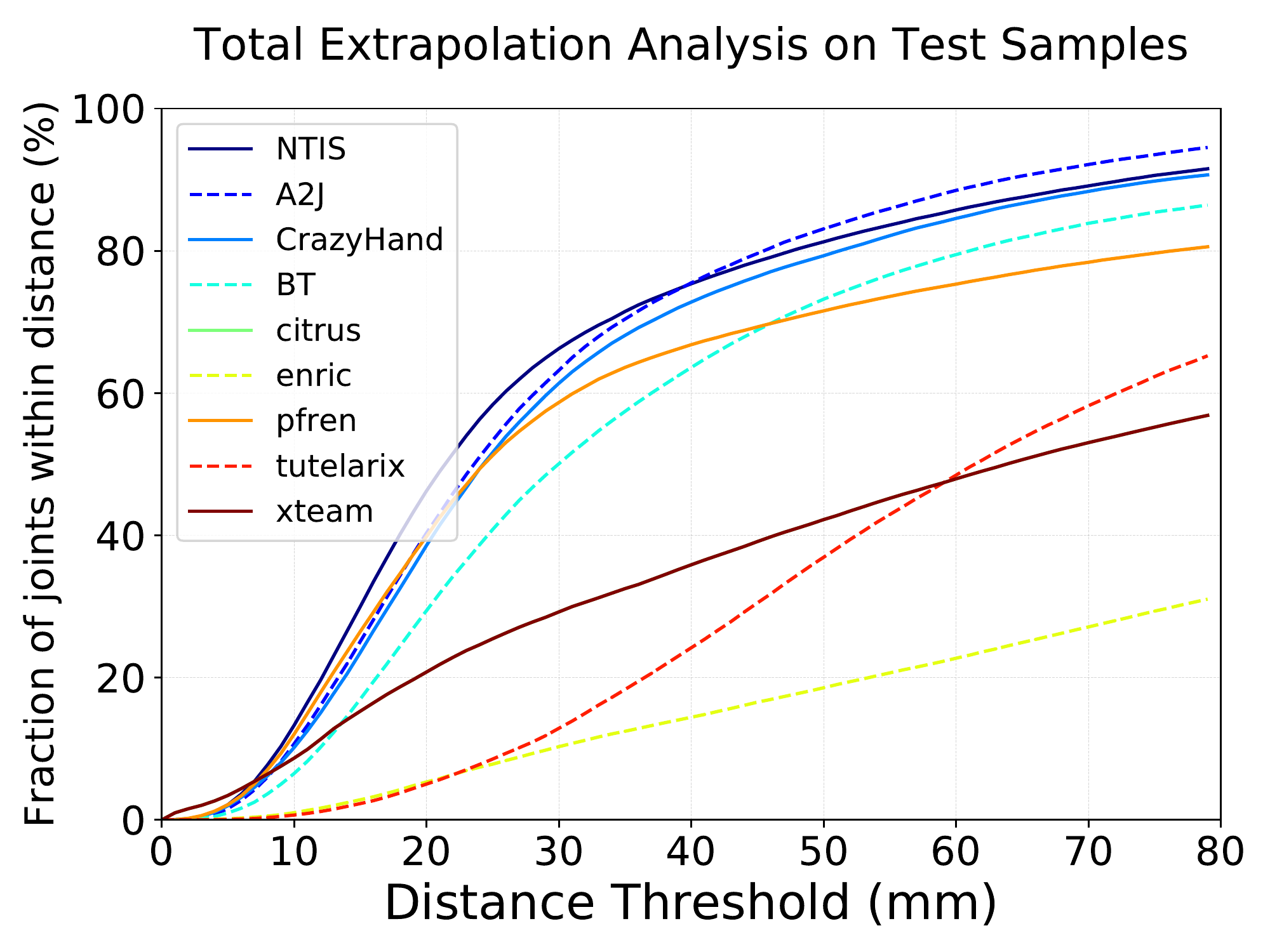}} \vspace{-0pt}\\
(c) Task 2 - Extrapolation frame &
(d) Task 2 - Extrapolation joint \vspace{-0pt}
\end{tabular}
\end{minipage} 
\\
\begin{minipage}{1.\linewidth}
\centering
\begin{tabular}{c c}
{\includegraphics[width=0.4\linewidth]{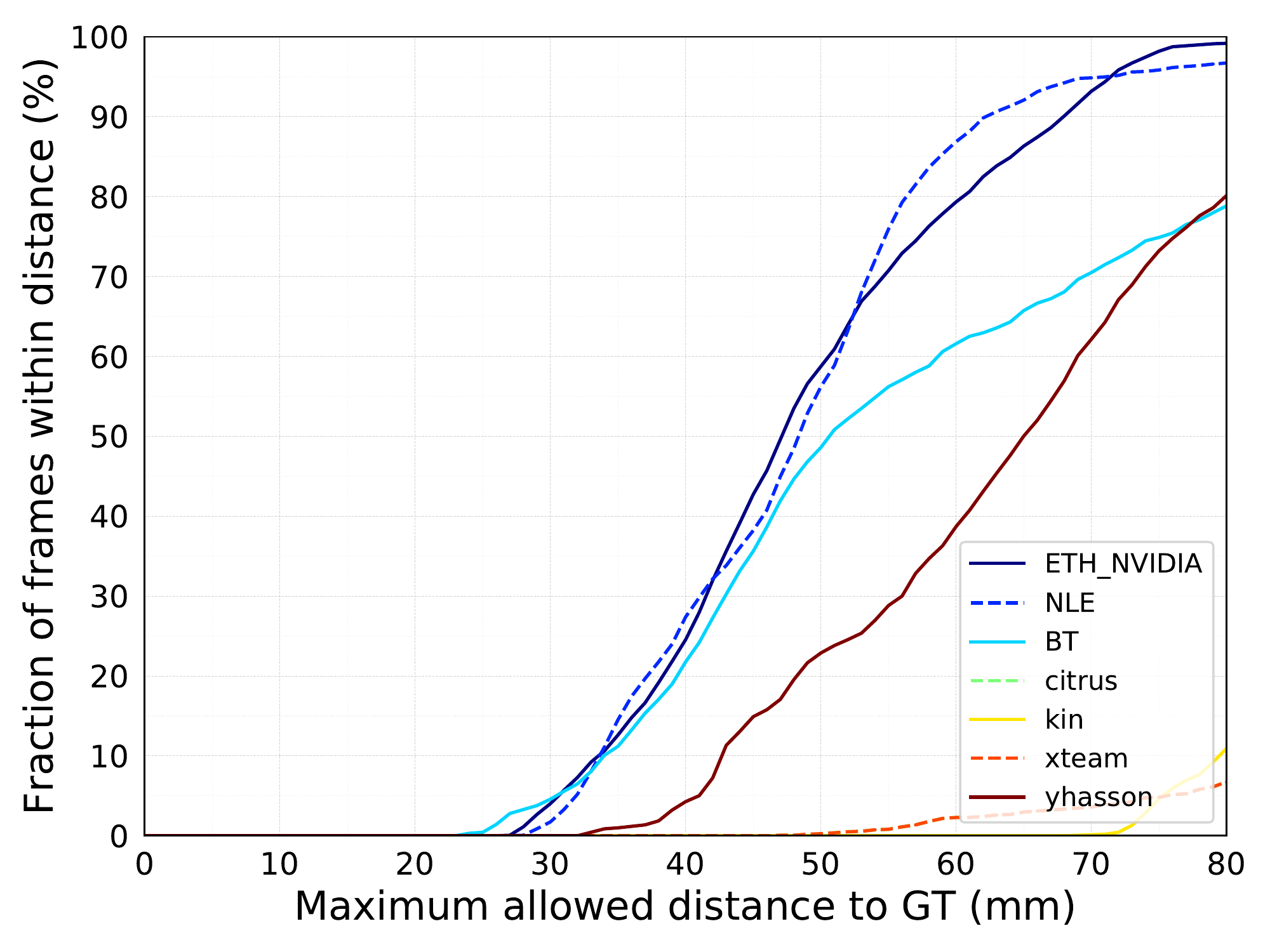}} & {\includegraphics[width=0.4\linewidth]{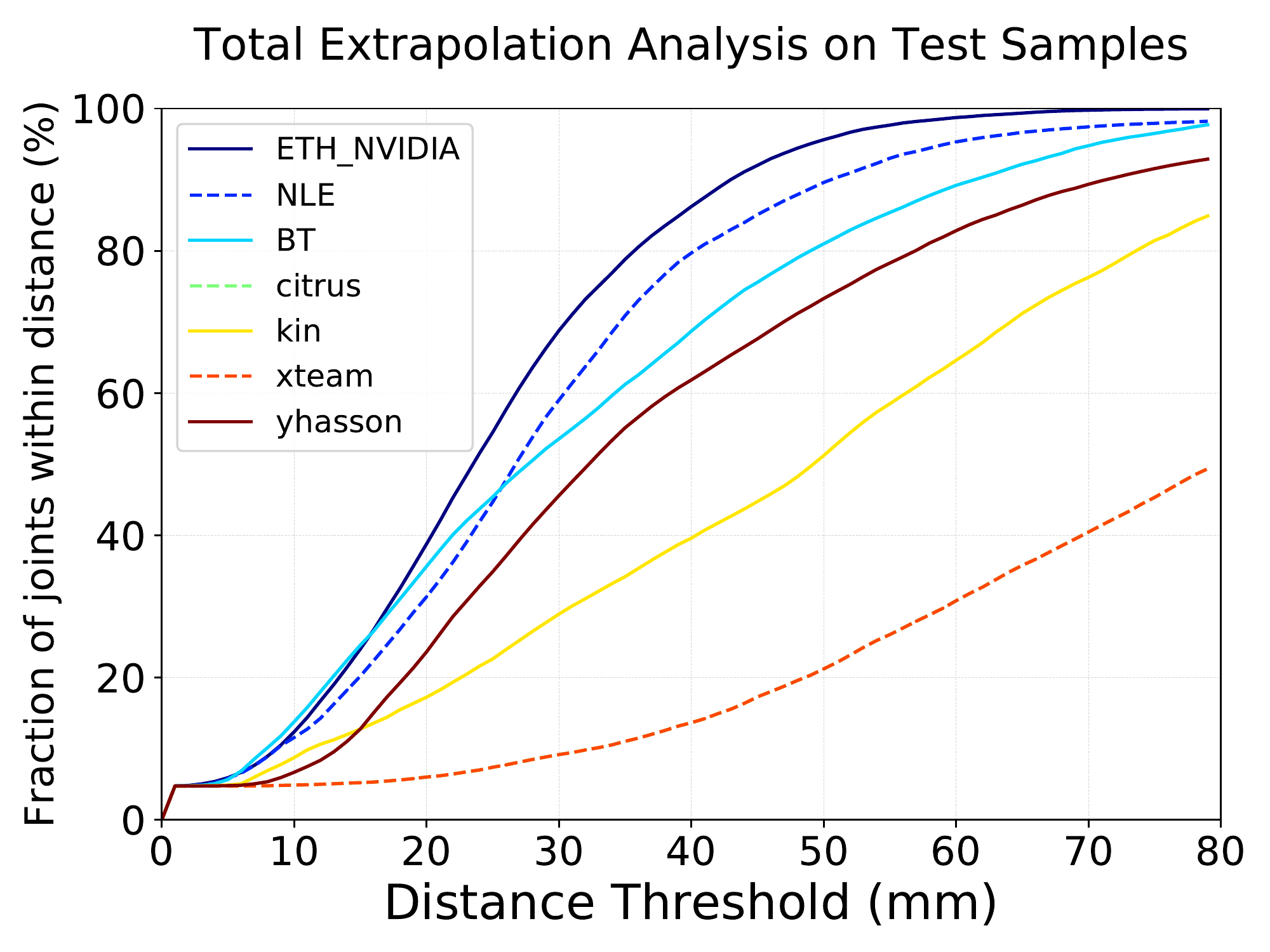}} \vspace{-0pt}\\
(e) Task3 - Extrapolation frame &
(f) Task3 - Extrapolation joint
\end{tabular}
\end{minipage} \vspace{-0pt}
\caption{All participated methods' total extrapolation accuracy analysis for each task. (a,c,e) represents the frame success rates where each frames' error is estimated by considering the maximum error of all joints in that frame. (b,d,f) shows the joint success rates.}
\label{fig:all_methods_extrapolation_plots}
\end{figure*}

Figure~\ref{fig:all_methods_extrapolation_plots} shows the analysis of all participated users in the challenge's tasks. We analysed the selected methods (6 for Task 1, 4 for Task 2 and 3 for Task 3) based on their methodological variances and results. The challenge have received 16 submissions for Task 1, 9 submissions for Task 2 and 7 for Task 3 to be evaluated from different users. 

\newpage
\subsection{Joint Success Rates of the Analysed Approaches}
\label{sec:appendix_joint_success_rates}
Success rate analyses for each of three tasks based on all joints in the test set are provided below. Please note the difference of the figures below compared to the success rate analysis based on frames as showed in Fig.~\ref{fig:task1_extrapolation_plots},~\ref{fig:task2_extrapolation_plots} and~\ref{fig:task3_extrapolation_plots}. Comparing the joint based analysis and the frame based analysis, we can note that all methods have different error variance for different joints and therefore the approaches tend to obtain higher accuracies based on considering each joint independently.
\begin{figure}[!htbp]
\centering
\begin{tabular}{c c}
{\includegraphics[width=0.4\linewidth]{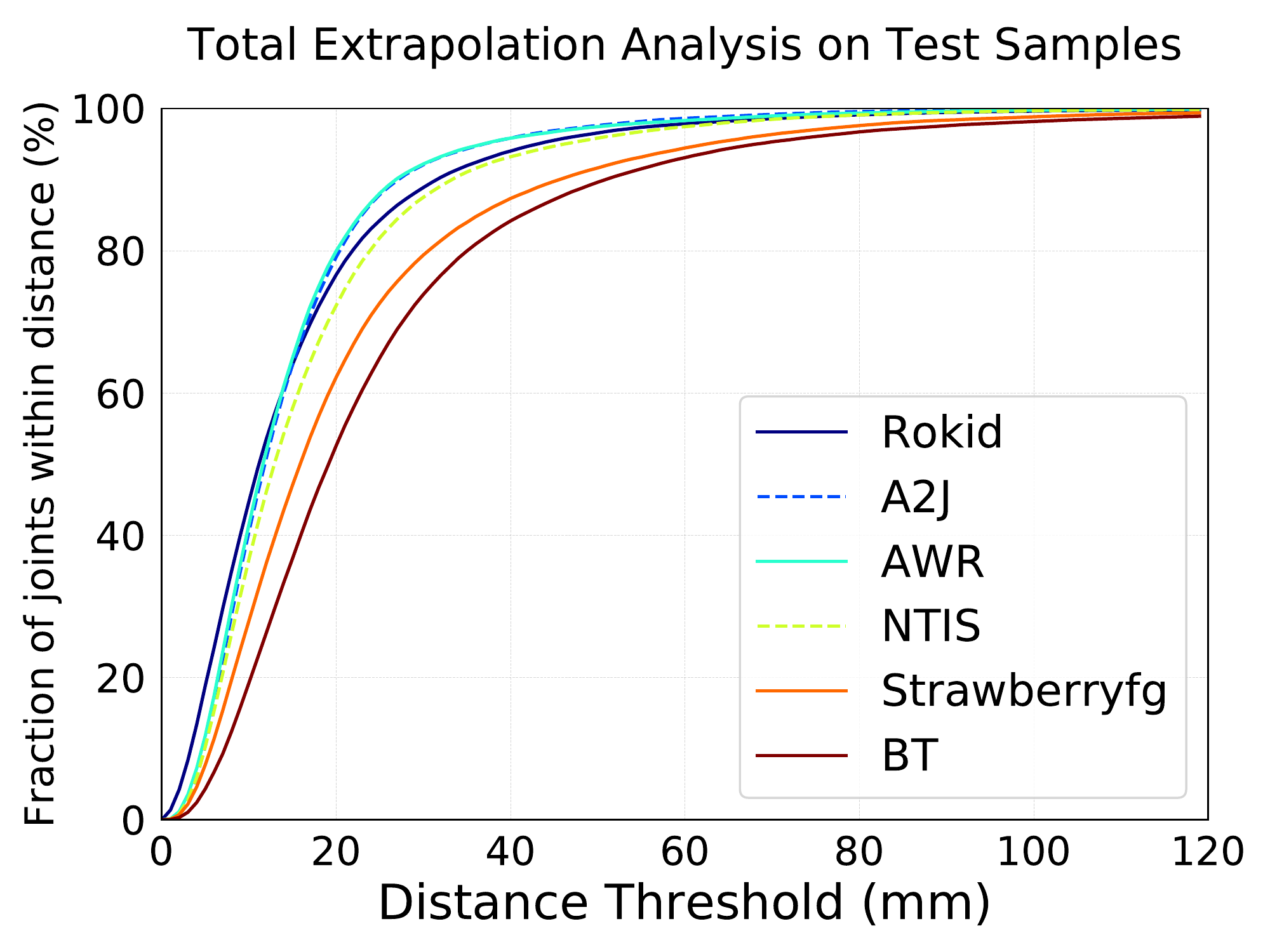}} &
{\includegraphics[width=0.4\linewidth]{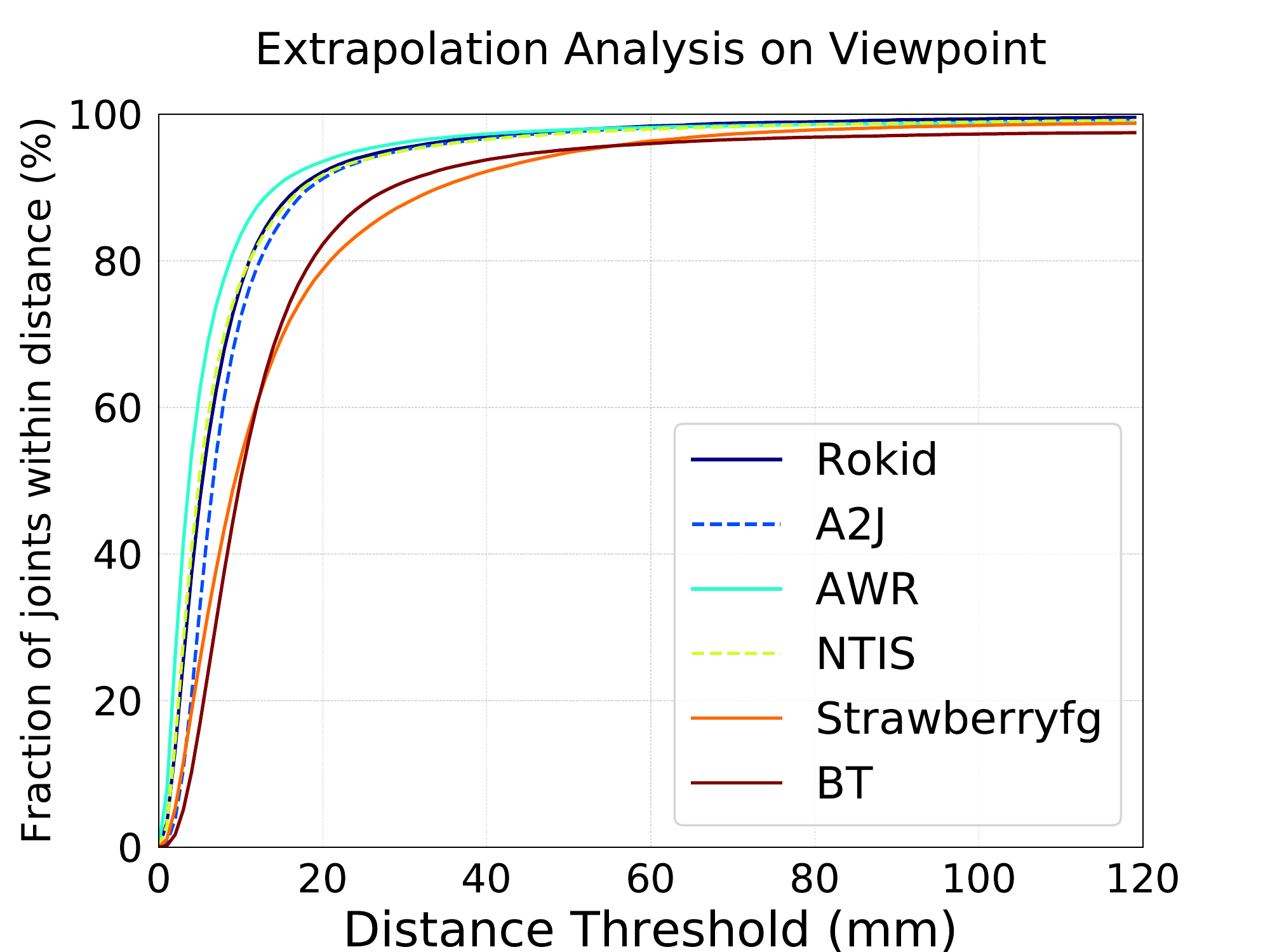}} \vspace{-0pt}\\
\small{(a) Extrapolation} & \small{(b) Viewpoint} \\
{\includegraphics[width=0.4\linewidth]{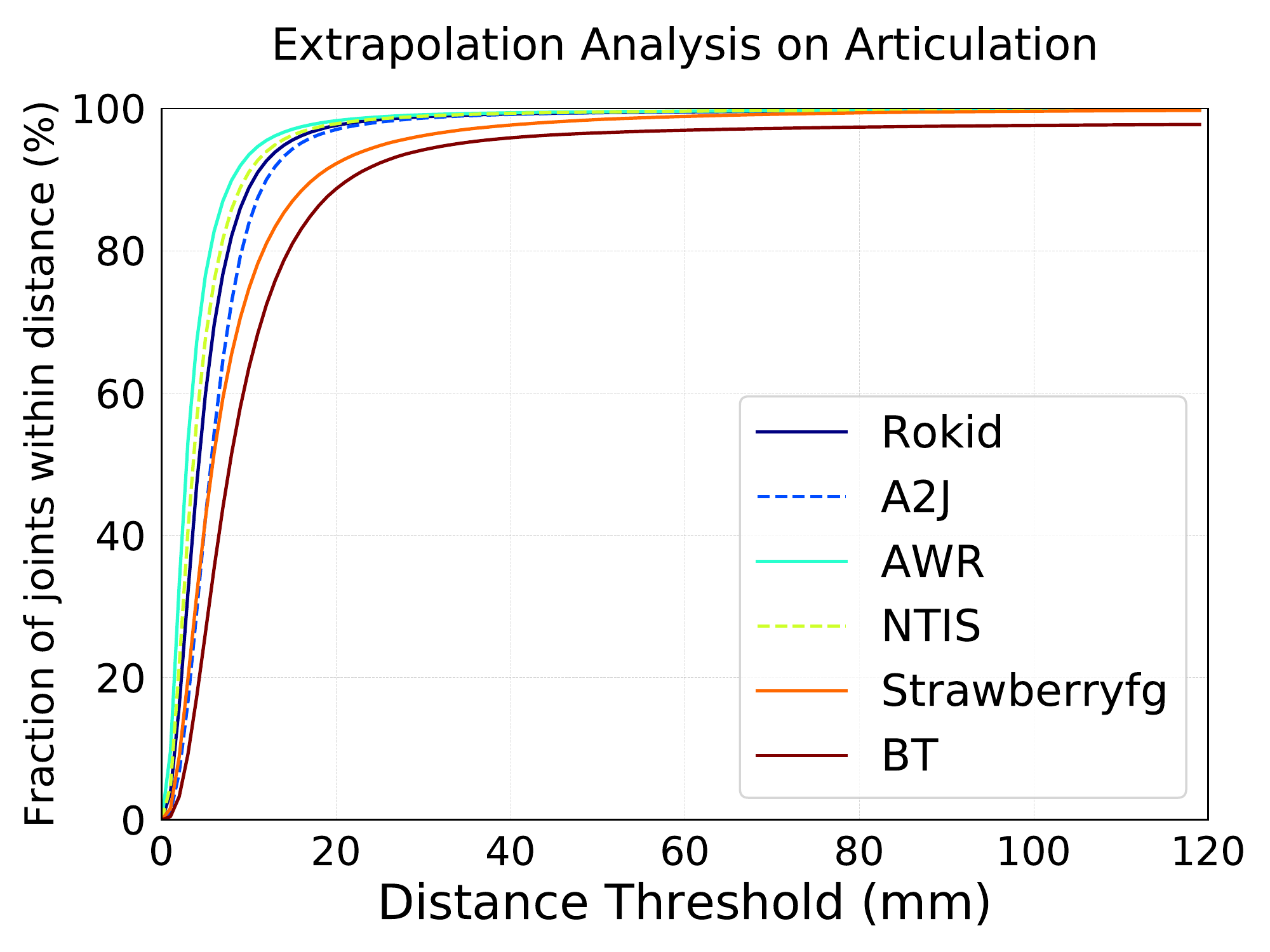}} & 
{\includegraphics[width=0.4\linewidth]{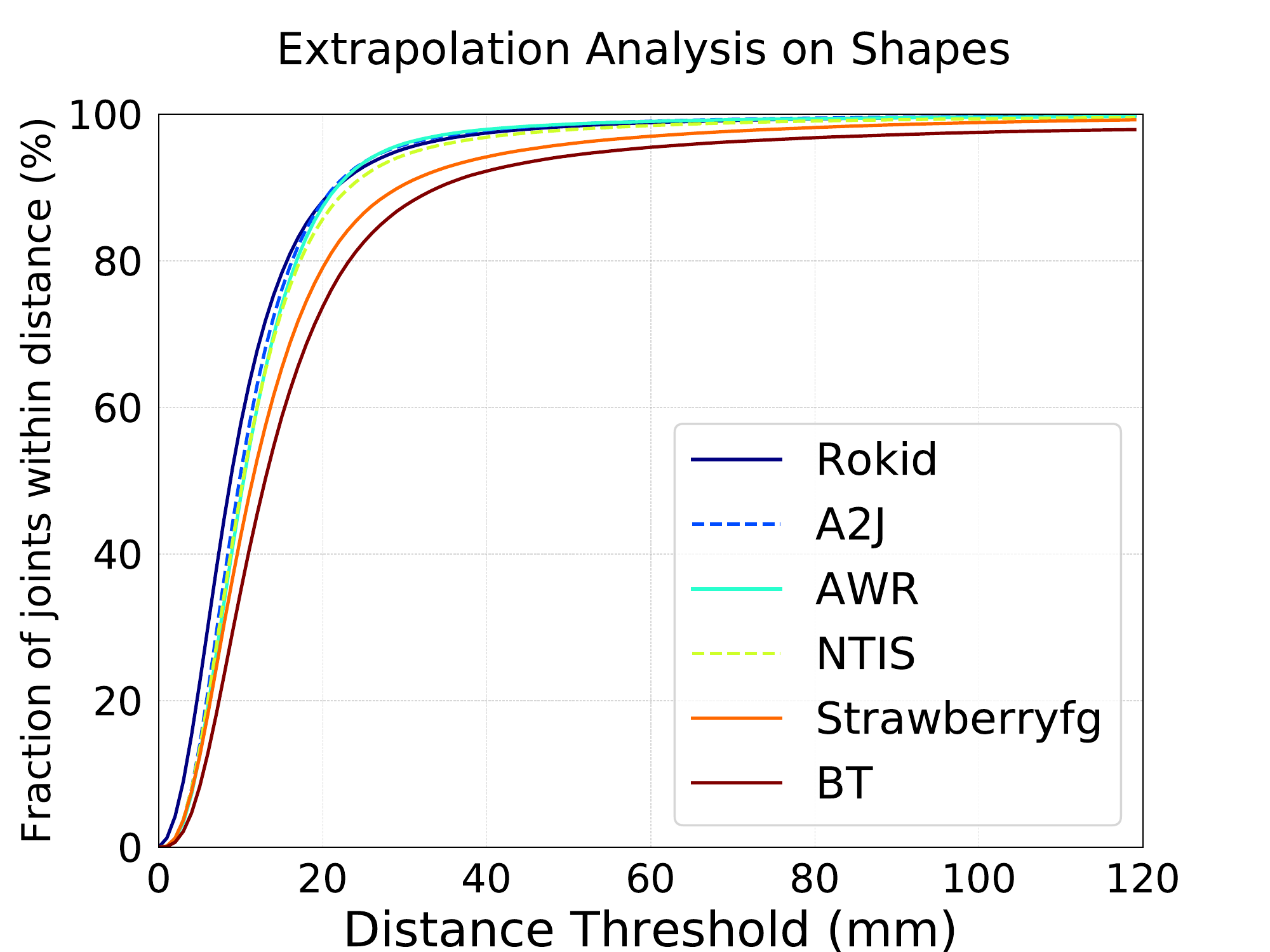}} \vspace{-0pt}\\
\small{(c) Articulation} & \small{(d) Shape}\\
{\includegraphics[width=0.4\linewidth]{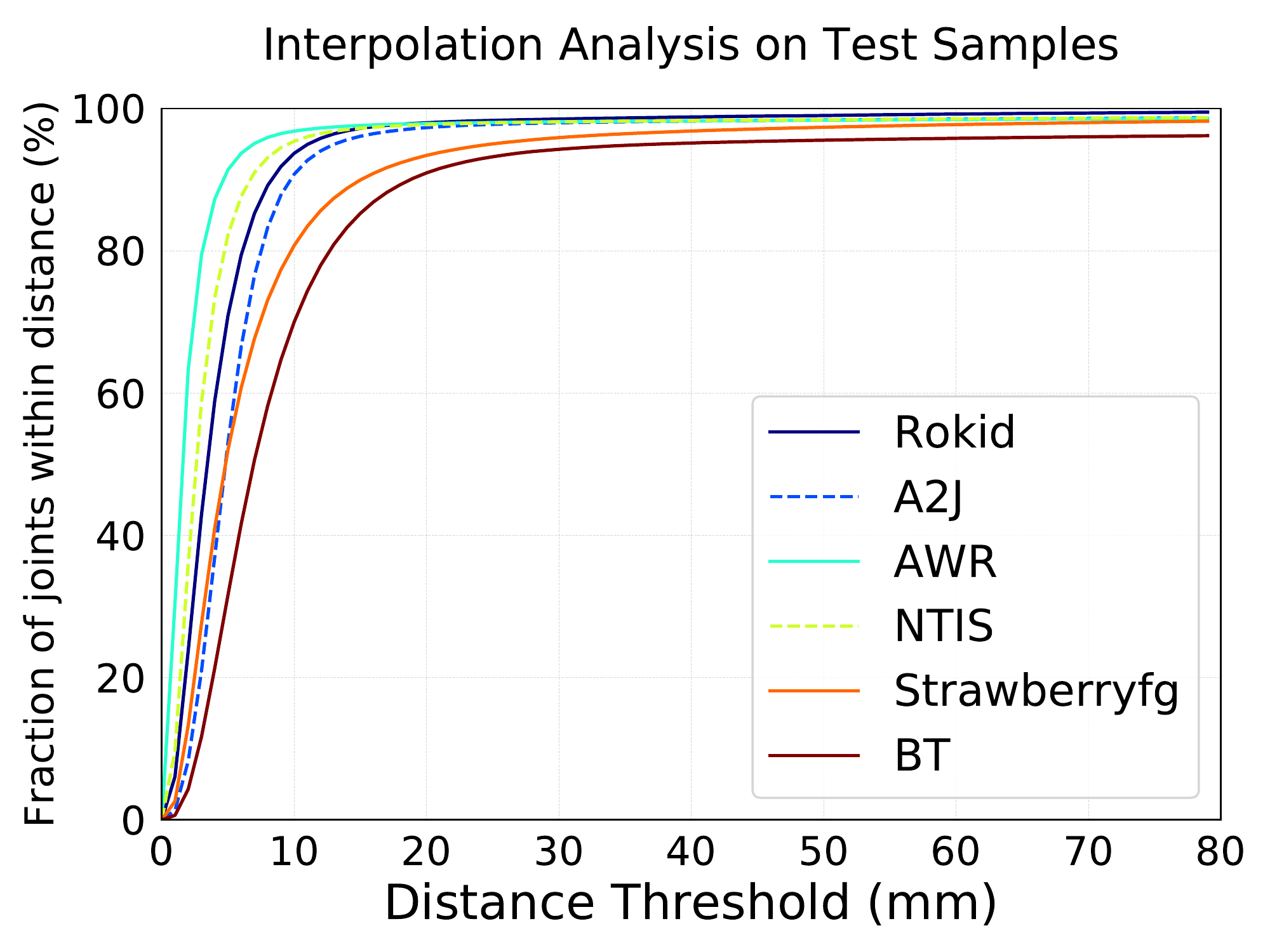}} & \vspace{-0pt}\\ \small{(e) Interpolation} &
\end{tabular}
\caption{Task 1 - Joint success rate analysis on different evaluation axis where each joints' error in the set is evaluated for measuring the accuracy.}
\label{fig:task1_extrapolation_plots_joint}
\end{figure} 

\begin{figure}[!h]
\centering
\begin{tabular}{c c}
{\includegraphics[width=0.4\linewidth]{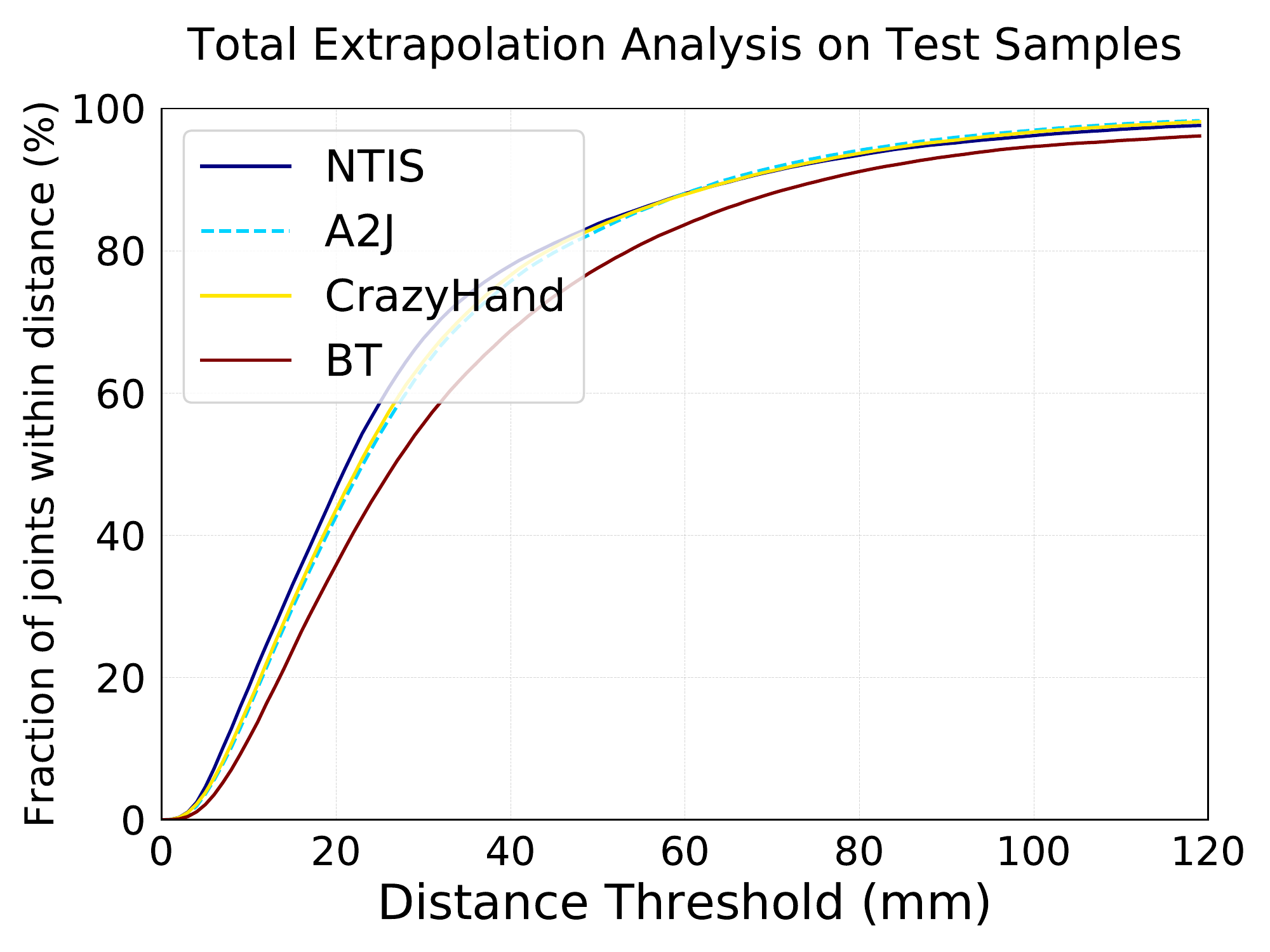}} &
{\includegraphics[width=0.4\linewidth]{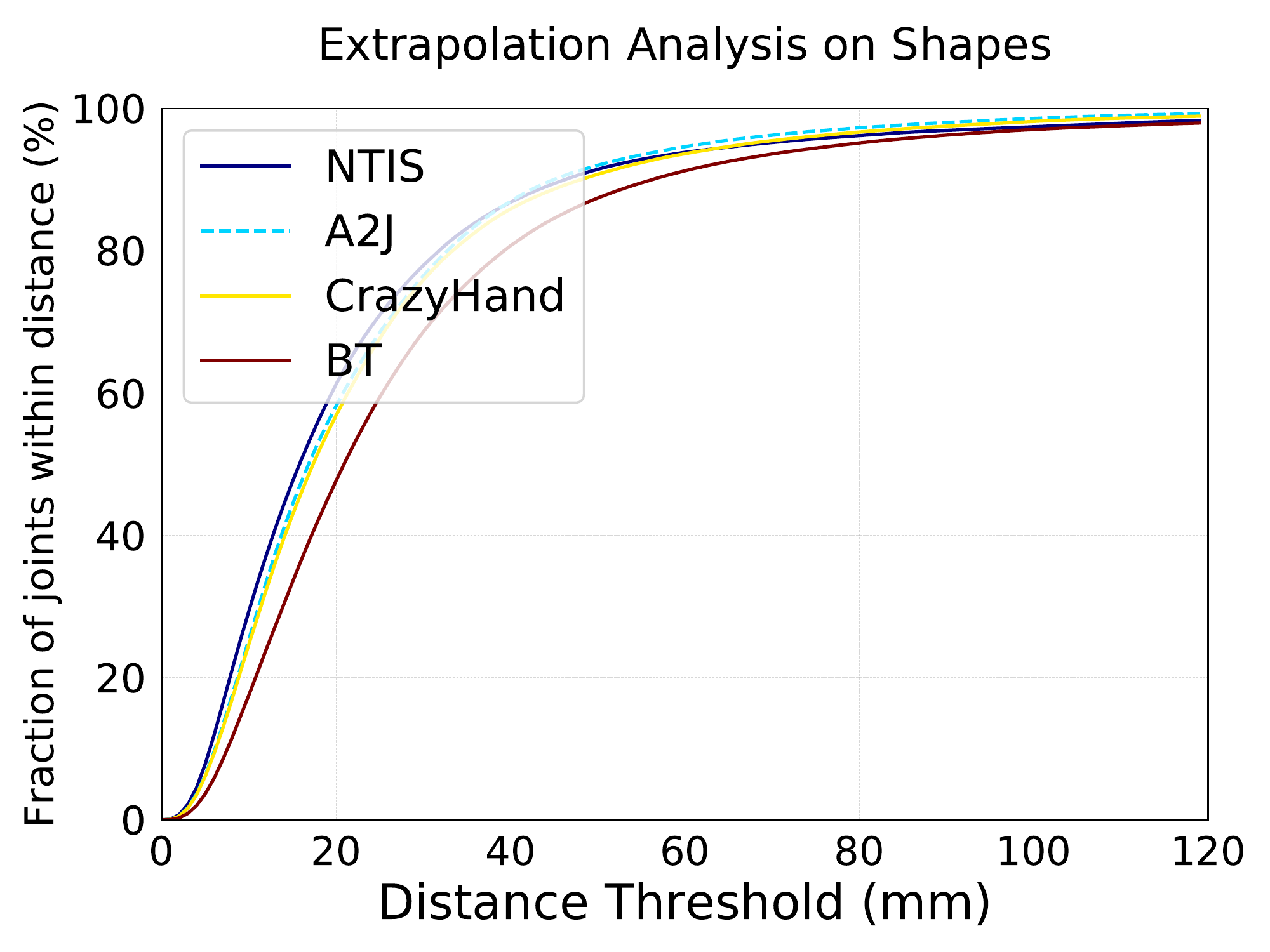}} \vspace{-0pt}\\
\small{(a) Extrapolation} & \small{(b) Shape} 
\\
{\includegraphics[width=0.4\linewidth]{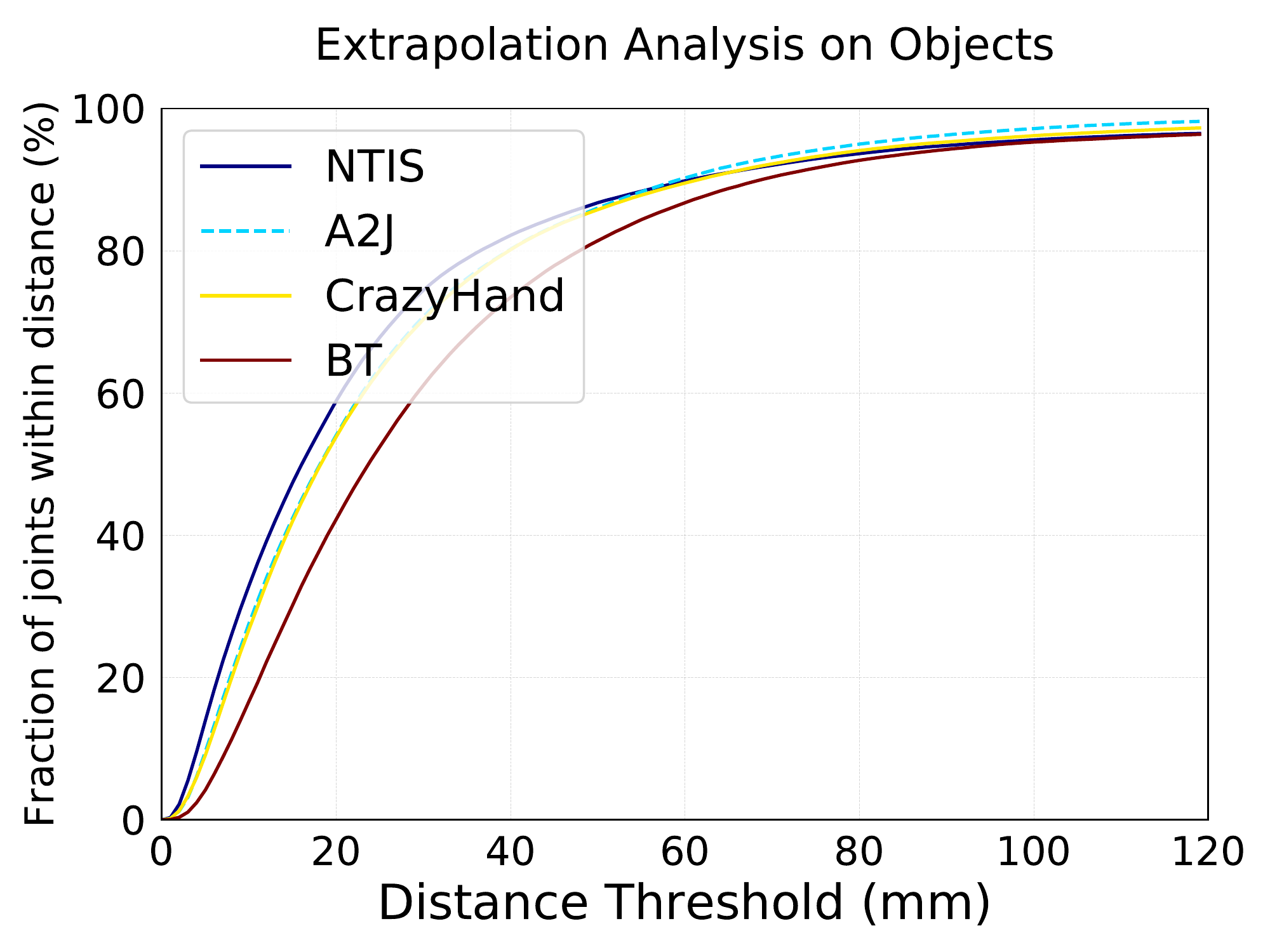}} &
{\includegraphics[width=0.4\linewidth]{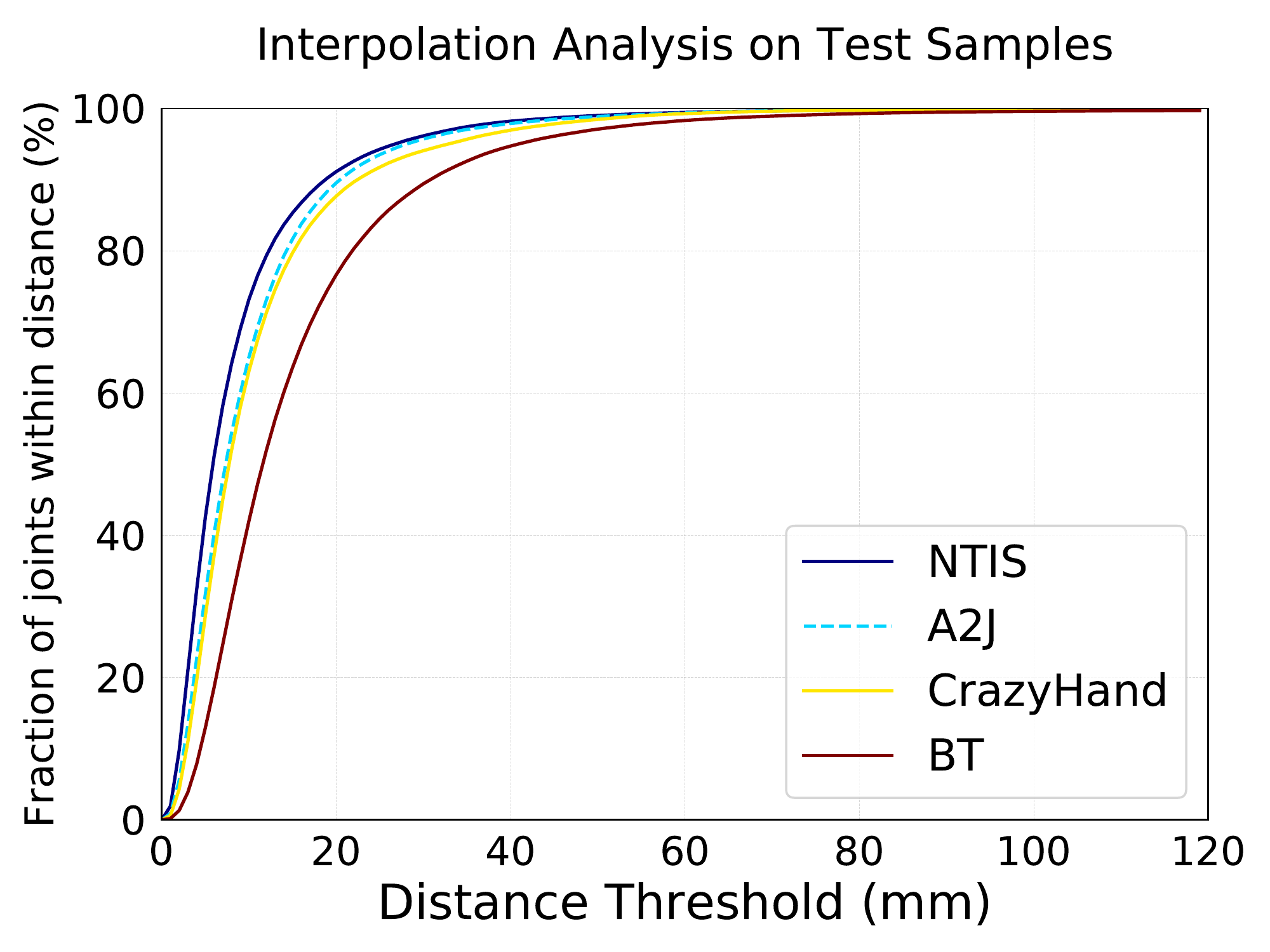}} \vspace{-0pt}\\
\small{(c) Object} & \small{(d) Interpolation} \vspace{-0pt} 
\end{tabular}
\caption{Task 2 - Joint success rate analysis on different evaluation axis where each joints' error in the set is evaluated for measuring the accuracy.}
\label{fig:task2_extrapolation_plots_joint}
\end{figure} 

\begin{figure}[!h]
\centering
\begin{tabular}{c c}
{\includegraphics[width=0.4\linewidth]{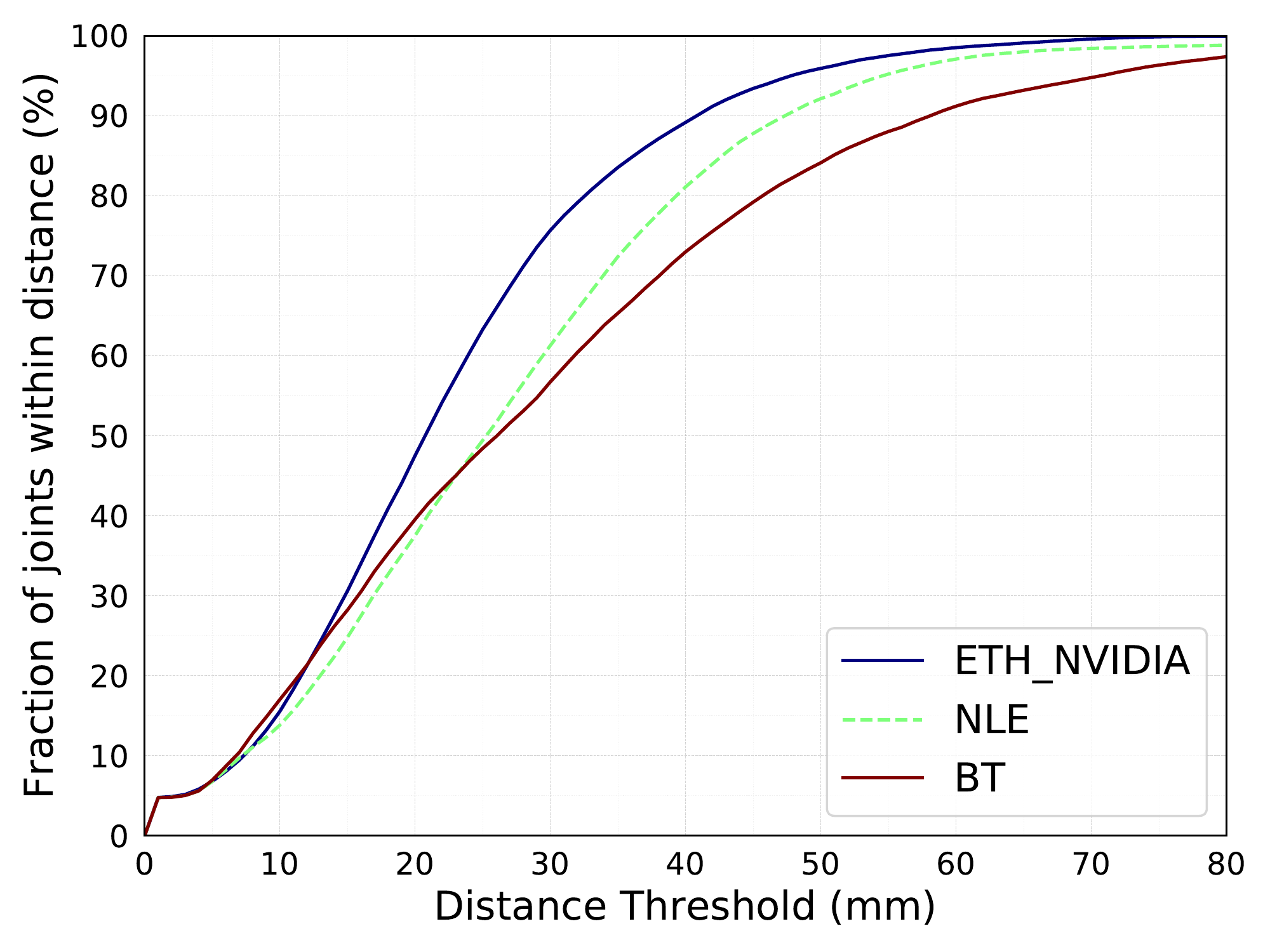}} &  
{\includegraphics[width=0.4\linewidth]{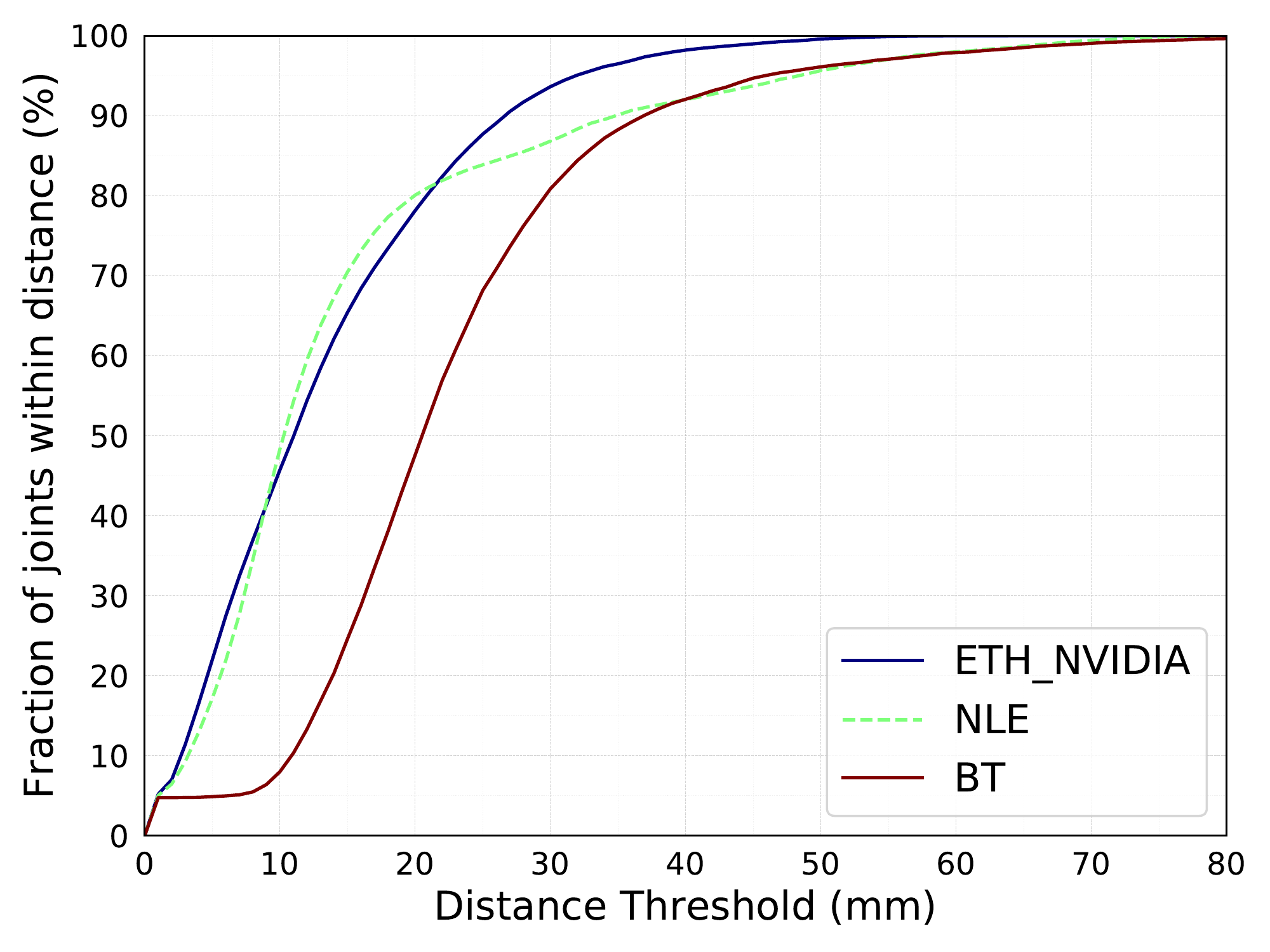}}\\
\small{(a) Extrapolation} & \small{(b) Shape} 
\\
{\includegraphics[width=0.4\linewidth]{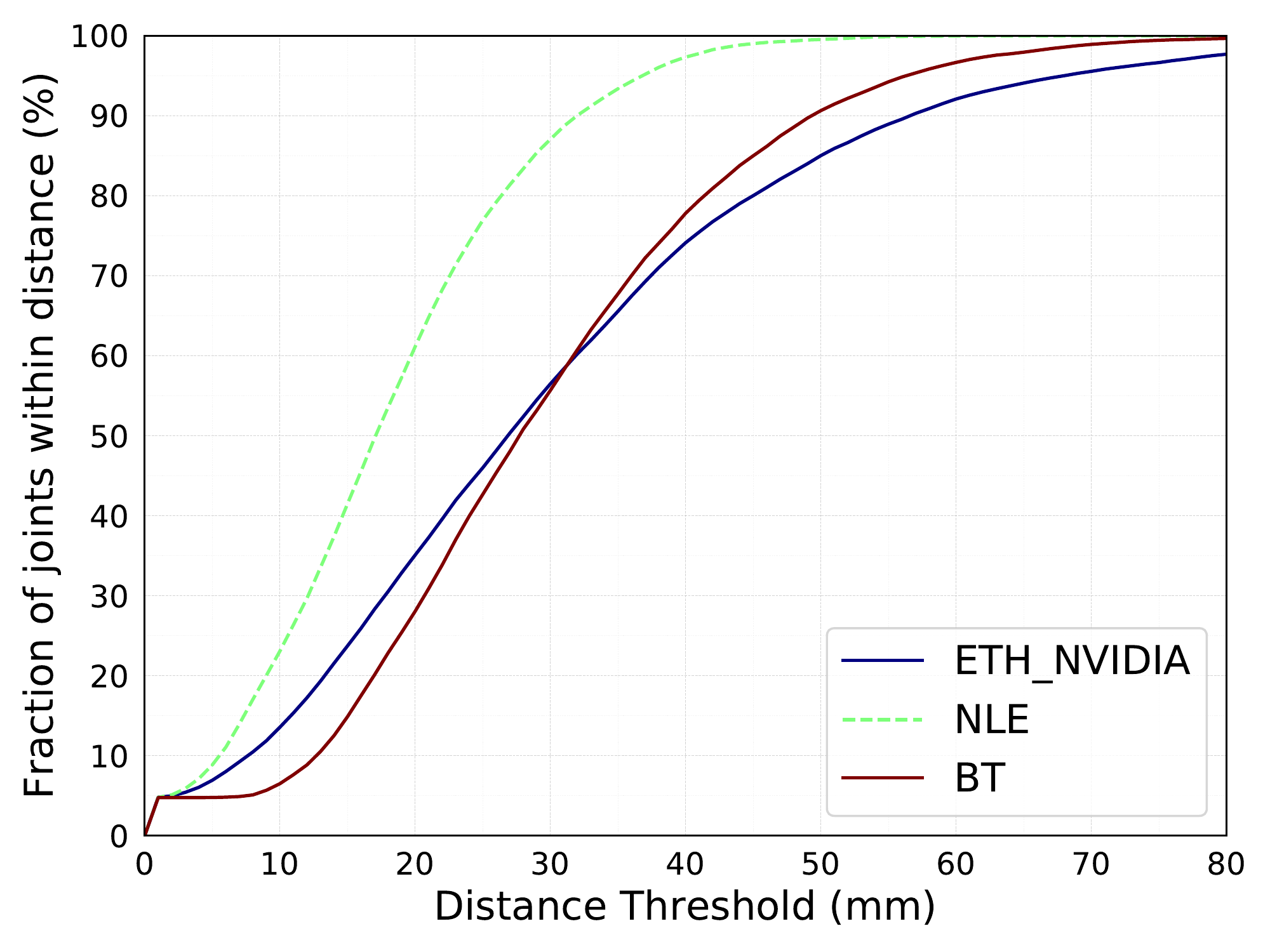}} &
{\includegraphics[width=0.4\linewidth]{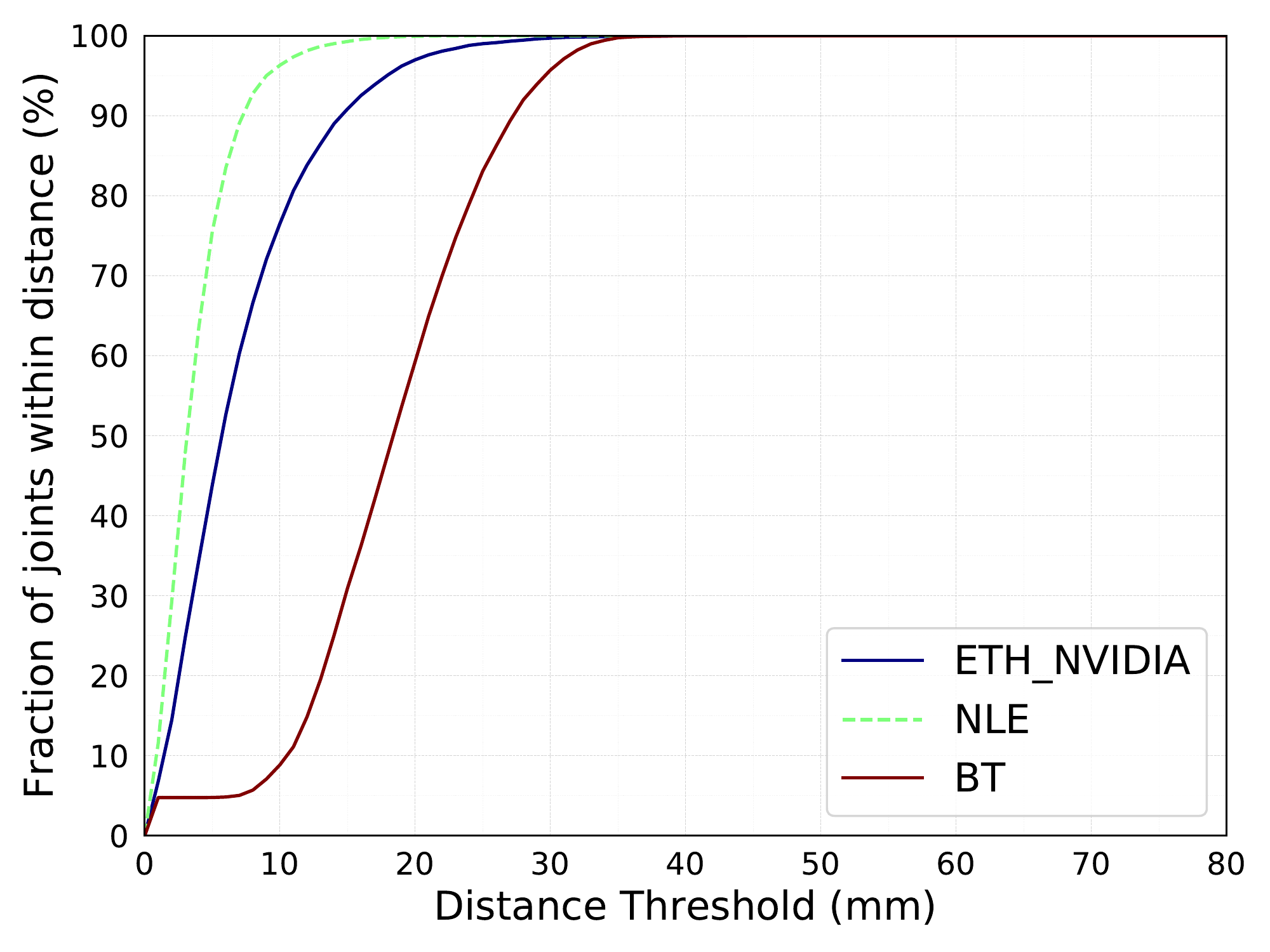}}\\
\small{(c) Object} & \small{(d) Interpolation}
\end{tabular}
\caption{ Task 3 - Joint success rate analysis on different evaluation axis where each joints' error in the set is evaluated for measuring the accuracy.}
\label{fig:task3_extrapolation_plots_joint}
\end{figure}
\clearpage

\subsection{Visualizations for Articulation Clusters, Hand Shapes and Object Types}
\label{appendix:visual}
\label{sec:appendix_vis}
\label{sec:data_details}

\begin{figure}[!h]
\centering
\begin{tabular}{c c c}
\begin{minipage}{.27\linewidth}
\centering
{\includegraphics[width=1\linewidth]{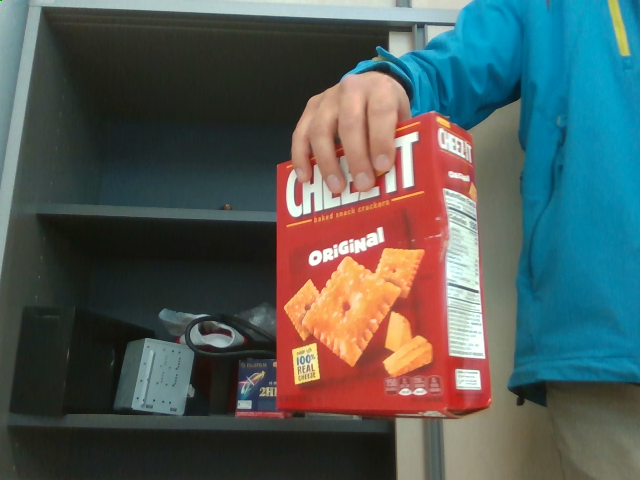}} \\
{(a) O1 - Cracker box}
\end{minipage} 
&
\begin{minipage}{.27\linewidth}
\centering
{\includegraphics[width=1\linewidth]{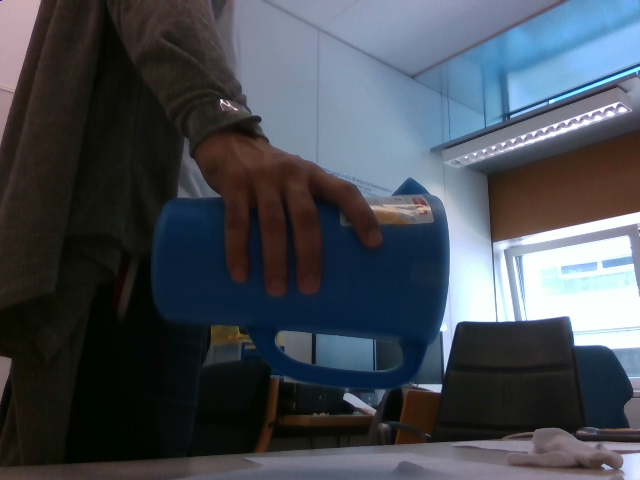}} \\
{(b) O2 - Pitcher base}
\end{minipage}
&
\begin{minipage}{.27\linewidth}
\centering
{\includegraphics[width=1\linewidth]{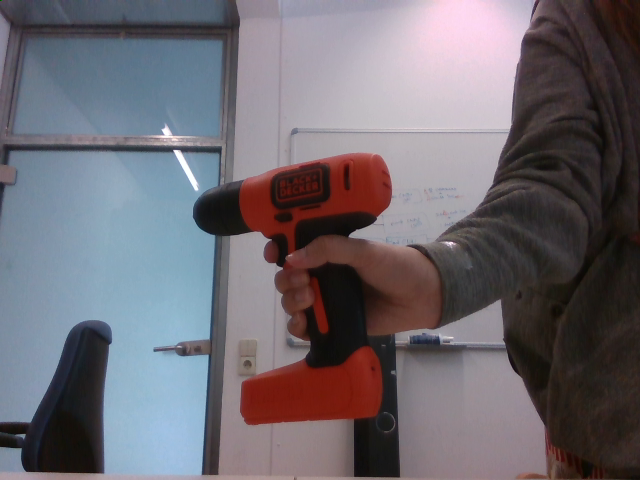}} \\
{(c) O3 - Power drill}
\end{minipage}
\\
\begin{minipage}{.27\linewidth}
\centering
{\includegraphics[width=1\linewidth]{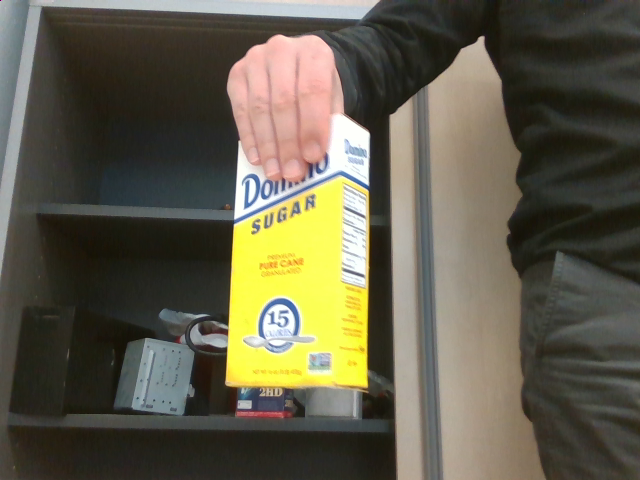}} \\
{(d) O4 - Sugar box}
\end{minipage}
&
\begin{minipage}{.27\linewidth}
\centering
{\includegraphics[width=1\linewidth]{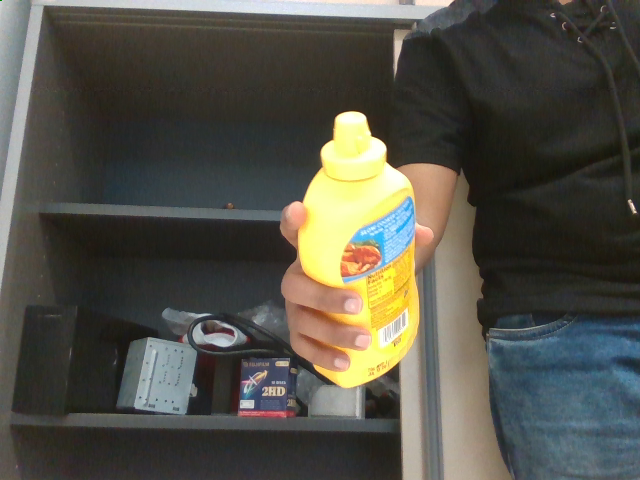}} \\
{(e) O5 - Mustard bottle}
\end{minipage}
&
\begin{minipage}{.27\linewidth}
\centering
{\includegraphics[width=1\linewidth]{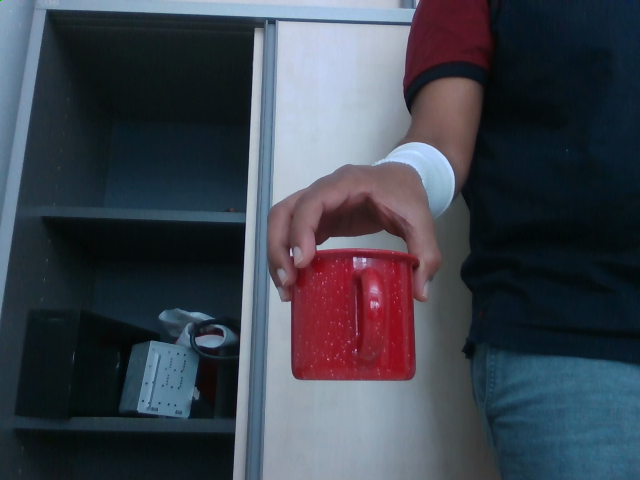}} \\
{(f) O6 - Mug}
\end{minipage}
\end{tabular}
\vspace{-10pt}
\caption{Example frames for the objects appear in Task 3, HO-3D~\cite{ho3d_2019} dataset.}
\label{fig:ho3d_objects}
\end{figure}


\begin{table}
\centering
    \begin{tabular}{c | c c}
        Object Id & Object Name & Seen in the Training Set\\
        \hline
        O1 & cracker box & \cmark\\
        O2 & pitcher base & \cmark\\
        O3 & power drill & \xmark\\
        O4 & sugar box & \cmark\\
        O5 & mustard bottle & \cmark\\
        O6 & mug & \xmark\\
    \end{tabular} \\
    \vspace{5pt}
    {(c) Object List }
\caption{List of seen and unseen objects in the training dataset of Task 3.}
\label{fig:task3_object_list}
\end{table}

\begin{figure}
\centering
{\includegraphics[width=0.45\linewidth]{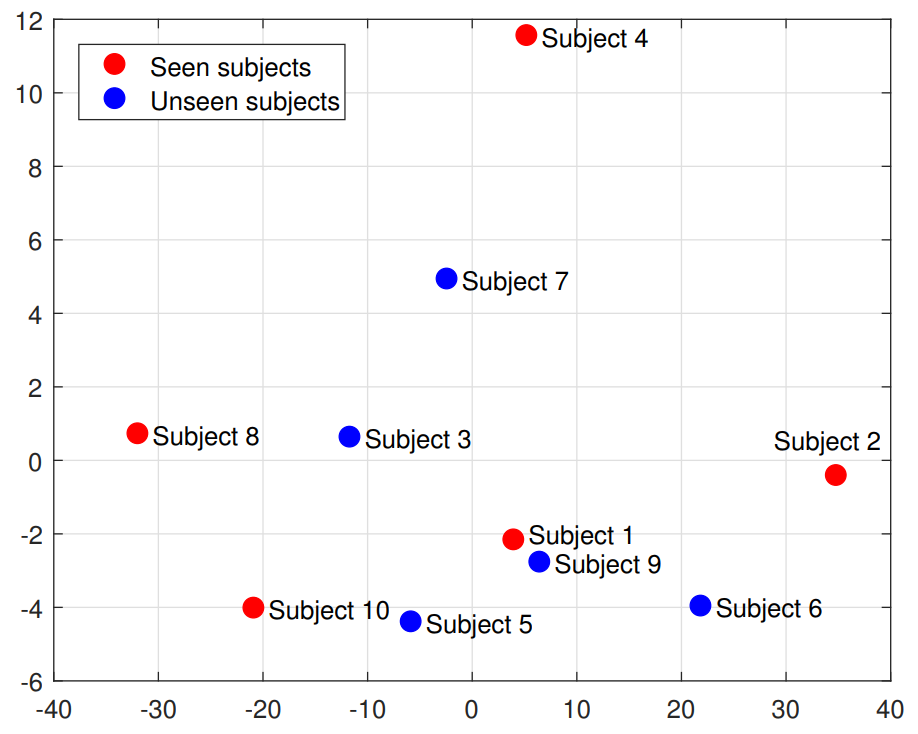}}
\vspace{-10pt}
\caption{Visualization of different hand shape distributions, appear in~\cite{yuan20172017}, by using the first two principal components of the hand shape parameters. Figure is taken from~\cite{yuan20172017}.}
\label{fig:hands17_shape_analysis}
\end{figure}

\newcommand{\clustersize}{0.39}
\begin{figure*}[!htbp]
\centering
\pagebreak
\begin{minipage}{.12\linewidth}
C1 00000\\C2 00001\\C3 00010
\end{minipage}
\begin{minipage}{.28\linewidth}
\centering
\includegraphics[width=\clustersize\linewidth,height=\clustersize\linewidth]{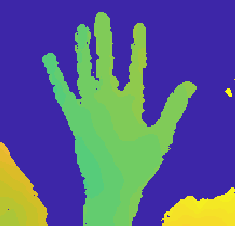}  \includegraphics[width=\clustersize\linewidth,height=\clustersize\linewidth]{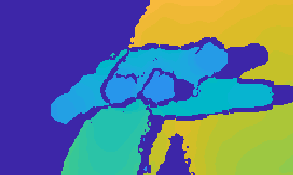} 
\end{minipage} \vline
\begin{minipage}{.28\linewidth}
\centering
\includegraphics[width=\clustersize\linewidth,height=\clustersize\linewidth]{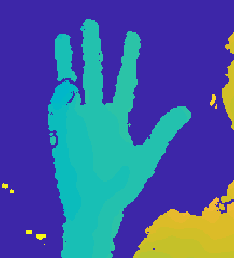}  \includegraphics[width=\clustersize\linewidth,height=\clustersize\linewidth]{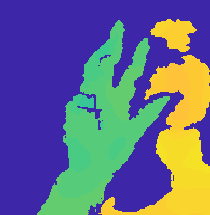} 
\end{minipage} \vline
\begin{minipage}{.28\linewidth}
\centering
\includegraphics[width=\clustersize\linewidth,height=\clustersize\linewidth]{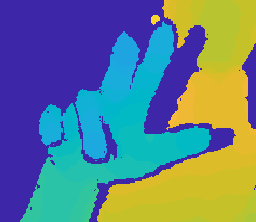}  \includegraphics[width=\clustersize\linewidth,height=\clustersize\linewidth]{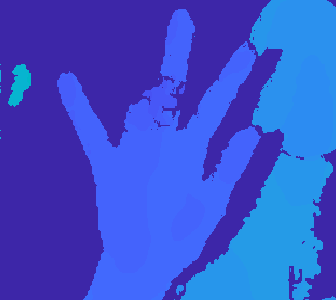} 
\end{minipage} 
\\

\begin{minipage}{.12\linewidth}
C4 00011\\C5 00100\\C6 00101
\end{minipage}
\begin{minipage}{.28\linewidth}
\centering
\includegraphics[width=\clustersize\linewidth,height=\clustersize\linewidth]{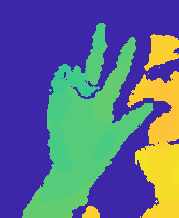}  \includegraphics[width=\clustersize\linewidth,height=\clustersize\linewidth]{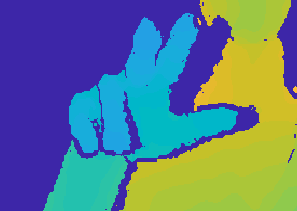} 
\end{minipage} \vline
\begin{minipage}{.28\linewidth}
\centering
\includegraphics[width=\clustersize\linewidth,height=\clustersize\linewidth]{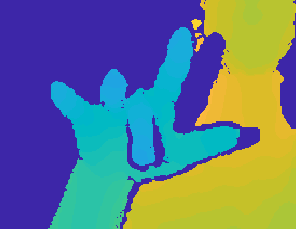}  \includegraphics[width=\clustersize\linewidth,height=\clustersize\linewidth]{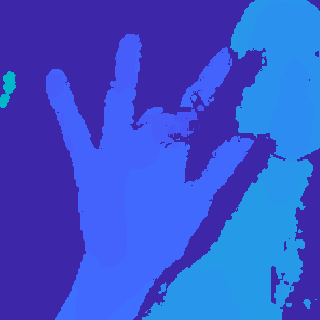} 
\end{minipage} \vline
\begin{minipage}{.28\linewidth}
\centering
\includegraphics[width=\clustersize\linewidth,height=\clustersize\linewidth]{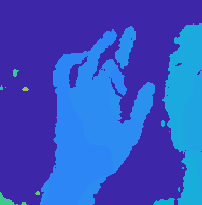}  \includegraphics[width=\clustersize\linewidth,height=\clustersize\linewidth]{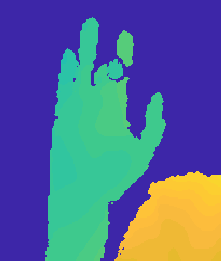} 
\end{minipage}
\\
\begin{minipage}{.12\linewidth}
C7 00110\\C8 00111\\C9 01000
\end{minipage}
\begin{minipage}{.28\linewidth}
\centering
\includegraphics[width=\clustersize\linewidth,height=\clustersize\linewidth]{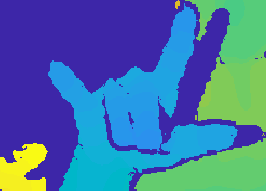}  \includegraphics[width=\clustersize\linewidth,height=\clustersize\linewidth]{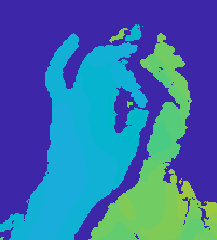} 
\end{minipage} \vline
\begin{minipage}{.28\linewidth}
\centering
\includegraphics[width=\clustersize\linewidth,height=\clustersize\linewidth]{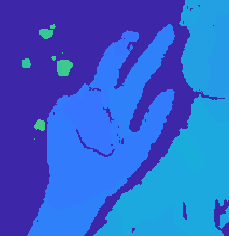}  \includegraphics[width=\clustersize\linewidth,height=\clustersize\linewidth]{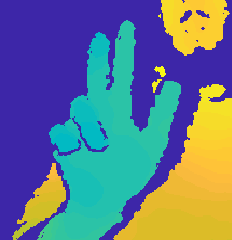} 
\end{minipage} \vline
\begin{minipage}{.28\linewidth}
\centering
\includegraphics[width=\clustersize\linewidth,height=\clustersize\linewidth]{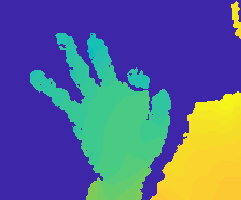}  \includegraphics[width=\clustersize\linewidth,height=\clustersize\linewidth]{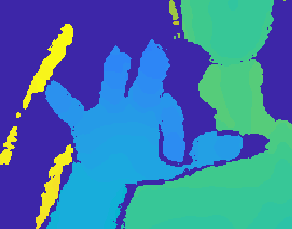} 
\end{minipage} 
\\
\begin{minipage}{.12\linewidth}
C10 01001\\C11 01010\\C12 01011
\end{minipage}
\begin{minipage}{.28\linewidth}
\centering
\includegraphics[width=\clustersize\linewidth,height=\clustersize\linewidth]{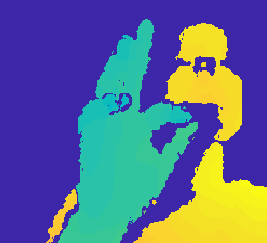}  \includegraphics[width=\clustersize\linewidth,height=\clustersize\linewidth]{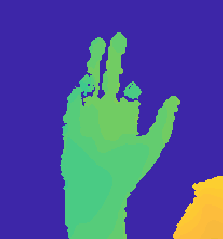} \\
\end{minipage} \vline
\begin{minipage}{.28\linewidth}
\centering
\includegraphics[width=\clustersize\linewidth,height=\clustersize\linewidth]{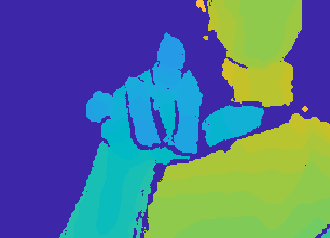}  \includegraphics[width=\clustersize\linewidth,height=\clustersize\linewidth]{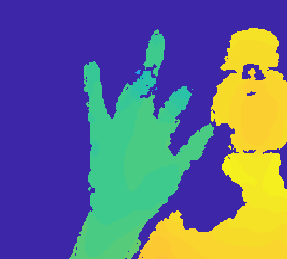} \\
\end{minipage} \vline
\begin{minipage}{.28\linewidth}
\centering
\includegraphics[width=\clustersize\linewidth,height=\clustersize\linewidth]{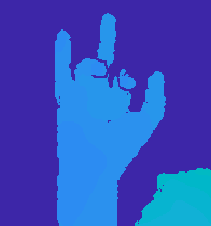}  \includegraphics[width=\clustersize\linewidth,height=\clustersize\linewidth]{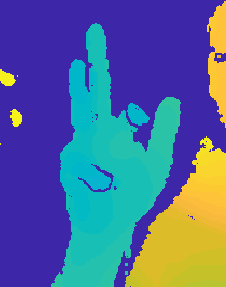} \\
\end{minipage} 
\\
\begin{minipage}{.12\linewidth}
C13 01100\\C14 01101\\C15 01110
\end{minipage}
\begin{minipage}{.28\linewidth}
\centering
\includegraphics[width=\clustersize\linewidth,height=\clustersize\linewidth]{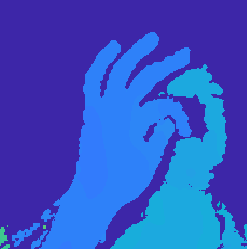}  \includegraphics[width=\clustersize\linewidth,height=\clustersize\linewidth]{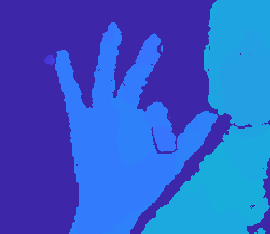} \\
\end{minipage} \vline
\begin{minipage}{.28\linewidth}
\centering
\includegraphics[width=\clustersize\linewidth,height=\clustersize\linewidth]{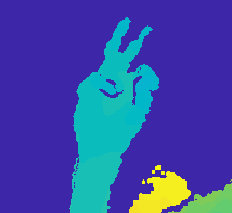}  \includegraphics[width=\clustersize\linewidth,height=\clustersize\linewidth]{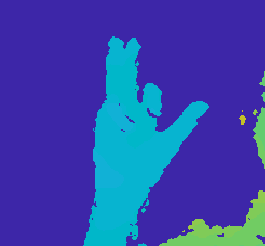} \\
\end{minipage} \vline
\begin{minipage}{.28\linewidth}
\centering
\includegraphics[width=\clustersize\linewidth,height=\clustersize\linewidth]{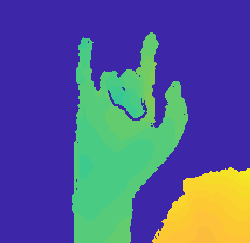}  \includegraphics[width=\clustersize\linewidth,height=\clustersize\linewidth]{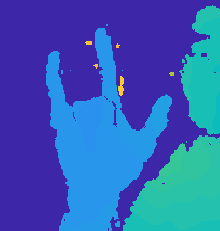} \\
\end{minipage}
\\
\begin{minipage}{.12\linewidth}
C16 01111\\C17 10000\\C18 10001
\end{minipage}
\begin{minipage}{.28\linewidth}
\centering
\includegraphics[width=\clustersize\linewidth,height=\clustersize\linewidth]{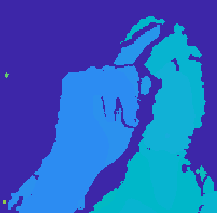}  \includegraphics[width=\clustersize\linewidth,height=\clustersize\linewidth]{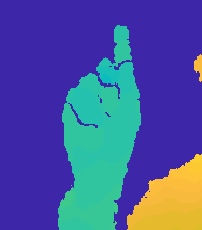} \\
\end{minipage} \vline
\begin{minipage}{.28\linewidth}
\centering
\includegraphics[width=\clustersize\linewidth,height=\clustersize\linewidth]{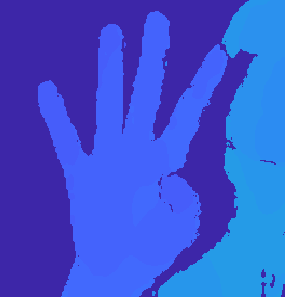}  \includegraphics[width=\clustersize\linewidth,height=\clustersize\linewidth]{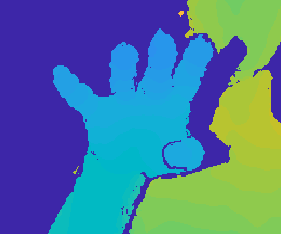} \\
\end{minipage} \vline
\begin{minipage}{.28\linewidth}
\centering
\includegraphics[width=\clustersize\linewidth,height=\clustersize\linewidth]{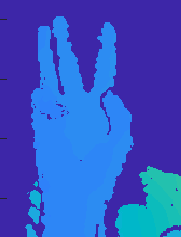} 
\includegraphics[width=\clustersize\linewidth,height=\clustersize\linewidth]{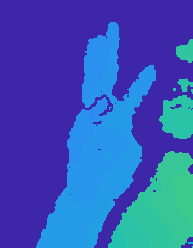} \\
\end{minipage} 
\\
\begin{minipage}{.12\linewidth}
C19 10010\\C20 10011\\C21 10100
\end{minipage}
\begin{minipage}{.28\linewidth}
\centering
\includegraphics[width=\clustersize\linewidth,height=\clustersize\linewidth]{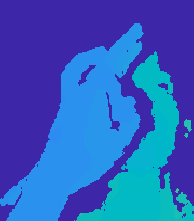}  \includegraphics[width=\clustersize\linewidth,height=\clustersize\linewidth]{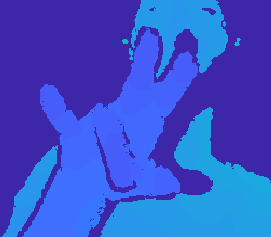} \\
\end{minipage} \vline
\begin{minipage}{.28\linewidth}
\centering
\includegraphics[width=\clustersize\linewidth,height=\clustersize\linewidth]{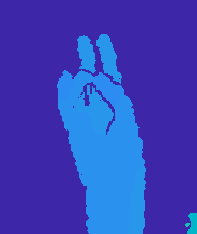}  \includegraphics[width=\clustersize\linewidth,height=\clustersize\linewidth]{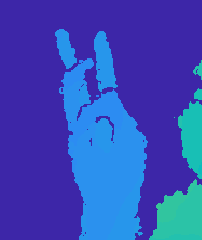} \\
\end{minipage} \vline
\begin{minipage}{.28\linewidth}
\centering
\includegraphics[width=\clustersize\linewidth,height=\clustersize\linewidth]{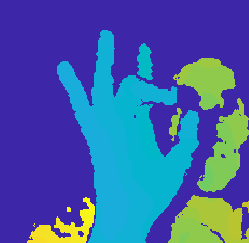}  \includegraphics[width=\clustersize\linewidth,height=\clustersize\linewidth]{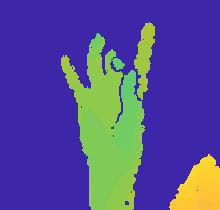} \\
\end{minipage} 
\\
\begin{minipage}{.12\linewidth}
C22 10101\\C23 10110\\C24 10111
\end{minipage}
\begin{minipage}{.28\linewidth}
\centering
\includegraphics[width=\clustersize\linewidth,height=\clustersize\linewidth]{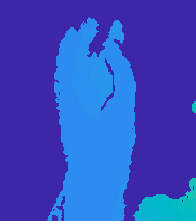}  \includegraphics[width=\clustersize\linewidth,height=\clustersize\linewidth]{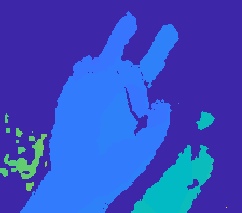} \\
\end{minipage} \vline
\begin{minipage}{.28\linewidth}
\centering
\includegraphics[width=\clustersize\linewidth,height=\clustersize\linewidth]{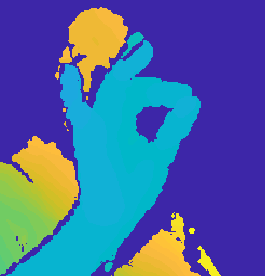}  \includegraphics[width=\clustersize\linewidth,height=\clustersize\linewidth]{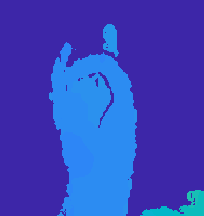} \\
\end{minipage} \vline
\begin{minipage}{.28\linewidth}
\centering
\includegraphics[width=\clustersize\linewidth,height=\clustersize\linewidth]{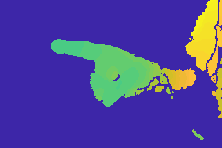}  \includegraphics[width=\clustersize\linewidth,height=\clustersize\linewidth]{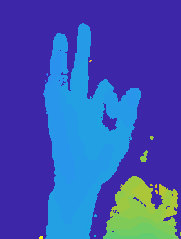} \\
\end{minipage} 
\\
\begin{minipage}{.12\linewidth}
C25 11000\\C26 11001\\C27 11010
\end{minipage}
\begin{minipage}{.28\linewidth}
\centering
\includegraphics[width=\clustersize\linewidth,height=\clustersize\linewidth]{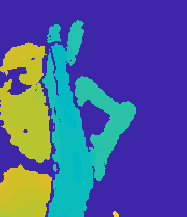}  \includegraphics[width=\clustersize\linewidth,height=\clustersize\linewidth]{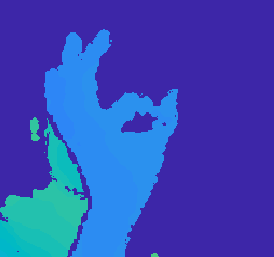} \\
\end{minipage} \vline
\begin{minipage}{.28\linewidth}
\centering
\includegraphics[width=\clustersize\linewidth,height=\clustersize\linewidth]{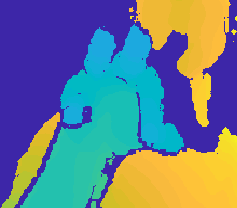}  \includegraphics[width=\clustersize\linewidth,height=\clustersize\linewidth]{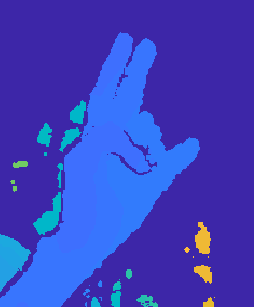} \\
\end{minipage} \vline
\begin{minipage}{.28\linewidth}
\centering
\includegraphics[width=\clustersize\linewidth,height=\clustersize\linewidth]{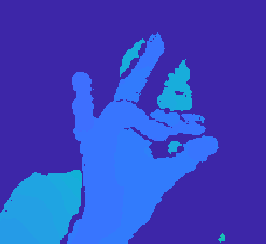}  \includegraphics[width=\clustersize\linewidth,height=\clustersize\linewidth]{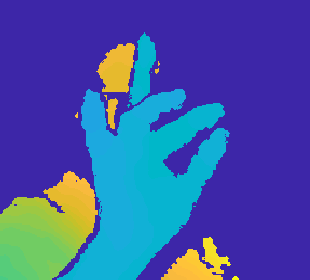} \\
\end{minipage} 
\\

\begin{minipage}{.12\linewidth}
C28 11011\\C29 11100\\C30 11101
\end{minipage}
\begin{minipage}{.28\linewidth}
\centering
\includegraphics[width=\clustersize\linewidth,height=\clustersize\linewidth]{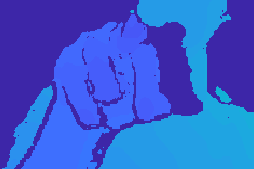}  \includegraphics[width=\clustersize\linewidth,height=\clustersize\linewidth]{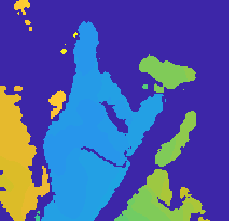} \\
\end{minipage} \vline
\begin{minipage}{.28\linewidth}
\centering
\includegraphics[width=\clustersize\linewidth,height=\clustersize\linewidth]{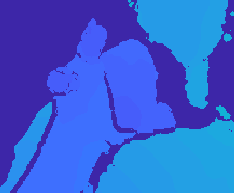}  \includegraphics[width=\clustersize\linewidth,height=\clustersize\linewidth]{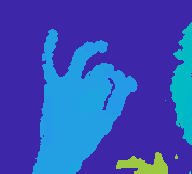} \\
\end{minipage} \vline
\begin{minipage}{.28\linewidth}
\centering
\includegraphics[width=\clustersize\linewidth,height=\clustersize\linewidth]{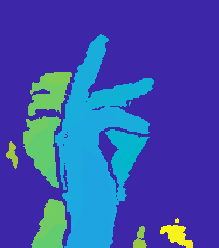}  \includegraphics[width=\clustersize\linewidth,height=\clustersize\linewidth]{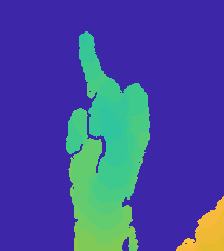} \\
\end{minipage} 
\\

\begin{minipage}{.12\linewidth}
C31 11110\\C32 11111
\end{minipage}
\begin{minipage}{.28\linewidth}
\centering
\includegraphics[width=\clustersize\linewidth,height=\clustersize\linewidth]{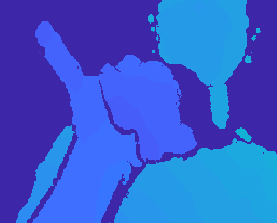}  \includegraphics[width=\clustersize\linewidth,height=\clustersize\linewidth]{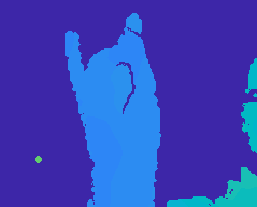} \\
\end{minipage} \vline
\begin{minipage}{.28\linewidth}
\centering
\includegraphics[width=\clustersize\linewidth,height=\clustersize\linewidth]{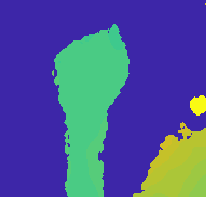}  \includegraphics[width=\clustersize\linewidth,height=\clustersize\linewidth]{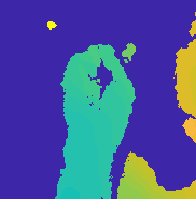} \\
\end{minipage} 

\caption{Examples frames for 32 articulation clusters used in the evaluations. Each row shows cluster ids and their respective binary representations for two example images of three clusters. Each binary representation is constructed from thumb to pinky fingers with 0 representing closed and 1 representing open fingers.}
\label{fig:task1_articulation_cluster_examples}
\end{figure*}

\twocolumn